%% file: arxiv.tex
\documentclass[a4paper,conference]{IEEEtran}
\IEEEoverridecommandlockouts 

\usepackage{ifpdf}
\usepackage{cite}
\usepackage[pdftex]{graphicx}
\usepackage{array}
\usepackage{mdwmath}
\usepackage{mdwtab}
\usepackage{amssymb,latexsym}
\usepackage{stfloats}
\usepackage{amsmath}
\usepackage{subfig}
\usepackage{algcompatible}
\usepackage{algorithm}
\usepackage[font={footnotesize}]{caption}
\usepackage{balance}
\usepackage[utf8]{inputenc}
\usepackage[T1]{fontenc}
\usepackage[compatible]{algpseudocode}

\input{math_commands.tex}

\usepackage[accsupp]{axessibility}
\usepackage{amsfonts}
\usepackage{bm}
\usepackage{booktabs}
\usepackage{xurl}
\usepackage{float}
\usepackage{makecell}
\usepackage{multirow}
\usepackage{needspace}
\usepackage[normalem]{ulem}
\usepackage{orcidlink}
\usepackage[percent]{overpic}
\usepackage{placeins}
\usepackage{siunitx}
\usepackage{standalone}
\standaloneconfig{mode=tex}
\usepackage{textcomp}
\usepackage{tikz}
\usetikzlibrary{arrows.meta,positioning,calc,fit,backgrounds,matrix}
\usepackage{xcolor}
\usepackage{hyperref}
\usepackage{doi}

\renewcommand{\figurename}{Figure}

\DeclareRobustCommand*{\IEEEauthorrefmark}[1]{%
\raisebox{0pt}[0pt][0pt]{\textsuperscript{\footnotesize\ensuremath{#1}}}}

\renewcommand\IEEEkeywordsname{Keywords}

\newcommand{\paneltag}[2][1.0em]{%
  \begin{tikzpicture}[baseline=(n.base)]
    \node[
      inner sep=0pt,
      minimum width=#1,
      minimum height=#1,
      fill=black!55,
      text=white,
      font=\bfseries\footnotesize,
      align=center
    ] (n) {#2};
  \end{tikzpicture}%
}

\newcommand{\panel}[3][0.32\linewidth]{%
  \begin{overpic}[width=#1]{#2}%
    \put(0,100){\makebox(0,0)[tl]{\paneltag{#3}}}%
  \end{overpic}%
}

\newcommand\copyrighttext{\footnotesize
  \textcopyright~2026 IEEE. Personal use of this material is permitted. Permission from IEEE must be obtained for all other uses, in any current or future media, including reprinting/republishing this material for advertising or promotional purposes, creating new collective works, for resale or redistribution to servers or lists, or reuse of any copyrighted component of this work in other works.}
\newcommand\copyrightnotice{%
\begin{tikzpicture}[remember picture,overlay]
\node[anchor=south,yshift=10pt] at (current page.south)
  {\fbox{\parbox{\dimexpr\textwidth-2\fboxsep-2\fboxrule\relax}{\copyrighttext}}};
\end{tikzpicture}}

\begin{document}
\pagestyle{plain}
\bstctlcite{IEEEexample:BSTcontrol}

\title{High-Fidelity Synthetic Transmission Electron Microscopy Image Generation Using Diffusion Probabilistic Models for Data-Limited Semiconductor Metrology}

\author{\IEEEauthorblockN{
Johannes Boehm\,\orcidlink{0009-0007-6564-0243}\IEEEauthorrefmark{1}\textsuperscript{,}\IEEEauthorrefmark{2},
Bappaditya Dey\,\orcidlink{0000-0002-0886-137X}\IEEEauthorrefmark{2}}
\IEEEauthorblockA{\IEEEauthorrefmark{1}
Chemnitz University of Technology, Chemnitz, Germany}
\IEEEauthorblockA{\IEEEauthorrefmark{2}
Interuniversity Microelectronics Centre (imec), Leuven, Belgium}
{\it bappaditya.dey@imec.be}
}

\maketitle
\thispagestyle{plain}
\copyrightnotice

\begin{abstract}
Advanced semiconductor nodes drastically increased demand for Transmission Electron Microscopy (TEM), yet destructive sample preparation, slow imaging and high costs severely limit the availability of diverse datasets needed for downstream machine learning (ML). Synthetic data generation is becoming essential, but current generative models often miss TEM-specific noise, structural detail, and stochastic variability crucial for evaluation. We present a Denoising Diffusion Probabilistic Model (DDPM) framework for synthetic TEM image generation under extreme data scarcity. A progressive patch-based training strategy scales from low-resolution patches to full images, enabling from-scratch training with only 15 samples. We integrate a custom TrivialAugment adaptation, cross-process domain transfer, classifier guidance, and RePaint-style inpainting, culminating in full-image generation that preserves the global structural and spatial relationships relevant to FAB metrology. Beyond synthesis, we repurpose DDPM feature representations for segmentation, partitioning encoder feature maps to obtain coherent region masks. On the closest 1\% of nearest-neighbor pairs, our synthetic DEVICE-TEM images reach MS-SSIM $0.985 \pm 0.002$ (full-set mean $0.587$), and a blinded qualitative expert assessment is consistent with the structural similarity results. The synthetic images are intended as training data for downstream ML in defect detection, segmentation, and metrology.
\end{abstract}

\begin{IEEEkeywords}
Transmission Electron Microscopy; Diffusion Probabilistic Models; Synthetic Data; Semiconductor Metrology; Image Generation; Segmentation
\end{IEEEkeywords}

\section{Introduction}
\label{sec:introduction}

Semiconductor scaling has reached advanced nodes at 2\,nm and beyond, enabled by Extreme Ultraviolet Lithography (EUVL) and emerging high-NA EUVL. As critical dimensions shrink, the challenges associated with process control, defect inspection, and metrology have escalated substantially and are expected to intensify further \cite{lorusso_metrology_2022, chen_improved_2024}.

Transmission Electron Microscopy (TEM) is indispensable for advanced-node metrology, delivering atomic-scale resolution for visualizing nanometer-scale structures and defects. High-quality TEM images are critical for defect classification, thickness measurements, and process optimization, yet comprehensive dataset generation is constrained by destructive sample preparation, limited sample diversity, and time- and expertise-intensive acquisition. This data scarcity presents a fundamental challenge to the development of both conventional and ML-based analysis tools, despite growing evidence that synthetic data can improve downstream performance \cite{govind_deep_2024}.

Denoising Diffusion Probabilistic Models (DDPMs) have shown great success in generating high-fidelity synthetic images across domains \cite{ho_denoising_2020b, nichol_improved_2021b, dhariwal_diffusion_2021b}, and their application to semiconductor imaging has been explored for SEM-based defect inspection \cite{deridder_semidiffusioninst_2023, dey_addressing_2024}. Unlike SEM imaging, which is primarily surface-sensitive and non-destructive, TEM enables transmission imaging through ultra-thin samples of \textless100\,nm with sub-angstrom resolution, making it suited for analyzing internal device structures and atomic-scale features. While data augmentation techniques have been shown to substantially enhance ML model training in electron microscopy \cite{shagadevan_improved_2021, chen_advancing_2024, kazimi_selfsupervised_2024}, applying diffusion models for robust TEM image synthesis remains largely unexplored.

We propose custom-trained diffusion models for high-fidelity synthetic TEM generation from limited real-wafer datasets. The approach trains from scratch, enabling domain-specific synthesis that preserves the characteristics of real TEM data. A progressive patch-to-image training strategy reproduces both fine-grained textures and large-scale structural features while remaining grounded in the original data distribution. Beyond synthesis, we leverage intermediate DDPM representations for unsupervised segmentation \cite{baranchuk_labelefficient_2021, couairon_diffcut_2024a}, demonstrating that generative features capture transferable microstructural cues and can yield practical region proposals that are efficiently curated with minimal effort. This generative framework complements existing efforts on automated acquisition and downstream analysis \cite{spurgeon_datadriven_2021, botifoll_machine_2022} by alleviating TEM data bottlenecks and providing controllable, domain-consistent synthetic imagery and feature representations.

The main contributions of this research are: \textbf{(i)} a patch-based progressive training methodology for DDPMs that enables high-fidelity, high-resolution synthetic TEM image generation from extremely limited datasets, including scenarios with as few as 15 training images, producing synthetic images largely indistinguishable from real wafer TEM data that satisfy the structural and noise-statistical criteria evaluated in Sec.~\ref{sec:results}; \textbf{(ii)} adaptation of the TA algorithm \cite{muller_trivialaugment_2021b} for TEM imaging and integration of semiconductor imaging-aware augmentation methods \cite{chen_improved_2024}; \textbf{(iii)} demonstration of domain transfer capabilities, facilitating synthetic TEM image generation that preserves characteristics from varying imaging conditions or process parameters; \textbf{(iv)} integration of guidance mechanisms for controllable synthesis, enabling generation of images with specific structural properties relevant to semiconductor metrology; \textbf{(v)} adaptation of state-of-the-art inpainting capability \cite{lugmayr_repaint_2022b} within the diffusion framework, enabling seamless image extension and resolution enhancement; \textbf{(vi)} experimental evaluation of DDPM intermediate feature representations for unsupervised TEM segmentation \cite{baranchuk_labelefficient_2021, couairon_diffcut_2024a}, providing coherent microstructural region proposals that serve as lightweight starting points for downstream segmentation workflows.

\section{Related Work}
\label{sec:related_work}

\subsection{Simulation-Based}
TEM image synthesis primarily relies on supervised methods like physics-based simulations modeling electron-specimen interactions. Several packages provide frameworks, including abTEM \cite{madsen_abtem_2021}, Prismatic \cite{rangeldacosta_prismatic_2021}, and conventional multislice algorithms \cite[p.~180]{kirkland_advanced_2020}. They allow precise control over imaging parameters and deliver physically accurate representations of electron scattering processes. However, these methods face limitations when applied to large-scale device structures in semiconductors, as they require extensive prior knowledge of specimen composition, imaging conditions, and microscope parameters, which may not always be available. While simulation frameworks such as Construction Zone \cite{rangeldacosta_robust_2024} have demonstrated success in generating synthetic datasets for nanoparticle segmentation, conventional simulations remain constrained by their reliance on predefined physical models. They may not capture the full complexity of real experimental conditions and become increasingly computationally demanding as the simulated specimen size grows. Consequently, they are not well-suited for generating large-scale datasets intended for training ML models \cite{botifoll_machine_2022}.

\subsection{Rule-Based and Concept-Oriented}
Rule-based semi-supervised approaches represent another category of synthetic TEM image generation, relying on predefined structural concepts and procedures. \cite{govind_deep_2024} proposed a parametric model for generating synthetic training data for dislocation segmentation, demonstrating that synthetic images can yield superior performance compared to models trained solely on real data. \cite{saleh_novel_2025} introduced a concept-oriented approach that integrates edge detection with diffusion models for denoising and connecting broken segments. They use diffusion models primarily for post-processing and noise reduction rather than direct image generation. While rule-based approaches offer precise control over structural features and enable efficient large-scale dataset generation, they remain constrained by the predefined concepts and rules encoded in their algorithms and thereby, like simulations, may fail to capture the full complexity and stochastic variability observed under real experimental conditions.

\subsection{Generative Adversarial Networks}
Recently, Generative Adversarial Networks (GANs) have been introduced as an unsupervised deep-learning approach for synthetic Electron Microscopy image generation. \cite{khan_leveraging_2023} demonstrated the use of cycle-consistent GANs to generate realistic scanning transmission electron microscopy images (STEM), while \cite{eliasson_precise_2025} employed CycleGANs to bridge experimental and simulated images for catalyst nanoparticle analysis with high-angle annular dark-field STEM. In the biological domain, \cite{shagadevan_improved_2021} showed that GANs could effectively augment transmission electron microscopy datasets, improving automated detection performance by generating synthetic labeled images. While GAN-based approaches are capable of generating visually realistic images, they are susceptible to mode collapse and training instabilities, which can limit the diversity and quality of generated samples. Additionally, GANs often face challenges preserving the fine-grained structural details that are critical for quantitative analysis \cite{dhariwal_diffusion_2021b}.

\subsection{Diffusion Models}
The emergence of DDPMs represents a paradigm shift in generative modeling for unsupervised image synthesis. Initially introduced by \cite{ho_denoising_2020b}, DDPMs formulate image generation as a gradual denoising process, in which models are trained to reverse a diffusion procedure that progressively corrupts training images with Gaussian noise. Improvements by \cite{nichol_improved_2021b} enhanced their quality, while the introduction of Denoising Diffusion Implicit Models (DDIM) \cite{song_denoising_2020} addressed the sampling efficiency during inference. Furthermore, \cite{dhariwal_diffusion_2021b} demonstrated that diffusion models achieve superior image quality compared to GANs on standard benchmarks. Diffusion models offer several advantages, including greater training stability, improved mode coverage, preservation of fine-grained structural details, and enhanced generative diversity. Their iterative denoising mechanism also captures noise characteristics and imaging artifacts inherent in real data, making them particularly well-suited for generating synthetic datasets that retain the statistical properties of experimental measurements. To the best of our knowledge, the work presented in this paper is among the first to investigate their application in TEM imaging.

\section{Data}
\label{sec:data}

We use high-resolution TEM images from semiconductor manufacturing research, acquired using actual FAB tools and stored as 3 $\times$ 8 bit grayscale JPEG files. They capture nanometer-scale device structures and potential defects representative of advanced node manufacturing processes.

\begin{figure*}[!t]
  \centering
  \begin{minipage}[t]{0.45\textwidth}
    \centering
    \panel[0.2\linewidth]{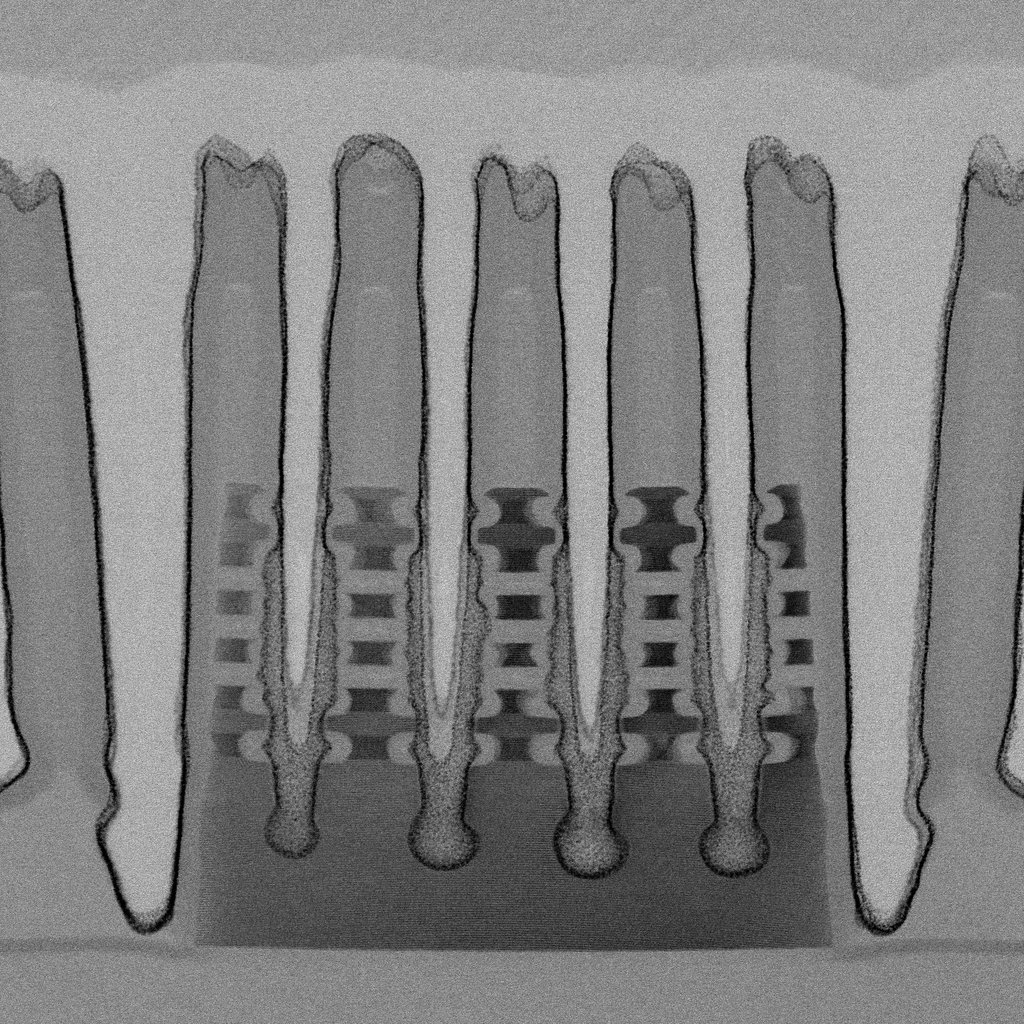}{a}
    \panel[0.2\linewidth]{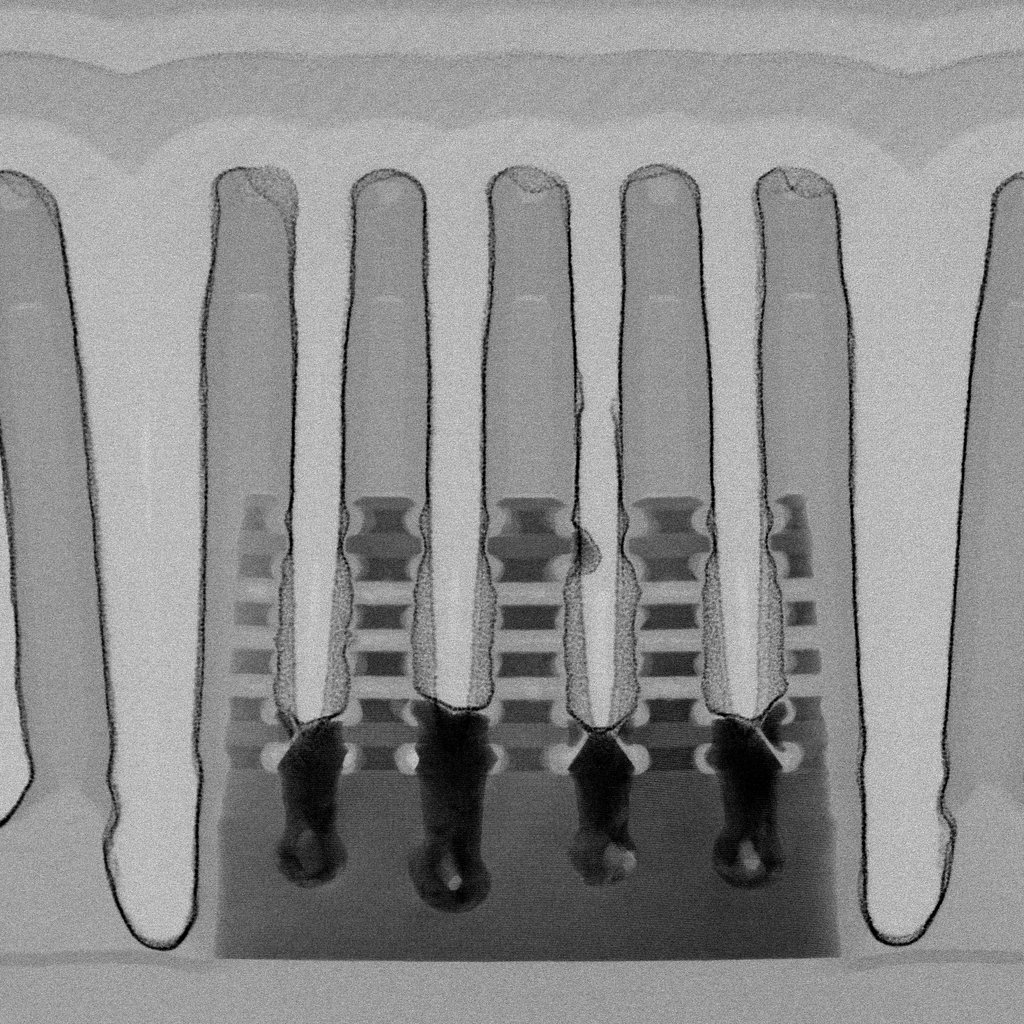}{b}
    \panel[0.2\linewidth]{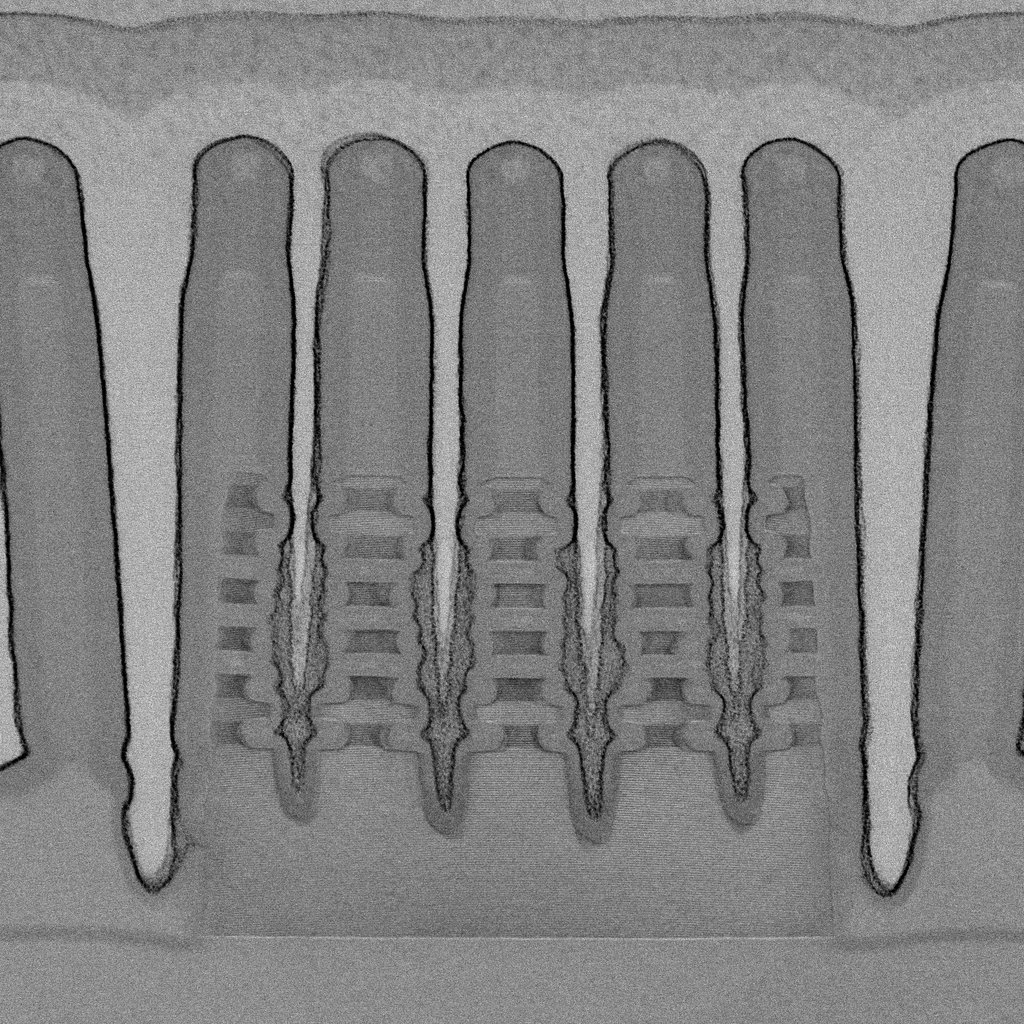}{c}
    \label{fig:device-tem}
  \end{minipage}
  \hfill
  \begin{minipage}[t]{0.45\textwidth}
    \centering
    \panel[0.2\linewidth]{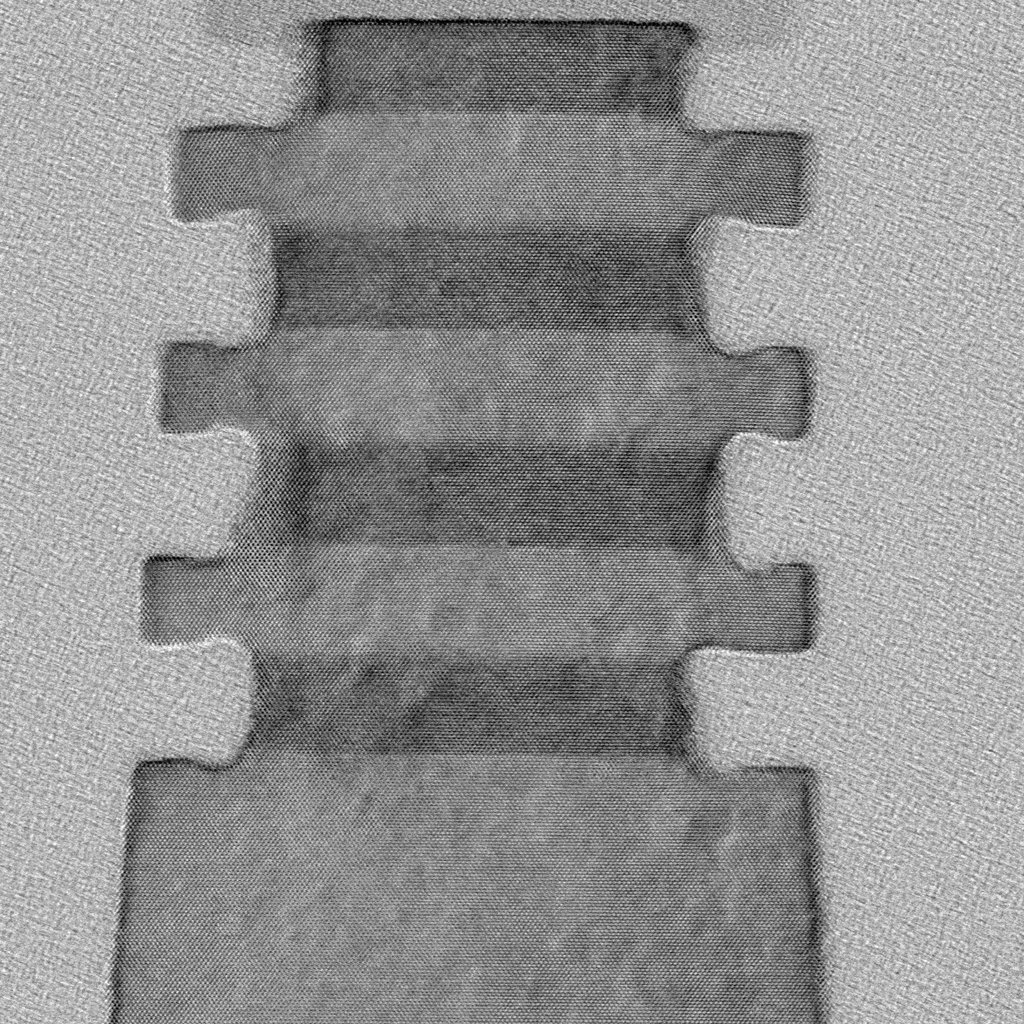}{d}
    \panel[0.2\linewidth]{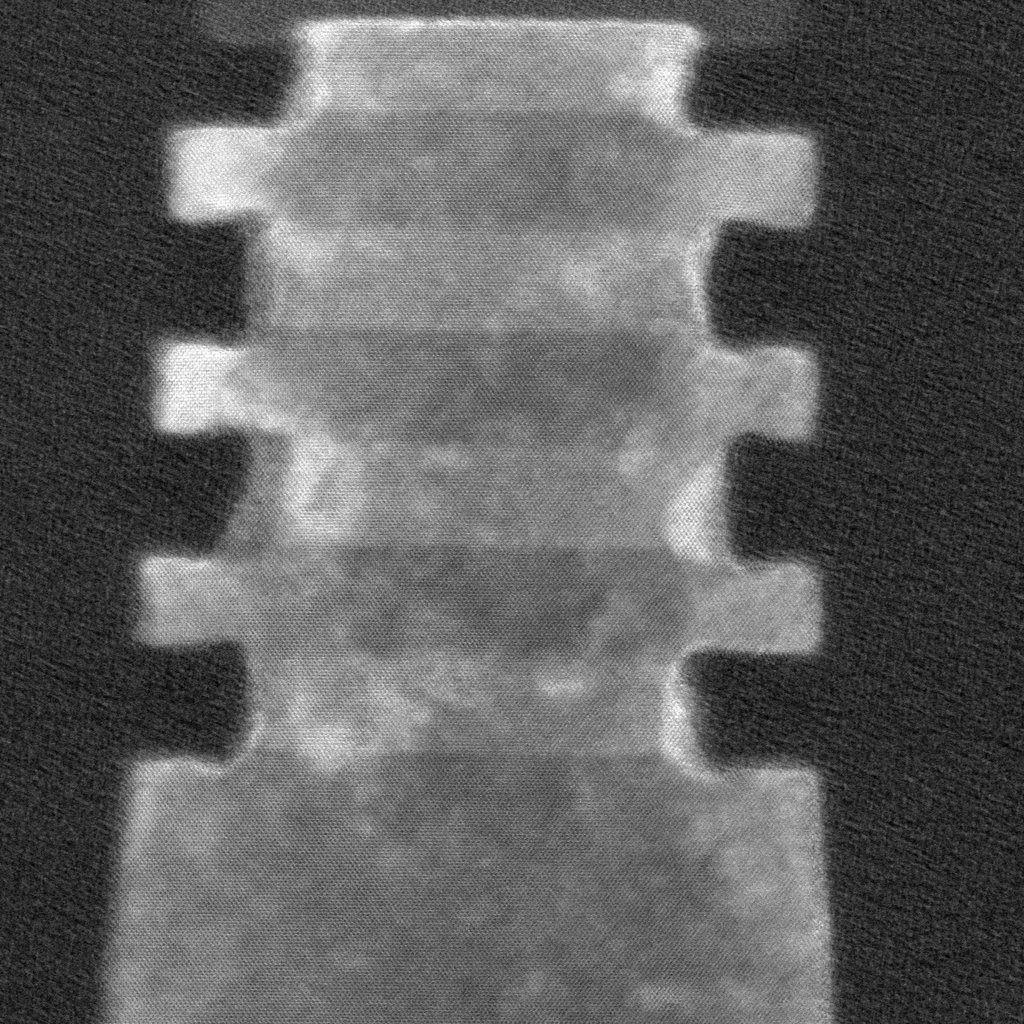}{e}
    \panel[0.2\linewidth]{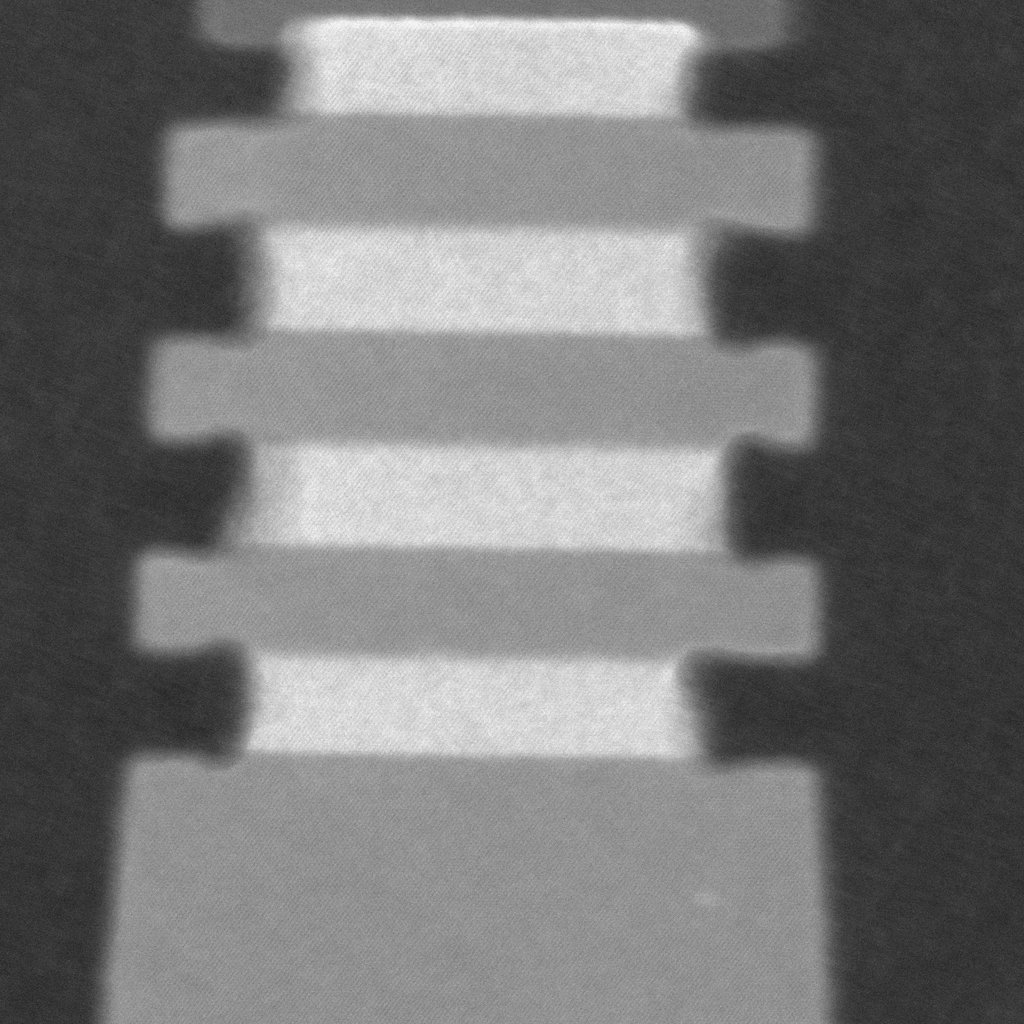}{f}
    \label{fig:nano-tem}
  \end{minipage}
  \caption{(a--c) DEVICE-TEM structures; (d--f) NANO-TEM BF, ADF, HAADF modes.}
  \label{fig:device-nano-row}
\end{figure*}

\textbf{DEVICE-TEM Dataset:} This dataset comprises 15 TEM images, of which 9 with a resolution of ($4096 \times 3874$) pixels and 6 at ($2048 \times 1936$) pixels. These lower-magnification images ($200{,}000\times$ to $300{,}000\times$) capture multiple structural features such as fins, cavities, and pillars, providing broader context for process-related geometries including cavity depth, sidewall profiles, and hard-mask integrity, critical for across-feature metrology and process variation studies. Three different imaging modes are featured: annular bright field (ABF), annular dark-field (ADF) and high-angle annular dark field (HAADF), covering three distinct structures as shown in Fig.~\ref{fig:device-nano-row}~(a-c).

\textbf{NANO-TEM Dataset:} 539 high-magnification ($960{,}000\times$) TEM images with a resolution of ($2048 \times 2048$) pixels, primarily focused on nanosheet and multilayer stacks. These images capture fine structural details at atomic or near-atomic resolution, enabling detailed analysis of layer thickness, interlayer spacing, and crystalline quality. The field of view is limited to a single structure. The dataset features three different imaging modes (BF, ADF, HAADF) across 180 structures, with one missing observation. Fig.~\ref{fig:device-nano-row}~(d-f) provides an overview of the modes featured in this dataset.

\section{Methodology}
\label{sec:methodology}

\subsection{Patch-based Progressive Training}

\begin{figure*}[t]
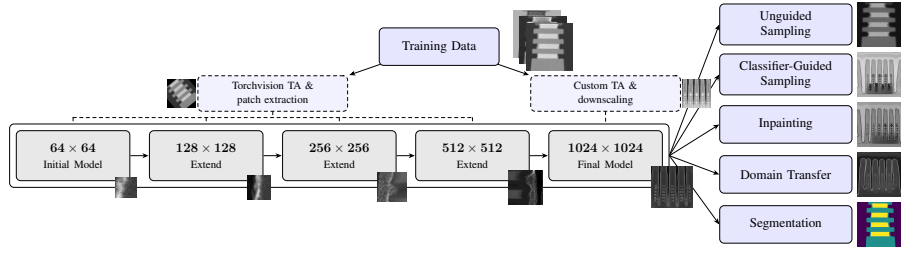

  \centering
  \resizebox{0.65\textwidth}{!}{\includestandalone{figs/framework}}
  \caption{Framework of the proposed patch-based training and generative process.}
  \label{fig:framework}
\end{figure*}

We implement patch-based progressive training as depicted in Fig.~\ref{fig:framework}, systematically increasing both spatial resolution and dataset size, transitioning from fine-grained textures at small scales to global structural features at larger scales. The training progresses through multiple resolution stages of ($64\times64$), ($128\times128$), ($256\times256$), and ($512\times512$) pixel patches, before finally moving to full-FOV images downscaled to ($1024\times1024$) resolution. From each training image, patches are extracted to maximize data utilization while preserving spatial coherence, allowing the integration of details that would otherwise be lost if training began directly on downscaled, full-perspective images. The extraction strategy ensures comprehensive coverage of image content while providing sufficient training samples. Edge regions are upscaled using Lanczos filtering \cite{duchon_lanczos_1979} to preserve texture. We train a separate model for each dataset. 

\subsection{Data Augmentation}
We implement a custom adaptation of the TA algorithm \cite{muller_trivialaugment_2021b}, tailored for semiconductor TEM imaging. TA offers a simple yet effective augmentation strategy by randomly selecting both the augmentation operation and its magnitude for each training sample, thereby avoiding the overhead of hyperparameter tuning. Our implementation incorporates grayscale-specific modifications and rejection criteria for extreme brightness values and hash-based duplicate filtering to ensure structural integrity and maintain control over the distribution of optical characteristics such as brightness and contrast. For the final training stage, we transitioned from the standard TorchVision TA implementation to a customized approach that incorporates augmentation strategies suggested in \cite{chen_improved_2024}, built on Albumentations and OpenCV. This prevents destructive transformations that could compromise the structural integrity of full-resolution images. Tab.~\ref{tab:combined_table} summarizes the dataset sizes after separately applying 20$\times$ augmentation to each stage. The final-stage operation set includes bounded photometric operations and additive Gaussian noise; the latter perturbs the noise statistics analyzed in Sec.~\ref{subsec:noise_device} and Sec.~\ref{subsec:noise_nano}, and is applied identically to all models trained for comparison.

\begin{table}[t]
\centering
\caption{Per-stage dataset sizes after 20$\times$ augmentation}
\label{tab:combined_table}
\footnotesize
\setlength{\tabcolsep}{4pt}
\renewcommand{\arraystretch}{1.12}
\begin{tabular}{lrr}
\toprule
\textbf{Resolution} & \textbf{DEVICE-TEM} & \textbf{NANO-TEM} \\
\midrule
$64\times64$     & 877{,}307  & 11{,}590{,}656 \\
$128\times128$   & 219{,}327  & 2{,}897{,}664 \\
$256\times256$   & 54{,}832   & 724{,}416 \\
$512\times512$   & 13{,}708   & 181{,}104 \\
$1024\times1024$ & 315        & 11{,}319 \\
\bottomrule
\end{tabular}
\end{table}

\subsection{Diffusion Model Architecture and Training}

\subsubsection{Model Architecture}
We employ a DDPM \cite{nichol_improved_2021b} in a U-Net configuration \cite{ronneberger_unet_2015}, implemented in PyTorch \cite{pytorch}. The model is constructed with increasing depth to enable weight transfer across resolution stages without architectural modifications. Channel multipliers and residual block configurations accommodate the progressive resolution increase from $64 \times 64$ to $1024 \times 1024$. Attention mechanisms at multiple resolution levels capture local texture patterns and global structural information.

\subsubsection{Hyperparameters and Training Configuration}
Patch-based training ran for 10 epochs per patch size, with batch-sizes increasing fourfold at each smaller resolution stage. Full-scale training at $1024 \times 1024$ required at least 40\,GB VRAM at batch-size 1 (DEVICE-TEM); NANO-TEM used an NVIDIA A100 80\,GB at batch-size 2. Settings: constant learning rate $3\times 10^{-5}$, gradient accumulation over 16 batches, dropout 0.2, maximum gradient norm 0.5, noise offset 0.05 (0.01 at final stage), minimum SNR gamma 5, cosine beta schedule, v-prediction parameterization, and EMA updates every 10 iterations with decay 0.995. DEVICE-TEM converged after 122{,}500 iterations at the final $1024 \times 1024$ stage (283.56\,h at 0.12\,it/s; 357.12\,h across all stages); NANO-TEM after 220{,}000 final-stage iterations (679.01\,h at 0.09\,it/s; 869.53\,h across all stages). The disproportionate cost of the last stage reflects the complexity of high-resolution features. Inference on an A100 40\,GB took 0.16\,s per DDIM timestep \cite{song_denoising_2020} at batch-size 1, scaling to batch-size 13 with \texttt{torch.compile}.

\subsubsection{Progressive Training and Weight Transfer}
For each resolution stage beyond the initial ($64 \times 64$), we froze encoder layers for half of the training iterations, except for the final NANO-TEM stage at $1024 \times 1024$, where freezing was limited to 60k of 220k iterations ($27\%$) because convergence was not known a priori. This allowed the model to adapt decoder layers to the new resolution while preserving low-level features learned in earlier stages, improving knowledge transfer and training stability. Higher-resolution models were initialized with the complete set of weights from the preceding stage, ensuring architectural and parameter compatibility across all resolution levels. Structural and perceptual metrics favor the proposed approach on MS-SSIM, SSIM and LPIPS, although each difference lies within one per-sample standard deviation, each configuration was trained only once, and FID is marginally lower without freezing and lowest for the non-patch-based baseline (Appendix~\ref{sec:ablation_patches}).

\subsection{Prediction Objective and Loss Weighting}
\label{subsec:objective}
Rather than predicting the noise $\boldsymbol{\epsilon}$ directly as in the original DDPM formulation \cite{ho_denoising_2020b}, the model is trained with the $v$-prediction parameterization introduced by \cite{salimans_progressive_2021}. In this formulation, the network $v_\theta$ predicts the velocity:
\begin{equation}
    \mathbf{v}_t = \sqrt{\bar{\alpha}_t}\,\boldsymbol{\epsilon} - \sqrt{1 - \bar{\alpha}_t}\,\mathbf{x}_0
    \label{eq:v_prediction}
\end{equation}
where $\mathbf{x}_0$ is the original TEM image, $\boldsymbol{\epsilon} \sim \mathcal{N}(0, I)$ is the noise added during the forward process, and $\bar{\alpha}_t$ is the cumulative product of the noise schedule at timestep $t$. The training objective minimizes the mean squared error between predicted and actual velocity:
\begin{equation}
    \mathcal{L} = \mathbb{E}_{t, \mathbf{x}_0, \boldsymbol{\epsilon} \sim \mathcal{N}(0,I)} \left[ w(t)\, \|\mathbf{v}_t - v_\theta(\mathbf{x}_t, t)\|^2 \right]
    \label{eq:v_loss}
\end{equation}
The $v$-prediction objective improves training stability and sample quality compared to $\boldsymbol{\epsilon}$-prediction \cite{salimans_progressive_2021}, particularly when combined with accelerated sampling schedules such as DDIM \cite{song_denoising_2020}. The per-timestep weight $w(t)$ is set following the min-SNR-$\gamma$ scheme \cite{hang_efficient_2023} with $\gamma = 5$, which down-weights low-noise timesteps that would otherwise dominate the loss and balances the contributions across the diffusion trajectory.

While the prediction objective remains consistent across all stages, the progressive training strategy facilitates hierarchical feature learning by progressively increasing data complexity and model scaling. At the ($64\times64$) resolution stage, training emphasizes image characteristics such as contrast, noise patterns, and basic textures. As the resolution increases, the model builds on these representations while incorporating larger-scale structural information, including boundaries and spatial relationships. Weight transfer and layer freezing ensure that fine-grained texture representations learned at lower resolutions are preserved and further refined.

\subsection{Structural Similarity Metrics and Methodology}
\label{sec:similarity}

\paragraph{Metrics}
To evaluate the structural similarity of original and synthetic images, we adopt domain-specific metrics proposed by \cite{treder_applications_2022} and implemented via TorchMetrics \cite{detlefsen_torchmetrics_2022}. As our augmentation algorithm includes operations on brightness and contrast \cite{chen_improved_2024}, we also employ the \textbf{Multi-Scale Structural Similarity Index Measure (MS-SSIM)} \cite{wang_multiscale_2003}, improving robustness by computing the \textbf{Structural Similarity Index Measure (SSIM)} \cite{wang_image_2004} across multiple scales. We measure perceptual similarity by employing the \textbf{Learned Perceptual Image Patch Similarity (LPIPS)} metric proposed by \cite{zhang_unreasonable_2018a} using the default AlexNet backbone. To quantify the distance between the distributions of real and synthetic images, we include the \textbf{Fréchet Inception Distance (FID)} \cite{heusel_gans_2017} and the \textbf{Kernel Inception Distance (KID)} \cite{binkowski_demystifying_2018}. FID uses all samples; KID is averaged over 100 random subsets of size $\min(1000, N_{\text{syn}}, N_{\text{real}})$. With $N_{\text{real}} = 15$ for DEVICE-TEM and $539$ for NANO-TEM, each subset contains all real images, so the reported KID standard deviation reflects only the sampling of synthetic images. At this sample size the FID covariance estimate for DEVICE-TEM is rank-deficient; we therefore read the DEVICE-TEM FID as indicative only and base distribution-level conclusions primarily on KID.

\paragraph{Methodology}
We evaluate structural and perceptual similarity via per-metric nearest-neighbor matching on the full set of generated samples (DEVICE-TEM $n=3976$, NANO-TEM $n=6656$). To prevent cross-mode pairings, candidate real images are restricted to those within the same acquisition mode, identified via Euclidean distance on z-score-normalized proxy features (brightness, pixel std, Laplacian-based noise estimate). 250 DDIM timesteps and an $\eta$-value of 0.5 were used, allowing for stochasticity in the sampling process \cite{song_denoising_2020}. Fig.~\ref{fig:ms-ssim-vs-ddim-side-by-side}~(b) shows the similarity metrics saturating well before 250 timesteps, so fewer timesteps would reduce sampling cost at a comparable similarity level. We further checked the robustness of the nearest-neighbor analysis with respect to the top-$k$ cutoff (results not shown). While the quantitative discussion below reports the top-$1\%$ subset, varying $k$ did not change the relative trends, and all metrics remained within the reported ranges.

\paragraph{Baseline}
For baseline comparisons, we employ MS-SSIM VAE \cite{snell_learning_2017b}, DCGAN \cite{radford_unsupervised_2016} and DDPM \cite{ho_denoising_2020b} architectures trained from scratch on full-FOV images using our augmentation and sampling strategies.  We provide a comprehensive quantitative and qualitative assessment of similarity based on: (1) pixel-level: MS-SSIM, SSIM, PSNR, MSE \cite{snell_learning_2017b, treder_applications_2022}; (2) feature/perceptual level: LPIPS \cite{zhang_unreasonable_2018a}; (3) distribution level: FID, KID \cite{heusel_gans_2017, binkowski_demystifying_2018}; (4) task-specific: noise distribution matching.

\section{Results}
\label{sec:results}

\subsection{Structural Similarity (DEVICE-TEM)}

\begin{figure}[t]
  \centering
  \setlength{\tabcolsep}{1.5pt}
  \begin{tabular}{ccccc}
    \panel[0.18\linewidth]{plots/device-tem1.jpg}{a} &
    \panel[0.18\linewidth]{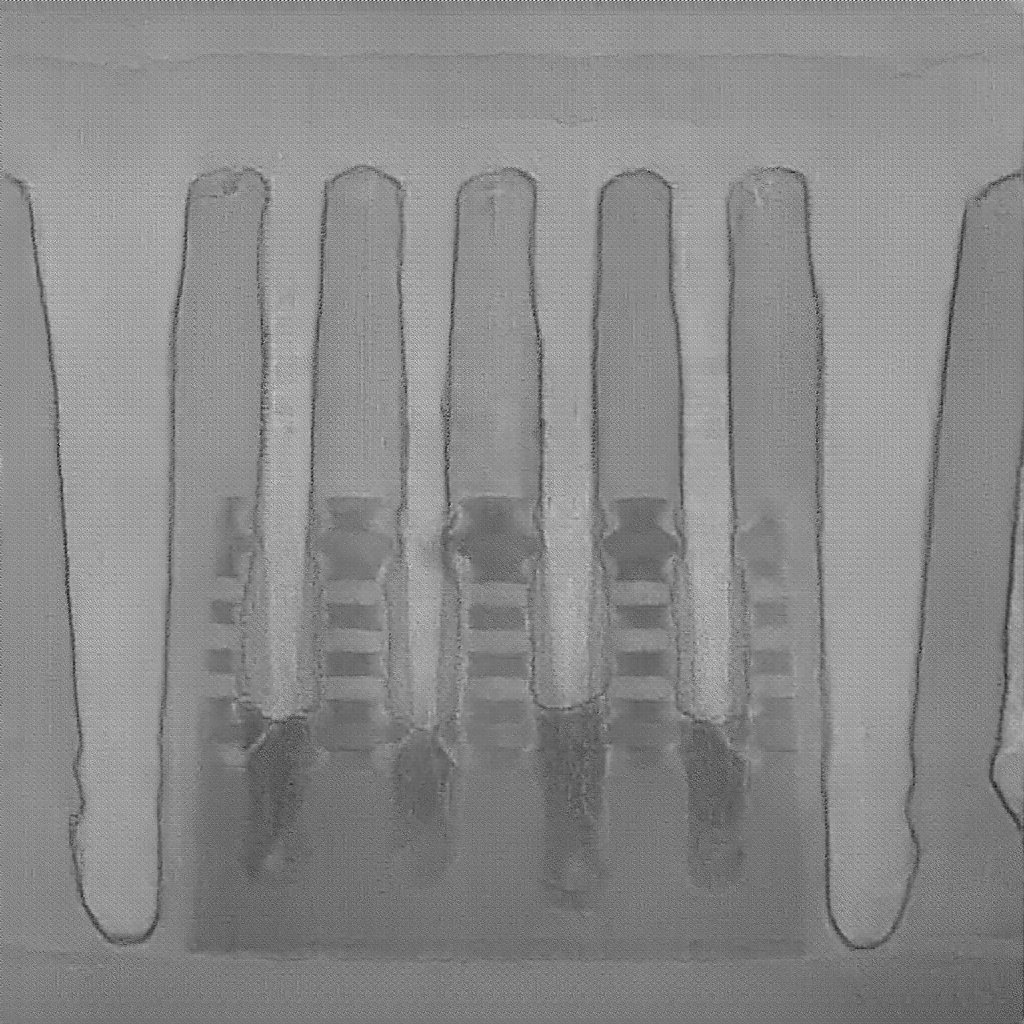}{b} &
    \panel[0.18\linewidth]{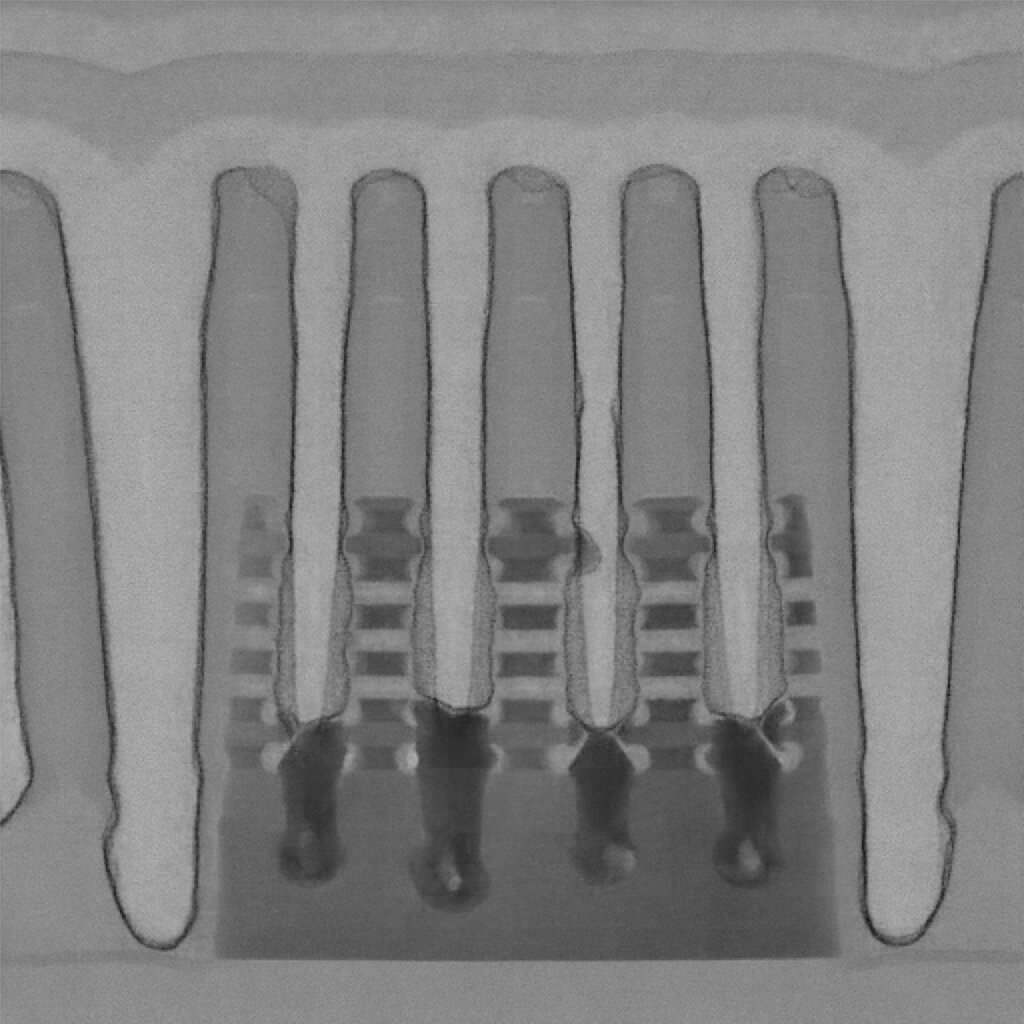}{c} &
    \panel[0.18\linewidth]{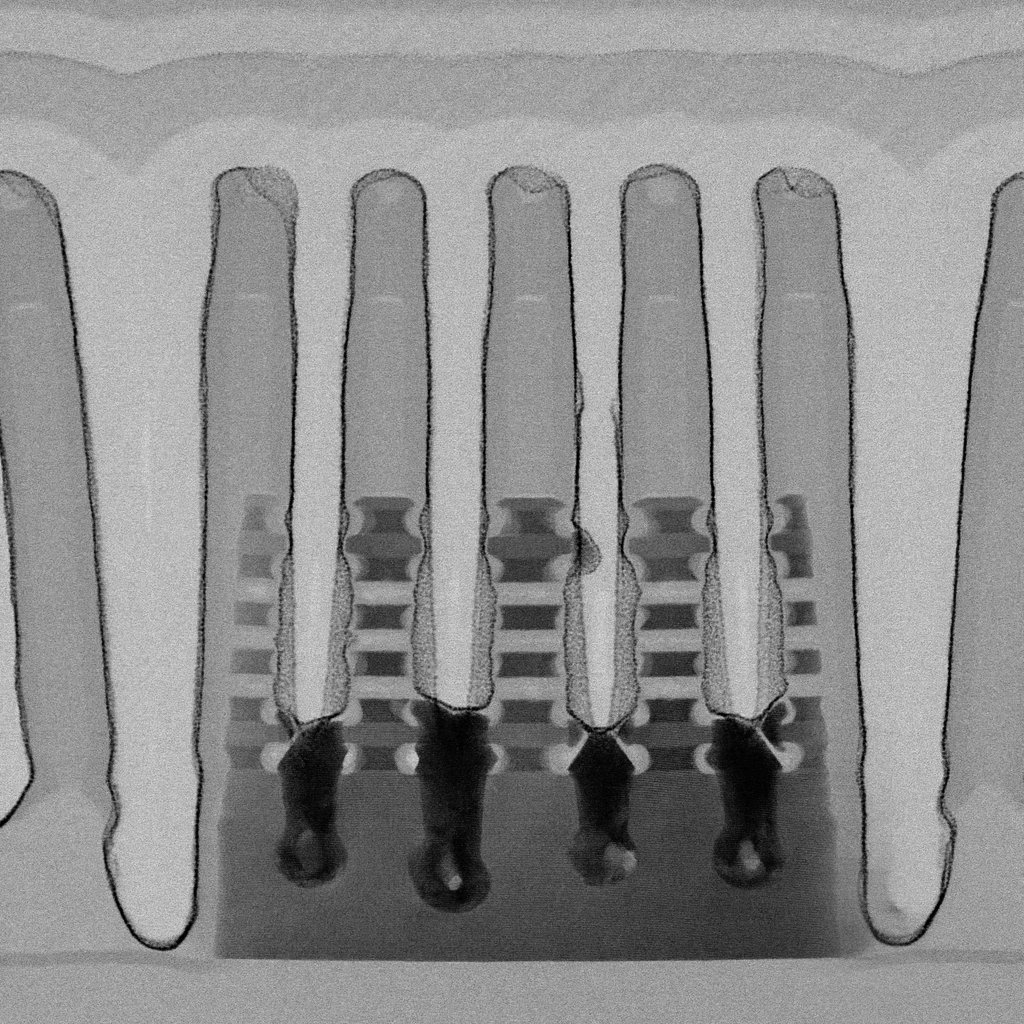}{d} &
    \panel[0.18\linewidth]{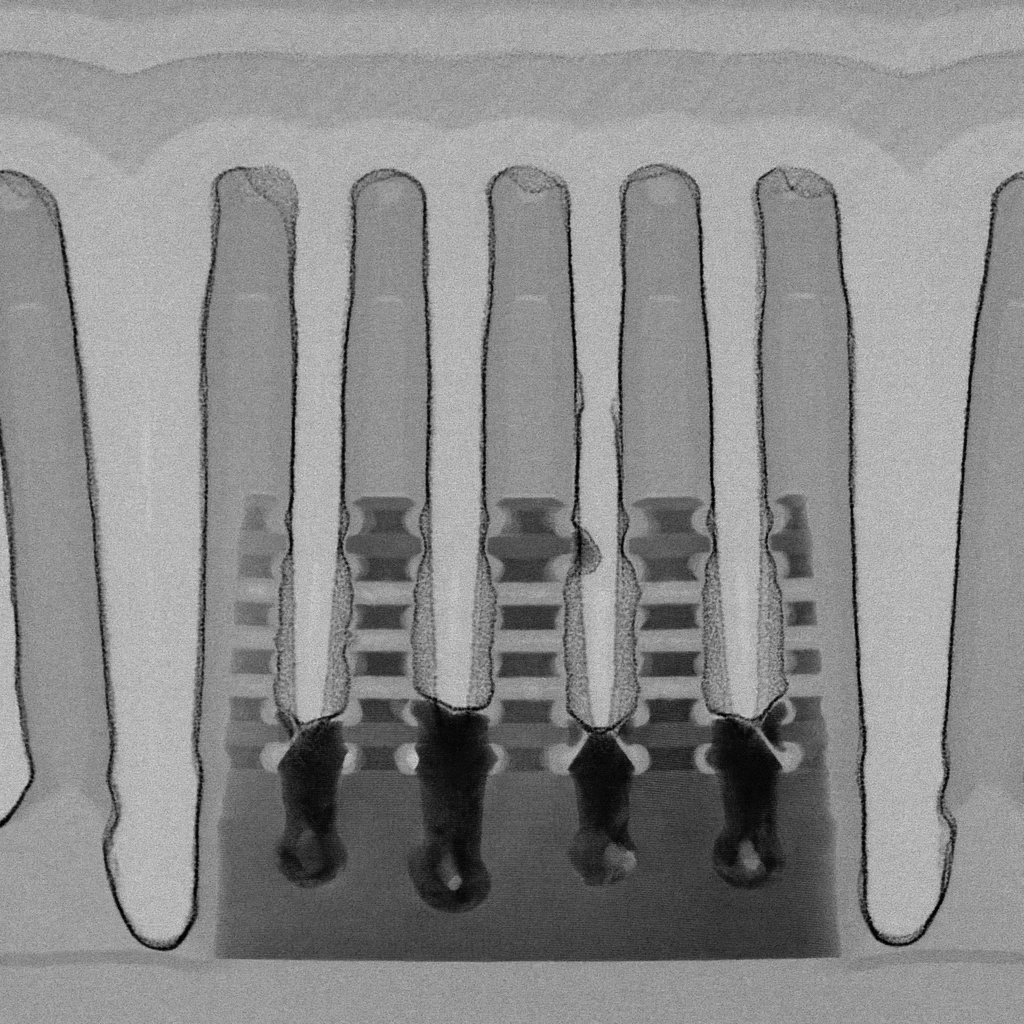}{e}
  \end{tabular}
  \caption{Original (a), DCGAN (b), MS-SSIM VAE (c), baseline DDPM (d) and proposed generative approach (e).}
  \label{fig:model-comparison}
\end{figure}

\FloatBarrier

\begin{table}[t]
\caption{Structural similarity metrics for synthetic DEVICE-TEM images (mean~$\pm$~std and median~[IQR], $n=3976$).}
\label{tab:similarity_metrics_device}
\centering
\scriptsize
\setlength{\tabcolsep}{1.5pt}
\renewcommand{\arraystretch}{1.12}
\resizebox{\columnwidth}{!}{%
\begin{tabular}{llrrrr}
\toprule
\textbf{Metric} & \textbf{Stat} & \textbf{Ours} & \textbf{Base} & \textbf{VAE} & \textbf{DCGAN} \\
\midrule
\multirow{2}{*}{MS-SSIM $\uparrow$}
  & mean   & \bm{$0.587{\pm}0.210$} & $0.547{\pm}0.196$ & $\underline{0.571{\pm}0.172}$ & $0.309{\pm}0.084$ \\
  & med.   & \bm{$0.546\,[0.348]$}  & $0.512\,[0.296]$  & $\underline{0.535\,[0.270]}$  & $0.265\,[0.089]$  \\
\multirow{2}{*}{SSIM $\uparrow$}
  & mean   & $\underline{0.529{\pm}0.191}$ & $0.480{\pm}0.178$ & \bm{$0.574{\pm}0.102$} & $0.208{\pm}0.095$ \\
  & med.   & $\underline{0.569\,[0.292]}$  & $0.490\,[0.278]$  & \bm{$0.573\,[0.127]$}  & $0.166\,[0.092]$  \\
\multirow{2}{*}{PSNR $\uparrow$}
  & mean   & \bm{$18.96{\pm}4.57$} & $\underline{18.13{\pm}3.63}$ & $15.87{\pm}3.07$ & $14.17{\pm}2.37$ \\
  & med.   & \bm{$18.00\,[3.89]$}  & $\underline{17.47\,[3.26]}$  & $15.61\,[4.14]$  & $12.82\,[4.46]$  \\
\multirow{2}{*}{MSE $\downarrow$}
  & mean   & \bm{$0.019{\pm}0.018$} & $\underline{0.020{\pm}0.016}$ & $0.032{\pm}0.019$ & $0.044{\pm}0.020$ \\
  & med.   & \bm{$0.016\,[0.014]$}  & $\underline{0.018\,[0.013]}$  & $0.028\,[0.028]$  & $0.052\,[0.037]$  \\
\multirow{2}{*}{LPIPS $\downarrow$}
  & mean   & \bm{$0.342{\pm}0.173$} & $\underline{0.377{\pm}0.159}$ & $0.492{\pm}0.115$ & $0.618{\pm}0.128$ \\
  & med.   & \bm{$0.349\,[0.292]$}  & $\underline{0.381\,[0.245]}$  & $0.524\,[0.138]$  & $0.698\,[0.187]$  \\
\midrule
FID $\downarrow$   & & $\underline{58.73}$ & \bm{$55.43$} & $202.72$ & $251.36$ \\
KID $\to 0$        & & \bm{$-0.002{\pm}0.013$} & $\underline{-0.010{\pm}0.010}$ & $0.094{\pm}0.027$ & $0.132{\pm}0.043$ \\
\bottomrule
\end{tabular}}
\end{table}

\begin{figure}[!b]
  \centering
  \begin{tikzpicture}
    \node[inner sep=0, anchor=south west] (img1)
      {\includegraphics[width=0.42\linewidth]{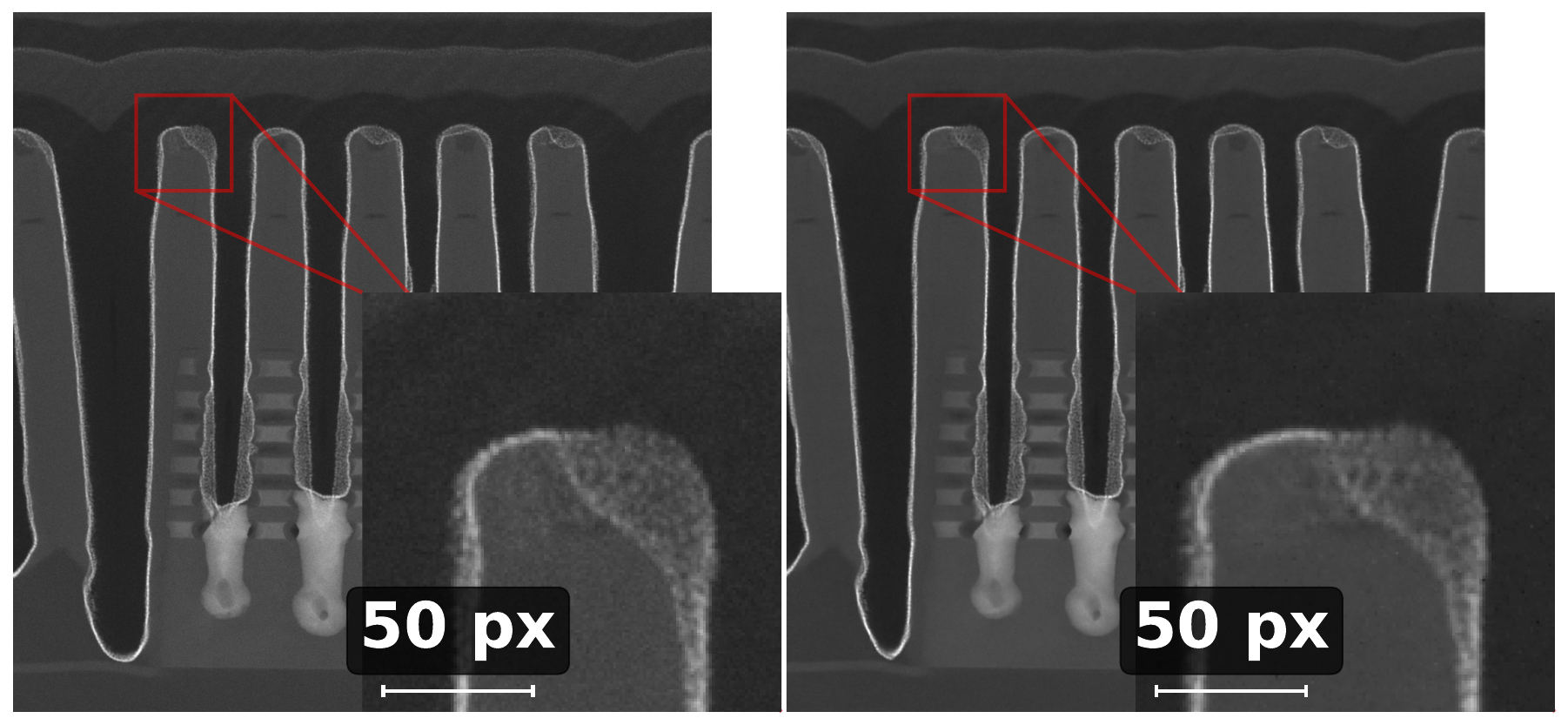}};
    \node[inner sep=0, anchor=south west] (img2) at ([xshift=4pt]img1.south east)
      {\includegraphics[width=0.42\linewidth]{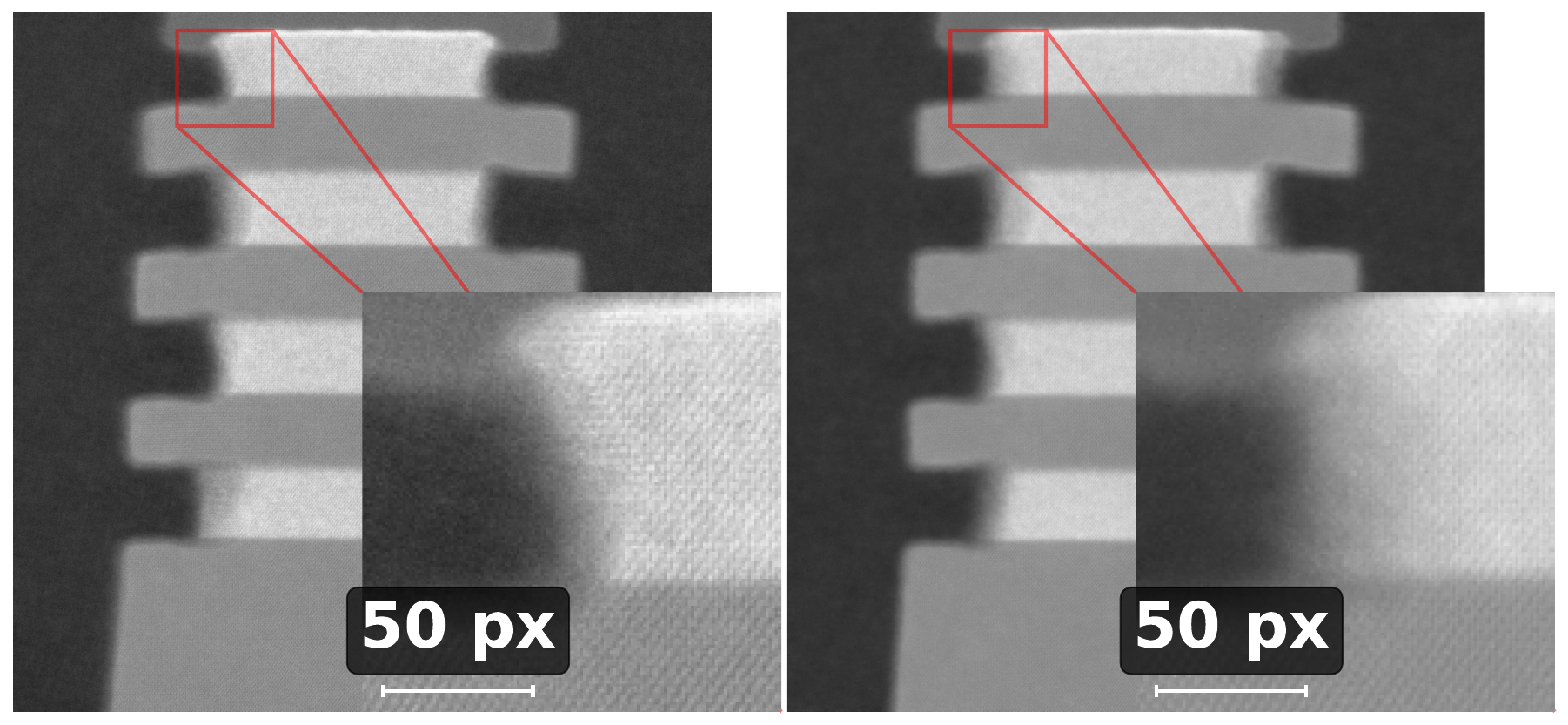}};

    \node[anchor=north west] at ([xshift=-2pt,yshift=2pt]img1.north west) {\paneltag{a}};
    \node[anchor=north west] at ([xshift=-3pt,yshift=2pt]$(img1.north west)!0.5!(img1.north east)$) {\paneltag{b}};

    \node[anchor=north west] at ([xshift=-2pt,yshift=2pt]img2.north west) {\paneltag{c}};
    \node[anchor=north west] at ([xshift=-3pt,yshift=2pt]$(img2.north west)!0.5!(img2.north east)$) {\paneltag{d}};
  \end{tikzpicture}
  \caption{Original and synthetic images: DEVICE-TEM (a, b) and NANO-TEM (c, d).}
  \label{nano_device_image_comparison}
\end{figure}
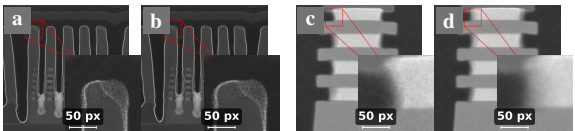

\begin{figure*}[t]
  \centering
  \begin{tikzpicture}
    \node[inner sep=0, anchor=south west] (img1)
      {\includegraphics[width=0.35\textwidth]{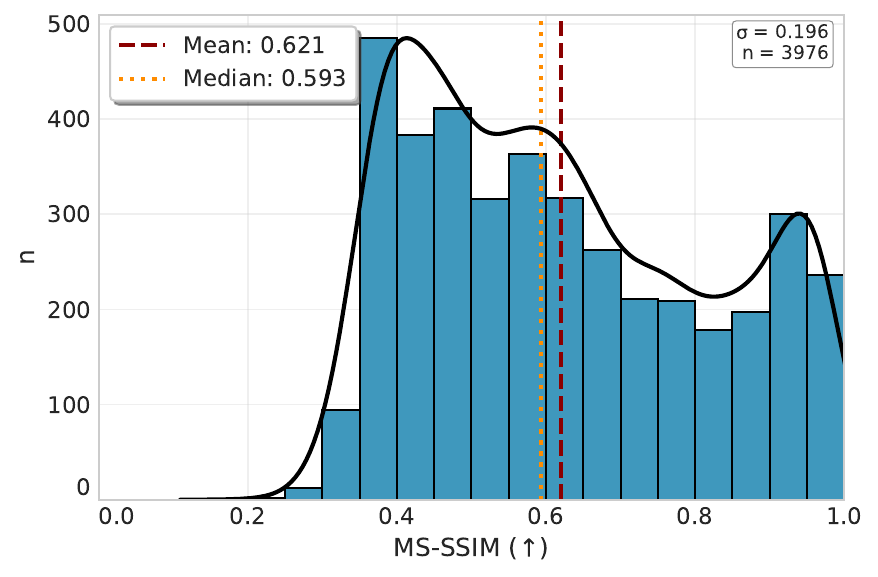}};
    \node[inner sep=0, anchor=south west] (img2) at ([xshift=16pt]img1.south east)
      {\includegraphics[width=0.35\textwidth]{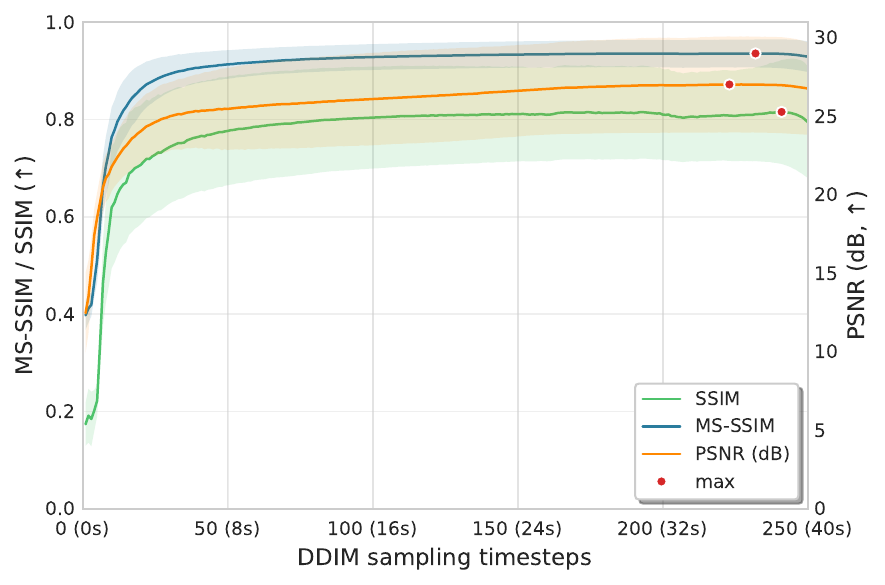}};

    \node[anchor=west] at ([xshift=-161pt, yshift=0.5pt]img1.east) {\paneltag{a}};
    \node[anchor=west] at ([xshift=-32.5pt]img2.east) {\paneltag{b}};
  \end{tikzpicture}
  \caption{(a) Distribution of nearest-neighbor DEVICE-TEM MS-SSIM ($n=3976$) matched against all original images, i.e. without the acquisition-mode restriction of Sec.~\ref{sec:similarity} (mode-restricted statistics in Tab.~\ref{tab:similarity_metrics_device}); (b) similarity (mean $\pm$ std) vs. timesteps ($n=100$).}
  \label{fig:ms-ssim-vs-ddim-side-by-side}
\end{figure*}

Tab.~\ref{tab:similarity_metrics_device} summarizes our similarity analysis; Fig.~\ref{fig:ms-ssim-vs-ddim-side-by-side}~(a) shows the MS-SSIM distribution when matching against all original images. On the top 1\% nearest-neighbor pairs ($n=40$), MS-SSIM rises to $0.985 \pm 0.002$ and SSIM to $0.918 \pm 0.005$, clearly ahead of the DDPM baseline ($0.977 / 0.881$), VAE ($0.955 / 0.867$), and DCGAN ($0.460 / 0.383$). PSNR is inherently bounded by the stochasticity of the reverse diffusion process, JPEG compression artifacts in the training data, and the sensitivity of PSNR to Gaussian noise variations \cite{hore_image_2010}. On the top-$1\%$ subset our approach leads on MS-SSIM and SSIM; across the full sample set it leads on all metrics except SSIM, where the VAE scores higher, and FID, where the baseline DDPM is marginally lower. KID, which is more reliable in the small-sample regime \cite{binkowski_demystifying_2018}, does not separate the two DDPM variants ($-0.002 \pm 0.013$ vs.\ $-0.010 \pm 0.010$). The DCGAN exhibits clear signs of mode collapse. The best synthetic samples closely resemble the original images while retaining subtle variations rather than being pixel-wise replicas, see Fig.~\ref{nano_device_image_comparison}.

\subsection{Structural Similarity (NANO-TEM)}

\begin{table}[t]
\caption{Structural similarity metrics for synthetic NANO-TEM images (mean~$\pm$~std and median~[IQR], $n=6656$).}
\label{tab:similarity_metrics_nano}
\centering
\scriptsize
\setlength{\tabcolsep}{1.5pt}
\renewcommand{\arraystretch}{1.12}
\resizebox{\columnwidth}{!}{%
\begin{tabular}{llrrrr}
\toprule
\textbf{Metric} & \textbf{Stat} & \textbf{Ours} & \textbf{Base} & \textbf{VAE} & \textbf{DCGAN} \\
\midrule
\multirow{2}{*}{MS-SSIM $\uparrow$}
  & mean   & $0.481{\pm}0.134$ & $0.412{\pm}0.099$ & \bm{$0.579{\pm}0.133$} & $\underline{0.508{\pm}0.131}$ \\
  & med.   & $0.414\,[0.202]$  & $0.377\,[0.080]$  & \bm{$0.587\,[0.217]$}  & $\underline{0.472\,[0.271]}$  \\
\multirow{2}{*}{SSIM $\uparrow$}
  & mean   & $0.270{\pm}0.129$ & $0.199{\pm}0.091$ & \bm{$0.397{\pm}0.126$} & $\underline{0.276{\pm}0.098}$ \\
  & med.   & $0.211\,[0.179]$  & $0.169\,[0.072]$  & \bm{$0.407\,[0.192]$}  & $\underline{0.268\,[0.206]}$  \\
\multirow{2}{*}{PSNR $\uparrow$}
  & mean   & $16.99{\pm}1.85$ & $16.34{\pm}1.55$ & \bm{$18.63{\pm}1.95$} & $\underline{18.15{\pm}1.98}$ \\
  & med.   & $16.89\,[1.89]$  & $16.42\,[1.62]$  & $\underline{18.45\,[2.51]}$  & \bm{$18.67\,[1.86]$}  \\
\multirow{2}{*}{MSE $\downarrow$}
  & mean   & $0.022{\pm}0.009$ & $0.025{\pm}0.010$ & \bm{$0.015{\pm}0.007$} & $\underline{0.018{\pm}0.015}$ \\
  & med.   & $0.021\,[0.009]$  & $0.023\,[0.009]$  & $\underline{0.014\,[0.008]}$  & \bm{$0.014\,[0.006]$}  \\
\multirow{2}{*}{LPIPS $\downarrow$}
  & mean   & $\underline{0.277{\pm}0.063}$ & \bm{$0.229{\pm}0.057$} & $0.417{\pm}0.086$ & $0.295{\pm}0.032$ \\
  & med.   & $\underline{0.266\,[0.079]}$  & \bm{$0.217\,[0.072]$}  & $0.409\,[0.122]$  & $0.305\,[0.050]$  \\
\midrule
FID $\downarrow$   & & \bm{$47.32$} & $\underline{68.52}$ & $170.96$ & $144.67$ \\
KID $\to 0$        & & \bm{$0.034{\pm}0.004$} & $\underline{0.067{\pm}0.005}$ & $0.147{\pm}0.007$ & $0.093{\pm}0.003$ \\
\bottomrule
\end{tabular}}
\end{table}

For NANO-TEM, the synthesis task is more challenging due to atomic-scale structures and stronger high-frequency content. Tab.~\ref{tab:similarity_metrics_nano} reports our results on 6656 generated samples. Our model achieves the best distribution-level FID and KID; the MS-SSIM VAE attains the highest pixel- and structure-level scores but the worst perceptual and distribution-level metrics, consistent with mode collapse inflating its similarity scores. Both DDPM variants outperform VAE and DCGAN on LPIPS, FID, and KID. On the top 1\% best-case samples ($n=67$), our model reaches MS-SSIM $0.813 \pm 0.013$ and SSIM $0.602 \pm 0.012$, surpassing the baseline DDPM ($0.774 / 0.560$) and DCGAN ($0.738 / 0.455$); the VAE leads here ($0.844 / 0.650$) but benefits from mode collapse. The gap to DEVICE-TEM indicates that reproducing atomic-scale periodicity remains more difficult than larger device-level structures.

\subsection{Perceptual Similarity (DEVICE-TEM)}
Fifteen original and fifteen synthetic ($1024 \times 1024$) DEVICE-TEM images were shown in mixed order to five domain experts, blinded to image origin and without author contact; none consistently identified the synthetic samples. A pre-registered forced-choice study is left for future work.

\subsection{Noise Characteristics (DEVICE-TEM)}
\label{subsec:noise_device}
We computed three noise metrics across the full set of generated samples described in Sec.~\ref{sec:similarity}. Tab.~\ref{tab:noise_comparison} summarizes our results for DEVICE-TEM. Unlike the structural similarity metrics in Tab.~\ref{tab:similarity_metrics_device}, noise metrics are reported at the dataset level; individual best-matched pairs may exhibit closer agreement. Our DDPM yields a slightly reduced noise standard deviation ($0.018 \pm 0.009$ vs. $0.022 \pm 0.011$) and modestly increased SNR ($25.53 \pm 3.58$\,dB vs. $23.78 \pm 2.66$\,dB), while the HFNR, a relative measure of spectral concentration computed identically for all models, remains closely matched ($1.139 \pm 0.018$ vs. $1.141 \pm 0.031$). We use HFNR only for comparison between models; its absolute value is set by the frequency partition of Eq.~(\ref{eq:hfnr}) and is not an absolute high-frequency fraction. The baseline DDPM performs comparably (noise std $0.019 \pm 0.011$, SNR $26.08 \pm 3.12$\,dB, HFNR $1.138 \pm 0.021$), suggesting that diffusion-based synthesis reproduces key noise characteristics while mildly smoothing the output. The MS-SSIM VAE produces over-smoothed images with markedly reduced noise standard deviation ($0.009 \pm 0.002$) and elevated SNR ($34.40 \pm 1.59$\,dB), consistent with a reconstruction-oriented loss that suppresses high-frequency components. Its HFNR ($1.209 \pm 0.006$) lies furthest from the original of all models, indicating the largest deviation in spectral concentration. DCGAN outputs appear noisier ($0.052 \pm 0.021$) with degraded SNR ($20.42 \pm 3.60$\,dB), and its HFNR ($1.090 \pm 0.010$) deviates further from the original than that of either DDPM variant, indicating differences in both amplitude and spectral concentration.

\begin{table}[t]
\caption{Dataset-level noise metrics for synthetic DEVICE-TEM images (mean~$\pm$~std and median~[IQR], $n=3976$); bold: closest to the originals, underlined: second closest.}
\label{tab:noise_comparison}
\centering
\scriptsize
\setlength{\tabcolsep}{1.5pt}
\renewcommand{\arraystretch}{1.12}
\resizebox{\columnwidth}{!}{%
\begin{tabular}{llrrrrr}
\toprule
\textbf{Metric} & \textbf{Stat} & \textbf{Orig.} & \textbf{Ours} & \textbf{Base} & \textbf{VAE} & \textbf{DCGAN} \\
\midrule
\multirow{2}{*}{Noise std}
  & mean   & $0.022{\pm}0.011$ & $\underline{0.018{\pm}0.009}$ & \bm{$0.019{\pm}0.011$} & $0.009{\pm}0.002$ & $0.052{\pm}0.021$ \\
  & med.   & $0.016\,[0.019]$  & \bm{$0.018\,[0.014]$}  & $\underline{0.018\,[0.015]}$  & $0.008\,[0.002]$  & $0.050\,[0.040]$  \\
\midrule
\multirow{2}{*}{SNR}
  & mean   & $23.78{\pm}2.66$ & \bm{$25.53{\pm}3.58$} & $\underline{26.08{\pm}3.12}$ & $34.40{\pm}1.59$ & $20.42{\pm}3.60$ \\
  & med.   & $23.58\,[2.25]$  & \bm{$25.87\,[4.05]$}  & $\underline{26.35\,[2.77]}$  & $34.73\,[2.08]$  & $18.87\,[7.01]$  \\
\midrule
\multirow{2}{*}{HFNR}
  & mean   & $1.141{\pm}0.031$ & \bm{$1.139{\pm}0.018$} & $\underline{1.138{\pm}0.021}$ & $1.209{\pm}0.006$ & $1.090{\pm}0.010$ \\
  & med.   & $1.155\,[0.053]$  & \bm{$1.134\,[0.027]$}  & $\underline{1.128\,[0.040]}$  & $1.210\,[0.007]$  & $1.085\,[0.019]$  \\
\bottomrule
\end{tabular}}
\end{table}

\subsection{Noise Characteristics (NANO-TEM)}
\label{subsec:noise_nano}

\begin{table}[t]
\caption{Dataset-level noise metrics for synthetic NANO-TEM images (mean~$\pm$~std and median~[IQR], $n=6656$); bold: closest to the originals, underlined: second closest.}
\label{tab:noise_comparison_nano}
\centering
\scriptsize
\setlength{\tabcolsep}{1.5pt}
\renewcommand{\arraystretch}{1.12}
\resizebox{\columnwidth}{!}{%
\begin{tabular}{llrrrrr}
\toprule
\textbf{Metric} & \textbf{Stat} & \textbf{Orig.} & \textbf{Ours} & \textbf{Base} & \textbf{VAE} & \textbf{DCGAN} \\
\midrule
\multirow{2}{*}{Noise std}
  & mean   & $0.041{\pm}0.016$ & $0.028{\pm}0.009$ & \bm{$0.043{\pm}0.017$} & $0.019{\pm}0.012$ & $\underline{0.035{\pm}0.025}$ \\
  & med.   & $0.037\,[0.027]$  & $\underline{0.028\,[0.015]}$  & \bm{$0.043\,[0.021]$}  & $0.015\,[0.013]$  & $0.026\,[0.011]$  \\
\midrule
\multirow{2}{*}{SNR}
  & mean   & $20.37{\pm}3.18$ & $25.50{\pm}3.55$ & \bm{$21.54{\pm}4.40$} & $28.81{\pm}4.80$ & $\underline{23.11{\pm}2.31}$ \\
  & med.   & $20.07\,[3.24]$  & $25.24\,[4.32]$  & \bm{$20.35\,[6.38]$}  & $29.17\,[6.92]$  & $\underline{23.77\,[1.07]}$  \\
\midrule
\multirow{2}{*}{HFNR}
  & mean   & $1.121{\pm}0.027$ & $\underline{1.140{\pm}0.020}$ & $1.148{\pm}0.019$ & $1.196{\pm}0.007$ & \bm{$1.128{\pm}0.012$} \\
  & med.   & $1.120\,[0.050]$  & $\underline{1.143\,[0.033]}$  & $1.152\,[0.020]$  & $1.197\,[0.009]$  & \bm{$1.131\,[0.013]$}  \\
\bottomrule
\end{tabular}}
\end{table}

On NANO-TEM (Tab.~\ref{tab:noise_comparison_nano}), three of the four generative models smooth the images: our DDPM, the VAE and DCGAN reduce the noise standard deviation and raise the SNR relative to the originals ($0.041 \pm 0.016$, $20.37 \pm 3.18$\,dB, HFNR $1.121 \pm 0.027$), whereas the baseline DDPM preserves noise amplitude and SNR most closely ($0.043 \pm 0.017$, $21.54 \pm 4.40$\,dB). Our DDPM yields $0.028 \pm 0.009$, $25.50 \pm 3.55$\,dB and HFNR $1.140 \pm 0.020$. DCGAN is closest on the spectral statistic (HFNR $1.128 \pm 0.012$) but deviates further than the baseline DDPM in amplitude and SNR. Together with the distribution-level results in Tab.~\ref{tab:similarity_metrics_nano}, this indicates that the DDPM models preserve the image distribution (FID, KID) and perceptual similarity (LPIPS) better than the adversarial baseline, while our variant smooths the atomic-resolution images during sampling.

\subsection{Domain Transfer}

\begin{figure}[t]
  \centering
  \begin{tikzpicture}
    \node[inner sep=0, anchor=south west] (img)
      {\includegraphics[width=\columnwidth]{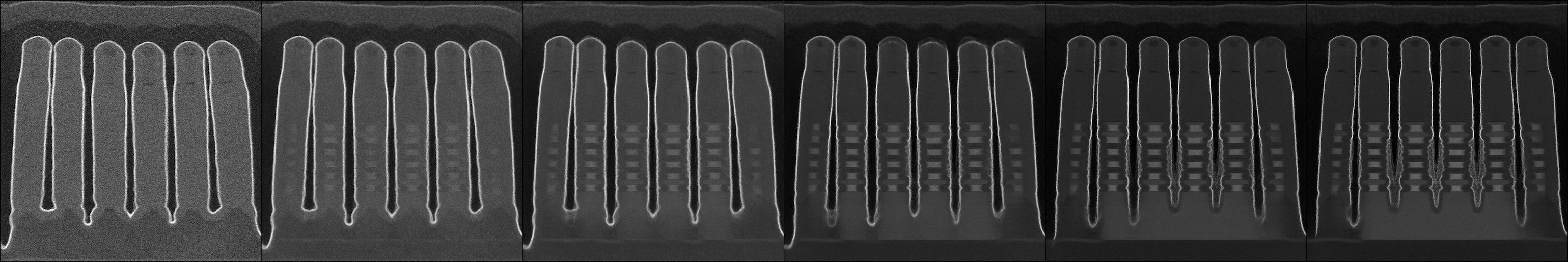}};
    \def\ncols{6}
    \def\pad{-3pt}
    \foreach \i/\lab in {0/a,1/b,2/c,3/d,4/e,5/f}{
      \node[anchor=north west]
        at ($ (img.north west)!{\i/\ncols}!(img.north east) + (\pad,-\pad) $)
        {\paneltag{\lab}};
    }
  \end{tikzpicture}
  \caption{(a) Original; (b--f) domain transfer outputs after timesteps $t \in \{50, 100, 200, 300, 400\}$.}
  \label{fig:domaintransfer}
\end{figure}

We incorporate domain transfer capabilities into our framework, adapted from \cite{smith_realistic_2022} and corresponding to SDEdit \cite{meng_sdedit_2022}, which enable the transformation of images across different imaging conditions or process parameters, while preserving high-level structural features. We apply the forward diffusion process to an input image $x_0$ for $T' \leq T$ timesteps, followed by reverse denoising using our trained DDPM. Specifically, given an input image $x_0$ outside the training set, we first apply noise according to:
\begin{equation}
x_{T'} = \sqrt{\bar{\alpha}_{T'}} x_0 + \sqrt{1 - \bar{\alpha}_{T'}} \epsilon
\end{equation}
where $\epsilon \sim \mathcal{N}(0, \mathbf{I})$ and $\bar{\alpha}_{T'} = \prod_{t=1}^{T'} \alpha_t$. The noisy image $x_{T'}$ is then processed through the reverse diffusion steps to generate an output adapted to the characteristics of the target domain, while maintaining structural properties of the input. The noise level parameter $T'$ governs the trade-off between domain adaptation and structural preservation. Higher values of $T'$ generate images closely aligned with the training distribution but with reduced retention of the original structure, whereas lower $T'$ values retain more structural details while still transferring domain-specific properties. As Fig.~\ref{fig:domaintransfer} demonstrates, this approach facilitates the transfer of structural features from a source to a target domain. It enables inter-domain transfer by adapting features from unseen input images (those not included in the training dataset and originating from different domains) to the target domain, aligning them with the characteristics of the training data. Furthermore, it supports the transfer of structural attributes across datasets derived from distinct training domains. For semiconductor characterization, this addresses cross-tool and cross-process compatibility by generating images that incorporate characteristics from different imaging conditions, accelerating voltages, or specimen preparation protocols. This approach aims to reduce the need for extensive and potentially destructive TEM data acquisition under specific configurations, thereby accelerating analysis workflows and enhancing flexibility across different equipment setups. Validation of this claim requires comprehensive experimental studies beyond the scope of the present work.

\subsection{Classifier Guidance}
\begin{figure}[t]
\centering
\panel[0.2\columnwidth]{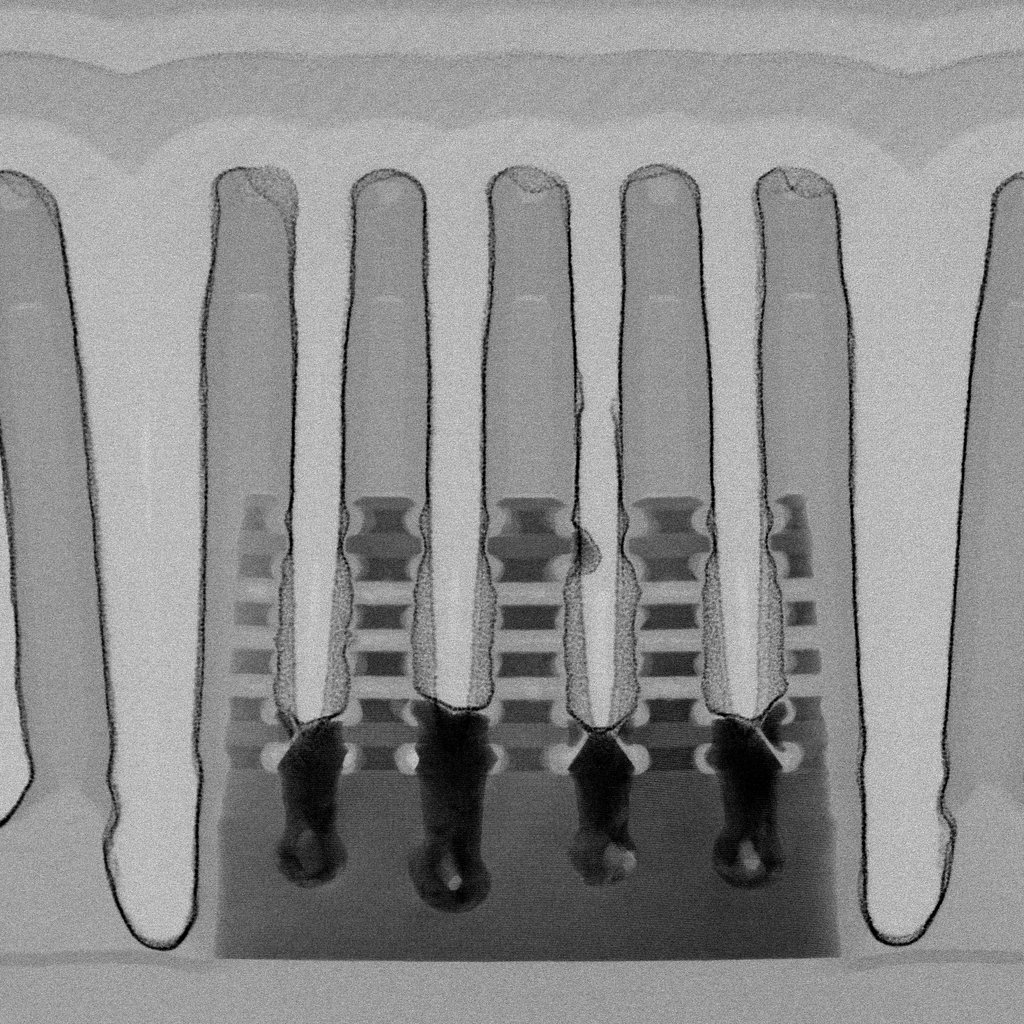}{a}\hfill
\panel[0.2\columnwidth]{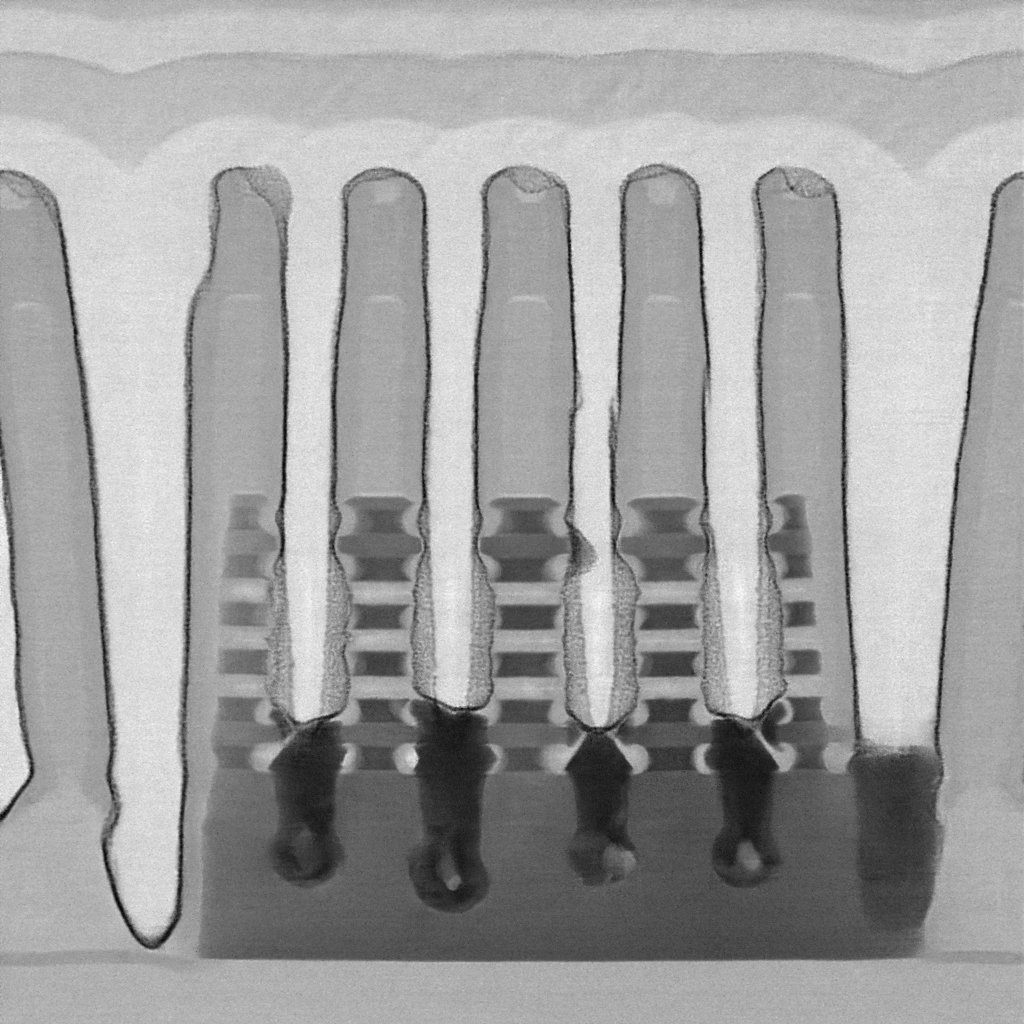}{b}\hfill
\panel[0.2\columnwidth]{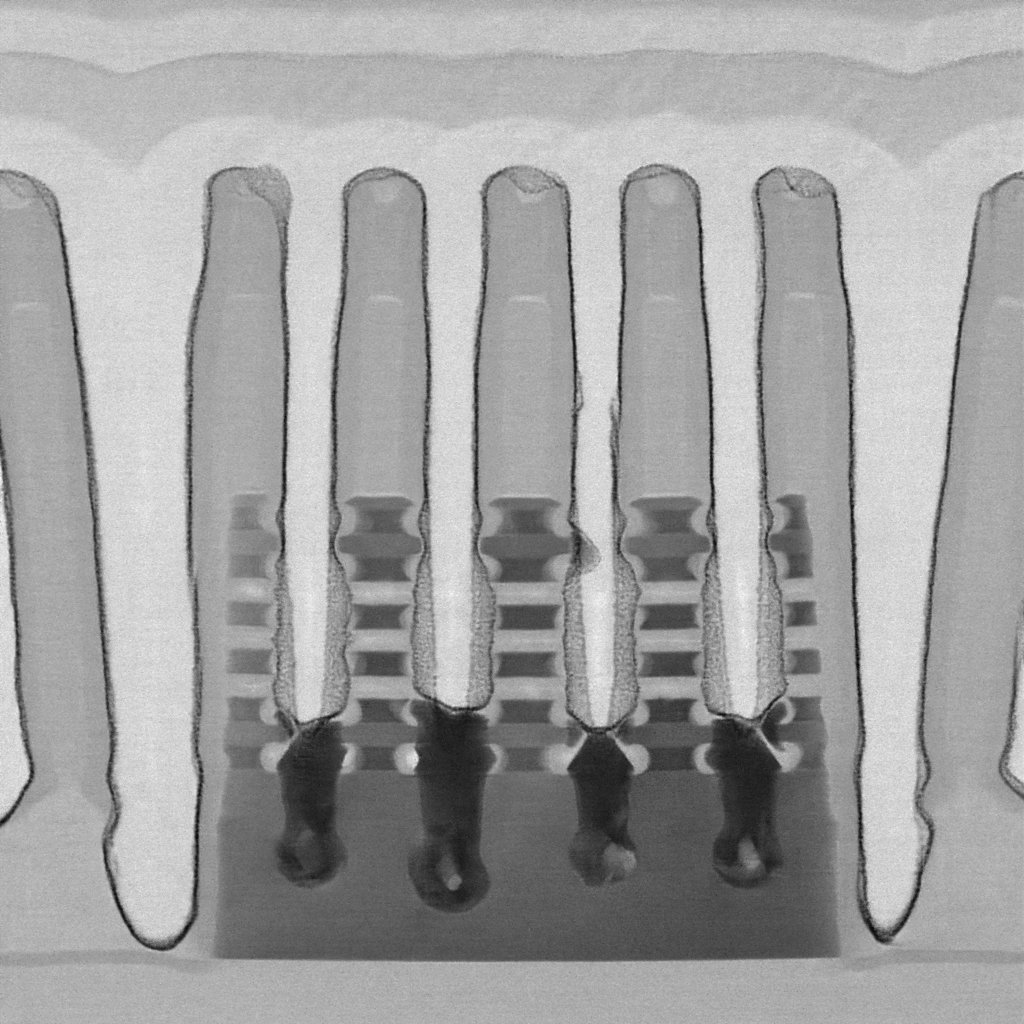}{c}\hfill
\panel[0.2\columnwidth]{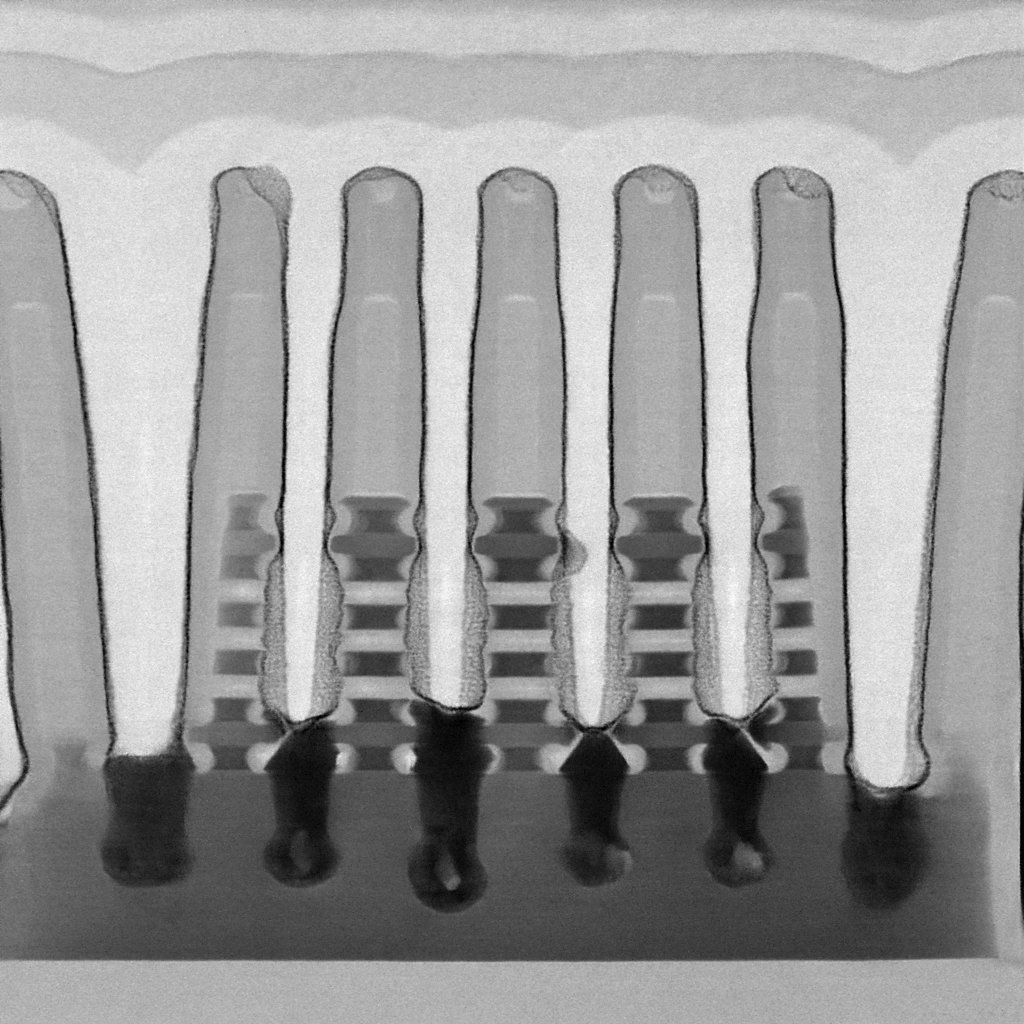}{d}\\[4pt]
\panel[0.2\columnwidth]{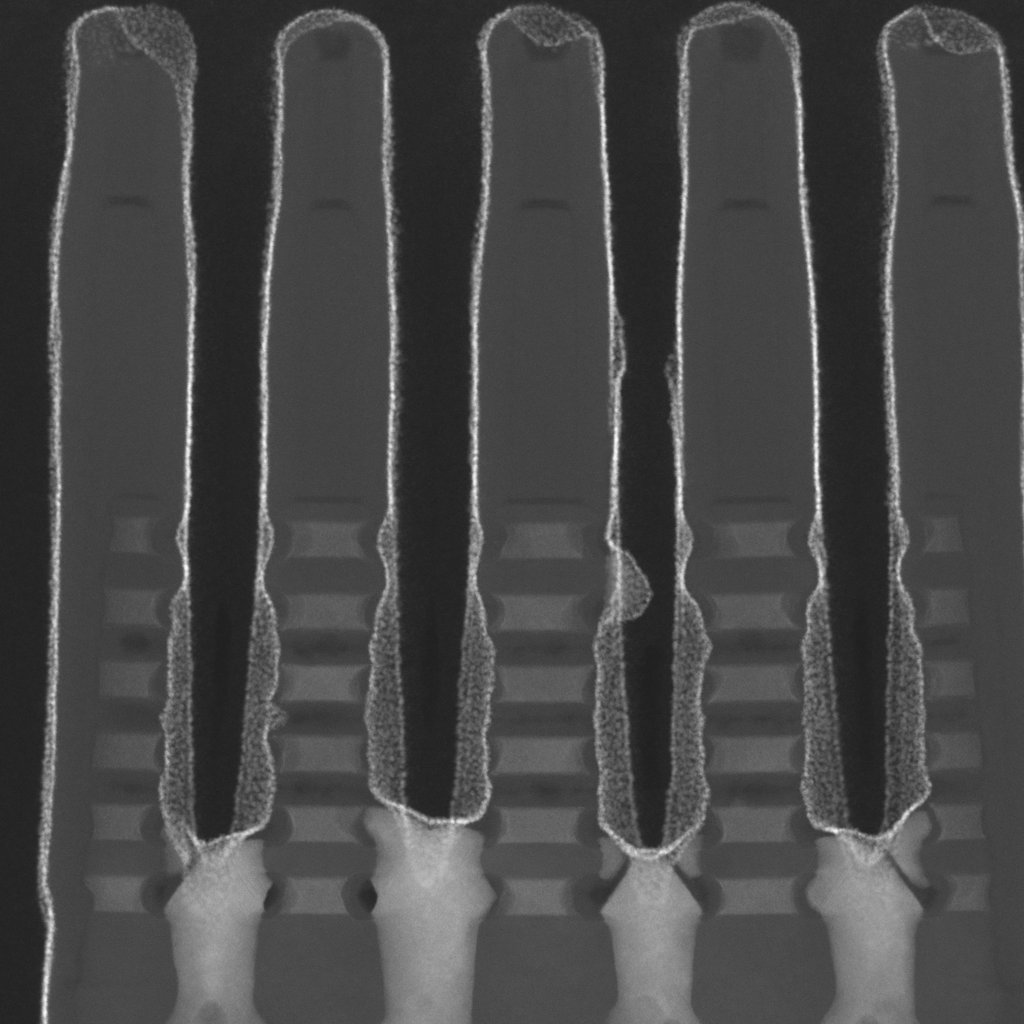}{e}\hfill
\panel[0.2\columnwidth]{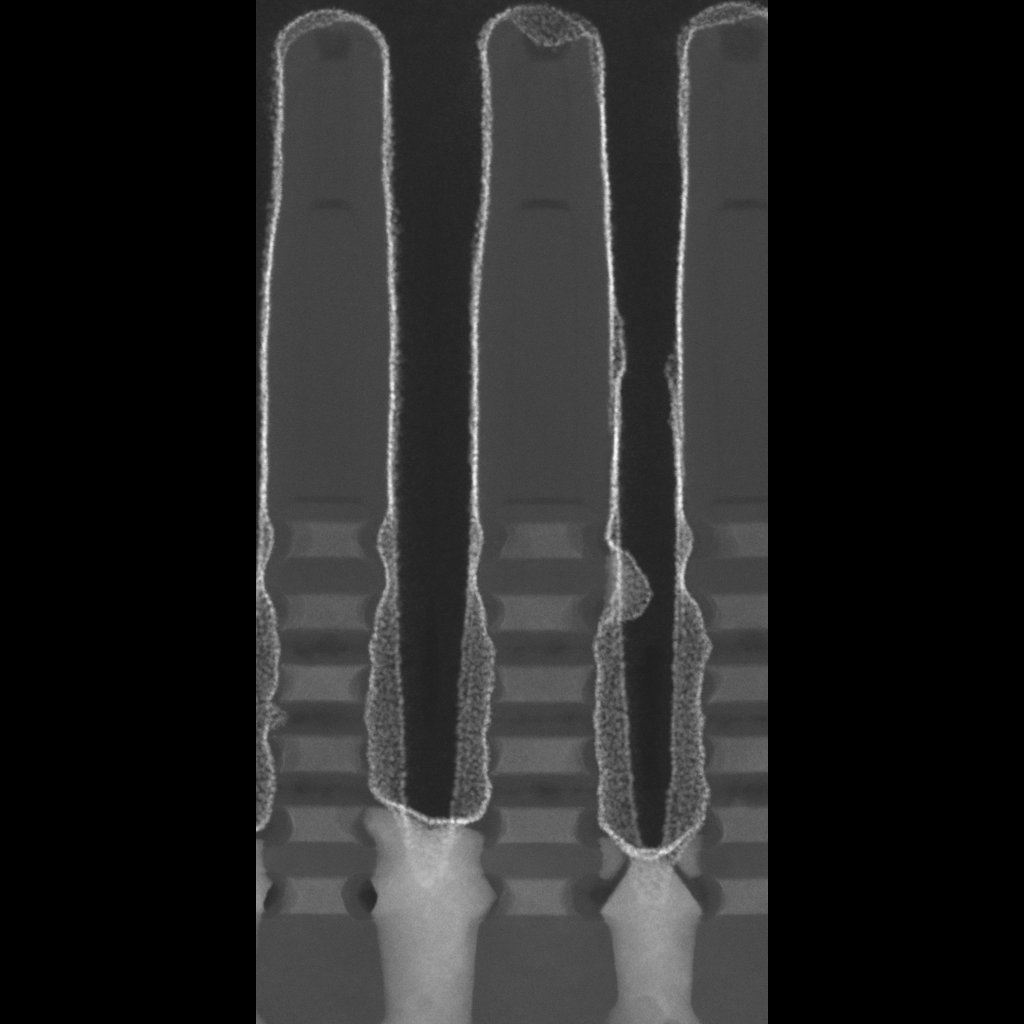}{f}\hfill
\panel[0.2\columnwidth]{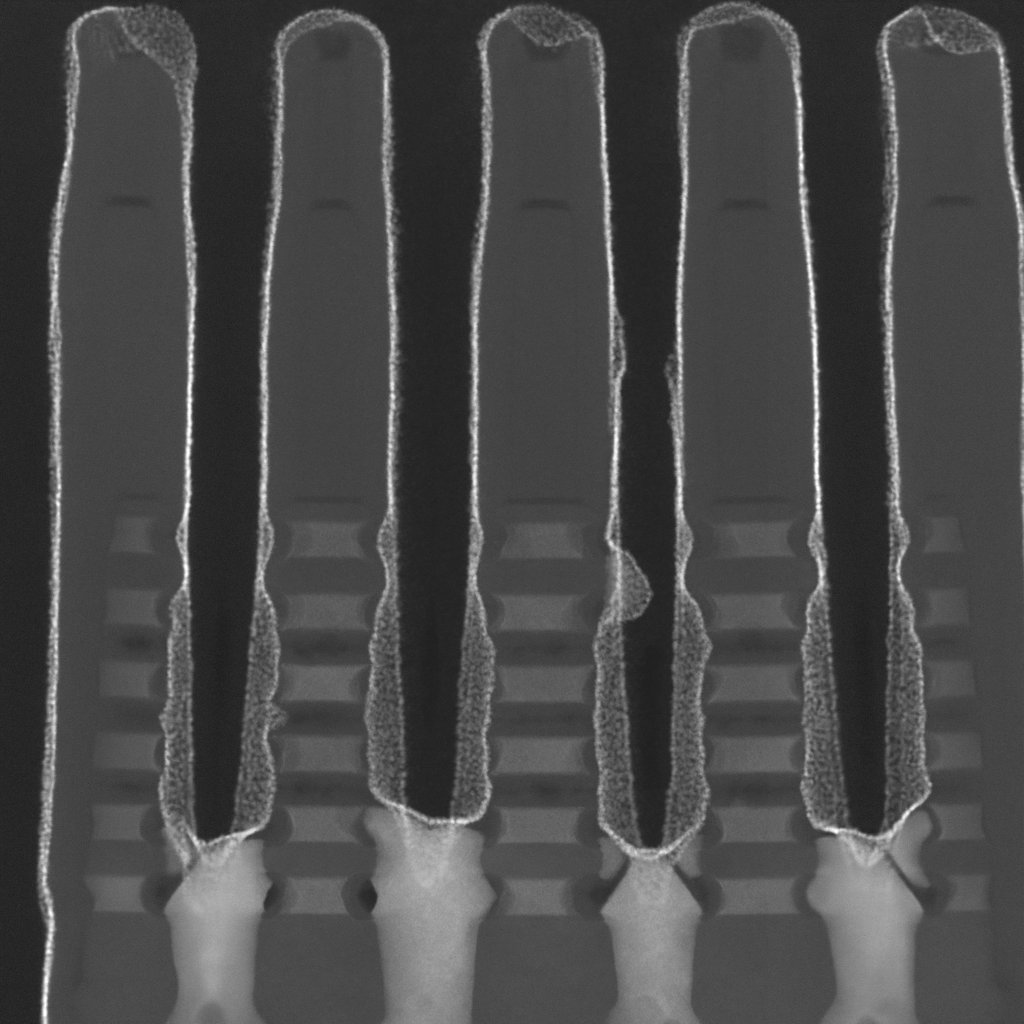}{g}\hfill
\phantom{\panel[0.2\columnwidth]{plots/inpainting_reconstructed.jpg}{}}
\caption{(a) Original; (b--d) guidance-scale ($s = 3, 5, 12$); (e) original, (f) mask, (g) inpainted reconstruction.}
\label{fig:guidance-inpainting}
\end{figure}

We enable the controllable generation of specific classes through classifier guidance adapted from \cite{dhariwal_diffusion_2021b}. This approach employs an external classifier trained on noisy images at multiple timesteps to guide the diffusion sampling process toward desired target classes. It modifies DDPM sampling by incorporating classifier gradients as follows:
\begin{equation}
x_{t-1} \sim \mathcal{N}\!\Big(
\mu_\theta(x_t,t) + s\,\Sigma_t\,\nabla_{x_t}\log p_\phi(y\mid x_t,t),\;\Sigma_t
\Big)
\label{eq:ddpm-guidance}
\end{equation}
Here, $s$ describes the guidance scale, $p_\phi(y|x_t, t)$ represents the classifier's predicted probability for target class $y$, and $\nabla_{x_t} \log p_\phi(y|x_t, t)$ provides the guidance gradient. We trained a noise-aware classifier conditioned on timesteps, using the augmented DEVICE-TEM dataset. As illustrated in Fig.~\ref{fig:guidance-inpainting}~(a-d), classifier guidance effectively steers the generation process toward specific classes, enhancing class-conditional sample quality. Increasing the guidance scale improves class fidelity but tends to reduce the diversity of generated structures and textures. However, external classifiers present several limitations: (i) they require separate training and maintenance of classification models, (ii) the fidelity of guidance depends directly on classifier robustness and accuracy, and (iii) the two-stage training process increases computational complexity and resource demands.

\subsection{Inpainting}
Inpainting enables the reconstruction of masked or missing regions within an image while ensuring consistency with surrounding content. In the context of TEM applications, this capability facilitates image extension, super-resolution, and seamless integration of images acquired under varying conditions into unified datasets. We adapt the RePaint methodology \cite{lugmayr_repaint_2022b}, leveraging the denoising process of DDPMs for conditional generation. In the reverse diffusion process, we combine information from known regions of the original image with predictions for unknown regions at each denoising step $t$:
\begin{equation}
\begin{aligned}
x_{t-1}^{\text{known}} &\sim \mathcal{N}(\sqrt{\bar{\alpha}_{t-1}} x_0, (1 - \bar{\alpha}_{t-1})\mathbf{I}) \\
x_{t-1}^{\text{unknown}} &\sim \mathcal{N}(\mu_\theta(x_t, t), \Sigma_\theta(x_t, t))
\end{aligned}
\end{equation}
\begin{equation}
x_{t-1} = m \odot x_{t-1}^{\text{known}} + (1 - m) \odot x_{t-1}^{\text{unknown}}
\end{equation}
Here, $m$ represents the binary mask indicating known regions, $x_{t-1}^{\text{known}}$ is sampled using corresponding known pixels from original image $x_0$, while $x_{t-1}^{\text{unknown}}$ is generated by the diffusion model for unknown, masked regions. The mask ensures that known regions remain untouched while unknown regions are synthesized in a manner consistent with the learned imaging characteristics. This method enables several TEM-specific applications: \textbf{(i)} extension of smaller Field-of-View (FoV) images to facilitate broader structural analysis, \textbf{(ii)} super-resolution reconstruction as detailed by \cite{lugmayr_repaint_2022b}, and \textbf{(iii)} integration of images acquired at varying magnifications or imaging conditions. Fig.~\ref{fig:guidance-inpainting}~(e-g) demonstrates the coherent reconstruction of a masked region. Here, 10 resampling iterations with a jump length of 10 steps, applied every 10 diffusion steps, produced the most coherent inpainting results. After 181 minutes of sampling on a NVIDIA L40S GPU, the reconstruction is visually coherent with the known region; computed over the full image, about half of which is the unmasked original, the metrics are MS-SSIM: 0.9901, SSIM: 0.9487, PSNR (dB): 37.0895, LPIPS: 0.0105.

\subsection{Segmentation}
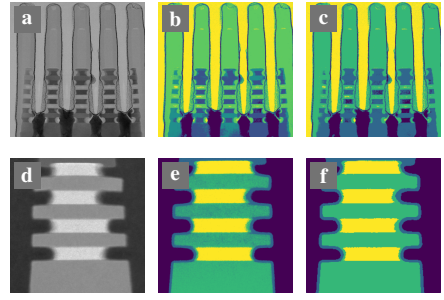
\begin{figure}[t]
  \centering
  \begin{tikzpicture}
    \node[inner sep=0, anchor=south west] (img1)
      {\includegraphics[width=0.2\linewidth]{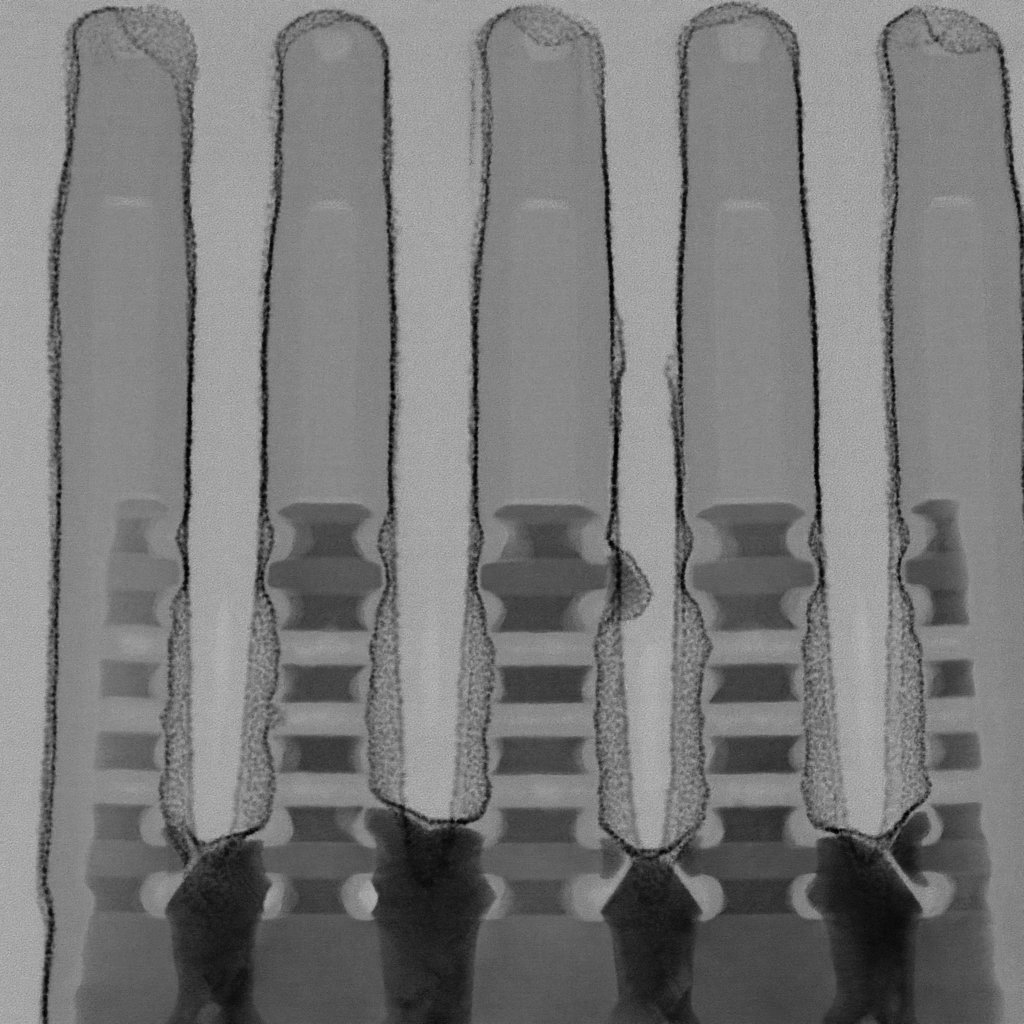}};
    \node[inner sep=0, anchor=south west] (img2) at ([xshift=5pt]img1.south east)
      {\includegraphics[width=0.2\linewidth]{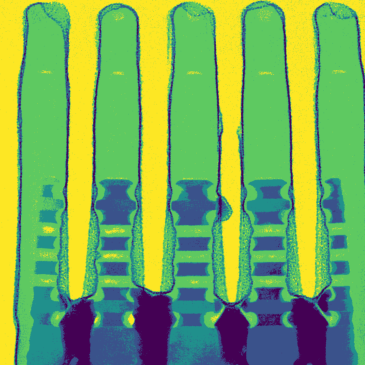}};
    \node[inner sep=0, anchor=south west] (img3) at ([xshift=5pt]img2.south east)
      {\includegraphics[width=0.2\linewidth]{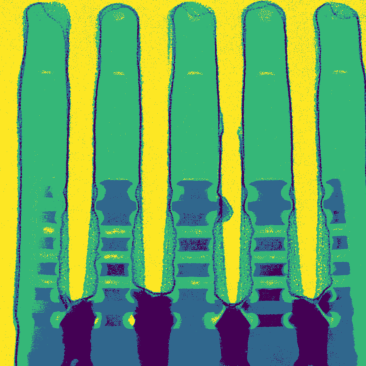}};

    \node[inner sep=0, anchor=north west] (img4) at ([yshift=-8pt]img1.south west)
      {\includegraphics[width=0.2\linewidth]{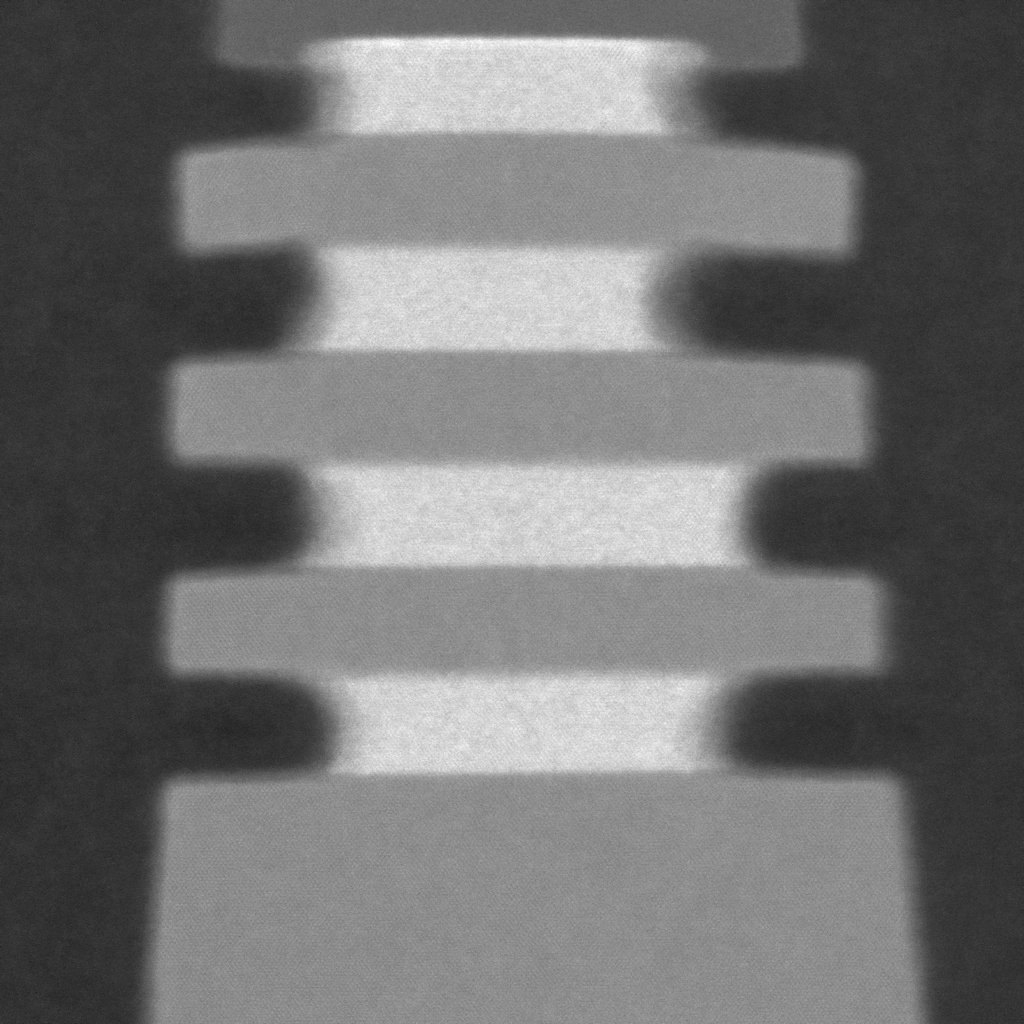}};
    \node[inner sep=0, anchor=south west] (img5) at ([xshift=5pt]img4.south east)
      {\includegraphics[width=0.2\linewidth]{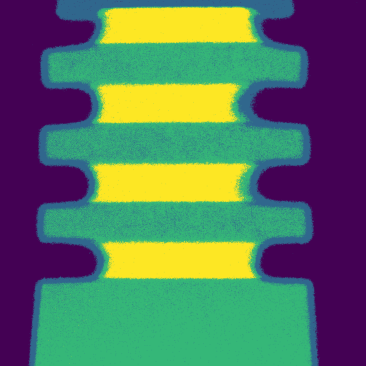}};
    \node[inner sep=0, anchor=south west] (img6) at ([xshift=5pt]img5.south east)
      {\includegraphics[width=0.2\linewidth]{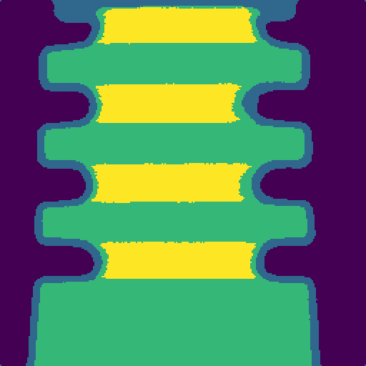}};

    \node[anchor=north west] at ([xshift=-2pt,yshift=2pt]img1.north west) {\paneltag{a}};
    \node[anchor=north west] at ([xshift=-2pt,yshift=2pt]img2.north west) {\paneltag{b}};
    \node[anchor=north west] at ([xshift=-2pt,yshift=2pt]img3.north west) {\paneltag{c}};
    \node[anchor=north west] at ([xshift=-2pt,yshift=2pt]img4.north west) {\paneltag{d}};
    \node[anchor=north west] at ([xshift=-2pt,yshift=2pt]img5.north west) {\paneltag{e}};
    \node[anchor=north west] at ([xshift=-2pt,yshift=2pt]img6.north west) {\paneltag{f}};
  \end{tikzpicture}
  \caption{Synthetic input (a, d) and partition based on intensity (b, e) and encoder features of the trained DDPM (c, f).}
  \label{fig:prelim-seg}
\end{figure}

Intermediate feature representations of trained DDPMs can be repurposed for image segmentation \cite{baranchuk_labelefficient_2021, couairon_diffcut_2024a}. Concretely, we extract encoder feature maps and apply simple partitioning (e.g., k-means and HDBSCAN clustering or graph-based splits) in this learned feature space to obtain coherent region masks that align well with visually salient microstructural domains. Our results confirm that DDPMs implicitly learn semantically meaningful representations like textures and boundaries that transfer to segmentation. As with most unsupervised segmentations, the resulting masks may be imperfect; however, they can be efficiently curated by visual selection and, when needed, corrected in a lightweight manner, making them practical starting points for downstream segmentation workflows. Figure~\ref{fig:prelim-seg} shows an example, exemplifying how the encoder features lead to a more nuanced segmentation.

\section{Limitations and Future Work}
\label{sec:limitations_future_work}

While we demonstrate successful synthetic TEM image generation, several limitations must be considered. On NANO-TEM, reproducing atomic-scale periodicity and fine contrast variations remains more difficult than generating the larger-scale structural geometries of the DEVICE-TEM dataset; although our model achieves the best distribution-level scores (FID, KID) and the best best-case structural similarity among non-mode-collapsing baselines, mean image-wise similarity stays below the values observed for DEVICE-TEM. Additionally, our approach requires substantial computational resources and careful hyperparameter tuning, as the models were developed from scratch. Patch extraction resamples partial tiles at the image edges to square patches, introducing an anisotropic rescaling for a fraction of the patches at each stage below $1024\times1024$; a tiling scheme that avoids this is left for future work.

Our noise analysis relies on the estimator definitions of Eqs.~(\ref{eq:laplacian_kernel})--(\ref{eq:hfnr}), which we apply identically to original and synthetic images. Two aspects limit their absolute interpretation. The Laplacian-based estimator of Eq.~(\ref{eq:noise_std}) is evaluated with zero padding at the image border and with a scale constant specific to this implementation, so reported noise standard deviations are estimator-dependent. Moreover, the Laplacian responds to fine structure such as edges and lattice fringes as well as to noise, and the zero-padded border adds a brightness-dependent term, so differences in $\sigma_n$ between models may partly reflect fine-structure contrast or brightness rather than noise alone. The frequency partition underlying Eq.~(\ref{eq:hfnr}) fixes the absolute scale of HFNR, so that quantity is meaningful as a relative comparison rather than as an absolute high-frequency fraction. We therefore base our conclusions on the ordering of the models and their distances from the original rather than on absolute values, and read them as differences in smoothness rather than in noise alone. A revised noise characterization (border-free estimation, an explicitly normalized spectral partition, and validation of both estimators against images with known noise) is left for future work, alongside the broader noise models (Poisson-Gaussian, shot, spatially correlated, signal-dependent) discussed below.

Future work includes guidance mechanisms that avoid external classifiers, hybrid coupling with physics-based electron-specimen models, broader noise characterizations (Poisson-Gaussian, shot, spatially correlated, signal-dependent), together with a revised and independently validated implementation of the noise and spectral estimators used here, and integration with automated acquisition pipelines toward semi-autonomous characterization \cite{spurgeon_datadriven_2021}. Foundation models and their fine-tuning are a further promising direction.

\section{Conclusion}
\label{sec:conclusion}

This work presents a generative framework that produces high-fidelity, high-resolution synthetic TEM images ($1024 \times 1024$) from extremely limited datasets. The proposed progressive patch-based generative framework was experimentally validated on two real wafer TEM datasets. On DEVICE-TEM, the resulting synthetic images lead the baselines on most structural, perceptual, and distribution-level metrics (all except SSIM and FID), indicating strong preservation of the structural and statistical properties relevant to semiconductor metrology. On the more challenging NANO-TEM dataset, our model offers the best balance of best-case structural fidelity and distribution-level fidelity, leading on FID and KID while avoiding the mode collapse that inflates the similarity scores of the VAE baseline. Beyond data generation, we show that intermediate DDPM feature representations can be repurposed for unsupervised image segmentation via simple partitioning in learned feature space, yielding coherent region masks that align with salient microstructural domains and provide practical, easily curated starting points for downstream segmentation workflows. By addressing the data scarcity imposed by destructive and costly TEM acquisition, and by providing auxiliary segmentation functionality from the same generative backbone, our approach is intended to support more robust learning for defect inspection, segmentation, and metrology, enabling broader adoption of machine learning-driven methods in advanced semiconductor manufacturing.

\section*{Acknowledgments}
The authors acknowledge the DDPM implementation by Phil Wang
(\href{https://gitlab.com/lucidrains/denoising-diffusion-pytorch/}{gitlab.com/lucidrains/denoising-diffusion-pytorch/}),
which provided the basis for the diffusion model architecture used in this work.
LLMs from the Anthropic Claude Opus 4 and OpenAI GPT-5 families
were used for assistance with code prototyping and proofreading. The authors reviewed
and validated all generated outputs.

\bibliographystyle{IEEEtranDOI}
\bibliography{bibliography}

\clearpage

\appendix 
\section{Appendix} 
\input{appendix.tex}

\end{document}

%% file: math_commands.tex
\usepackage{amsmath,amsfonts,bm}

\def\eqref#1{equation~\ref{#1}}

\def\1{\bm{1}}

\DeclareMathAlphabet{\mathsfit}{\encodingdefault}{\sfdefault}{m}{sl}
\SetMathAlphabet{\mathsfit}{bold}{\encodingdefault}{\sfdefault}{bx}{n}



%% file: figs/framework.tex
\begin{tikzpicture}[
  line cap=round, line join=round,
  >= {Stealth[length=2.0mm,width=1.6mm]},
  every node/.style={font=\large},
  stage/.style={rounded corners, draw, minimum height=18mm, inner sep=2.2mm, 
                line width=0.8pt, align=center, fill=gray!18, text width=42mm},
  proc/.style={rounded corners, draw, minimum height=12mm, inner sep=2.2mm, 
               line width=0.8pt, align=center, fill=blue!10, text width=38mm},
  aug/.style={draw, dashed, rounded corners, align=center, font=\normalsize, 
              fill=blue!5, minimum height=9mm, inner sep=2mm, text width=44mm},
  bus/.style={dashed, line width=0.8pt},
  underlay/.style={rounded corners, fill=cyan!12, inner sep=4mm, 
                   line width=0pt}
]

\newcommand{\squareimg}[2]{
  \begin{scope}
    \clip ([shift={(-7mm,-7mm)}]#1) rectangle ([shift={(7mm,7mm)}]#1);
    \node[inner sep=0pt, anchor=center] at (#1) {\includegraphics[width=14mm]{#2}};
  \end{scope}%
}

\matrix (M) [row sep=0mm, column sep=6mm] at (0,0) {
  \node[stage, text width=32mm, minimum height=16mm] (s64)   
       {$\mathbf{64\times 64}$\\{\normalsize Initial Model}}; &
  \node[stage, text width=32mm, minimum height=16mm] (s128)  
       {$\mathbf{128\times 128}$\\{\normalsize Extend}}; &
  \node[stage, text width=32mm, minimum height=16mm] (s256)  
       {$\mathbf{256\times 256}$\\{\normalsize Extend}}; &
  \node[stage, text width=32mm, minimum height=16mm] (s512)  
       {$\mathbf{512\times 512}$\\{\normalsize Extend}}; &
  \node[stage, text width=32mm, minimum height=16mm] (s1024) 
       {$\mathbf{1024\times 1024}$\\{\normalsize Final Model}}; \\
};

\coordinate (midS) at ($(s64)!0.5!(s512)$);
\node[aug, above=6mm of s1024] (augFull) {Custom TA \&\ \\downscaling};
\node[aug] (augPatch) at ($(midS |- augFull)$) {Torchvision TA \&\ \\patch extraction};

\coordinate (augMidC) at ($ (augPatch.center)!0.5!(augFull.center) $);
\node[proc, text width=34mm] (train) at ($(augMidC) + (0,1.5cm)$) {Training Data};

\draw[->, line width=0.9pt, shorten >=0.5mm] ($(train.south west) + (0.5mm,0.5mm)$) -- (augPatch.north east);
\draw[->, line width=0.9pt, shorten >=0.5mm] ($(train.south east) + (-0.5mm,0.5mm)$) -- (augFull.north west);

\node[draw=none, rounded corners, fit=(s64)(s128)(s256)(s512)(s1024),
      inner sep=2mm, line width=0.6pt] (stageGroup) {};

\def\appstep{16mm}
\node[proc, right=8mm of stageGroup] (sample) at ($(stageGroup.east) + (8mm,4.2)$) 
     {Unguided\\Sampling};
\node[proc] (guided) at ($(sample) + (0,-\appstep)$)   {Classifier-Guided\\Sampling};
\node[proc] (inpaint) at ($(sample) + (0,-2*\appstep)$) {Inpainting};
\node[proc] (domain) at ($(sample) + (0,-3*\appstep)$)  {Domain Transfer};
\node[proc] (segment) at ($(sample) + (0,-4*\appstep)$) {Segmentation};

\coordinate (busY) at ($(stageGroup.north)!0.5!(augPatch.south)$);
\coordinate (busL) at (s64.center |- busY);
\coordinate (busR) at (s512.center |- busY);
\draw[bus] (busL) -- (busR);
\draw[bus] (augPatch.south) -- ($(busL)!0.5!(busR)$);

\foreach \a/\b in {s64/s128, s128/s256, s256/s512, s512/s1024} {
  \draw[->, line width=0.9pt, shorten >=0.4pt, shorten <=0.4pt] 
       (\a.east) -- (\b.west);
}

\foreach \n in {s64,s128,s256,s512} {
  \draw[bus] (\n.center |- busL) -- (\n.center |- stageGroup.north);
}
\draw[bus] (augFull.south) -- (s1024.center |- stageGroup.north);

\draw[->, line width=0.9pt] (stageGroup.east) -- (sample.west);
\draw[->, line width=0.9pt] (stageGroup.east) -- (guided.west);
\draw[->, line width=0.9pt] (stageGroup.east) -- (inpaint.west);
\draw[->, line width=0.9pt] (stageGroup.east) -- (domain.west);
\draw[->, line width=0.9pt] (stageGroup.east) -- (segment.west);

\begin{pgfonlayer}{background}
  \node[draw, rounded corners, fit=(s64)(s128)(s256)(s512)(s1024),
        inner sep=2mm, line width=0.6pt, fill=white] {};
\end{pgfonlayer}

\def\imgWmax{22}
\def\imgratio{1.35}
\pgfmathsetmacro{\imgWmin}{\imgWmax/pow(\imgratio,4)}
\def\thumbdx{-1.4mm}
\def\thumbdy{-2.0mm}
\foreach \nodename/\k/\fname in {s64/0/64, s128/1/128, s256/2/256, s512/3/512} {
  \pgfmathsetmacro{\w}{\imgWmin*pow(\imgratio,\k)}
  \pgfmathsetmacro{\wuse}{(\k==1) ? \w*0.85 : ((\k==2) ? \w*0.9*0.85 : ((\k==3) ? \w*0.8*0.85 : \w))}
  \node[inner sep=0pt, anchor=center]
       at ($ (\nodename.south east) + (\thumbdx,\thumbdy) $)
       {\includegraphics[width=\wuse mm]{figs/\fname.png}};
}
\pgfmathsetmacro{\wfinal}{\imgWmin*pow(\imgratio,4)*0.7*0.85}
\node[inner sep=0pt, anchor=center]
     at ($ (s1024.south east) + (3mm,-2.0mm) $)
     {\includegraphics[width=\wfinal mm]{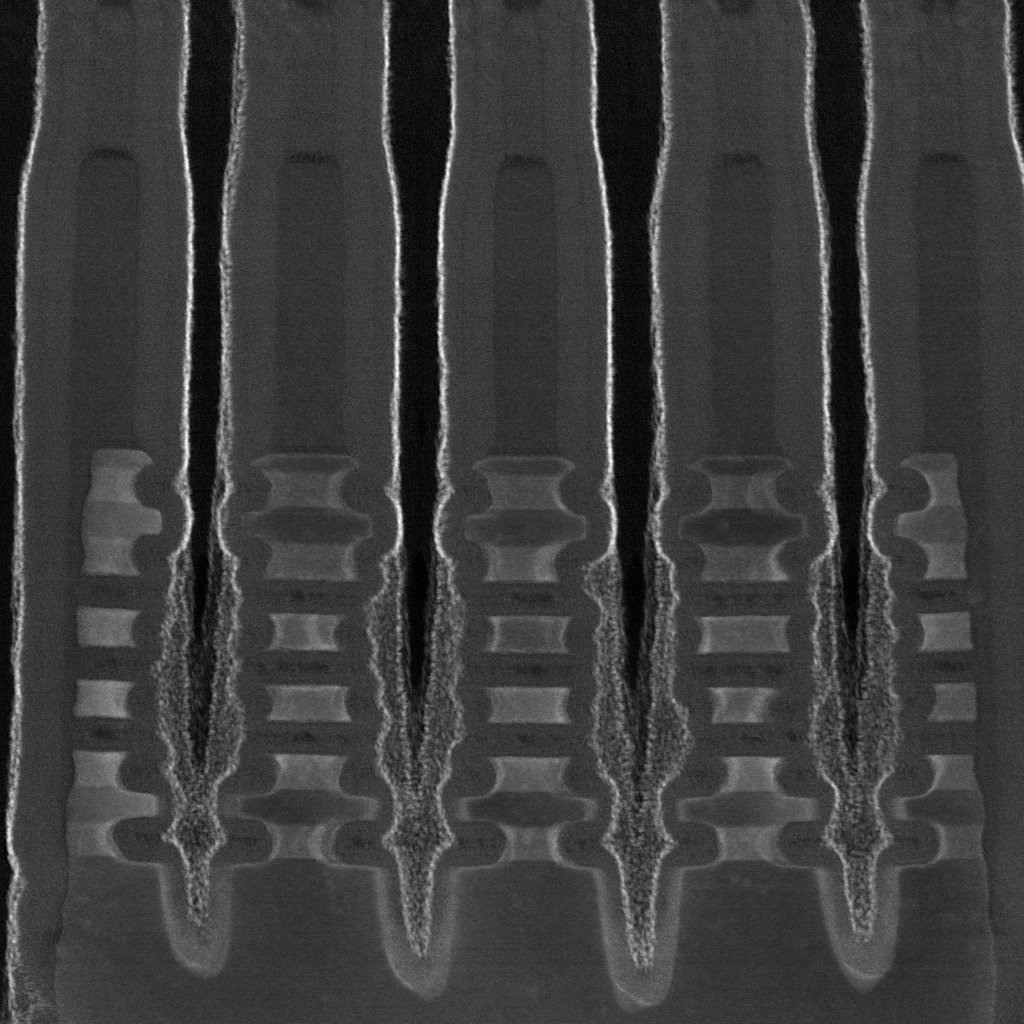}};

\foreach \i/\xoff/\yoff/\imgfile in {1/-2mm/2mm/nano-tem0.jpg, 2/0mm/0mm/nano-tem1.jpg, 3/2mm/-2mm/nano-tem2.jpg} {
  \node[inner sep=0pt, anchor=center]
    at ($ (train.east) + (14mm+\xoff,\yoff+2mm) $)
    {\includegraphics[width=15mm]{plots/\imgfile}};
}

\coordinate (imgSample)  at ($ (sample.east)  + (8.5mm,0) $);
\coordinate (imgGuided)  at ($ (guided.east)  + (8.5mm,0) $);
\coordinate (imgInpaint)  at ($ (inpaint.east)  + (8.5mm,0) $);
\coordinate (imgDomain)  at ($ (domain.east)  + (8.5mm,0) $);
\coordinate (imgSegment)  at ($ (segment.east)  + (8.5mm,0) $);

\squareimg{imgSample}{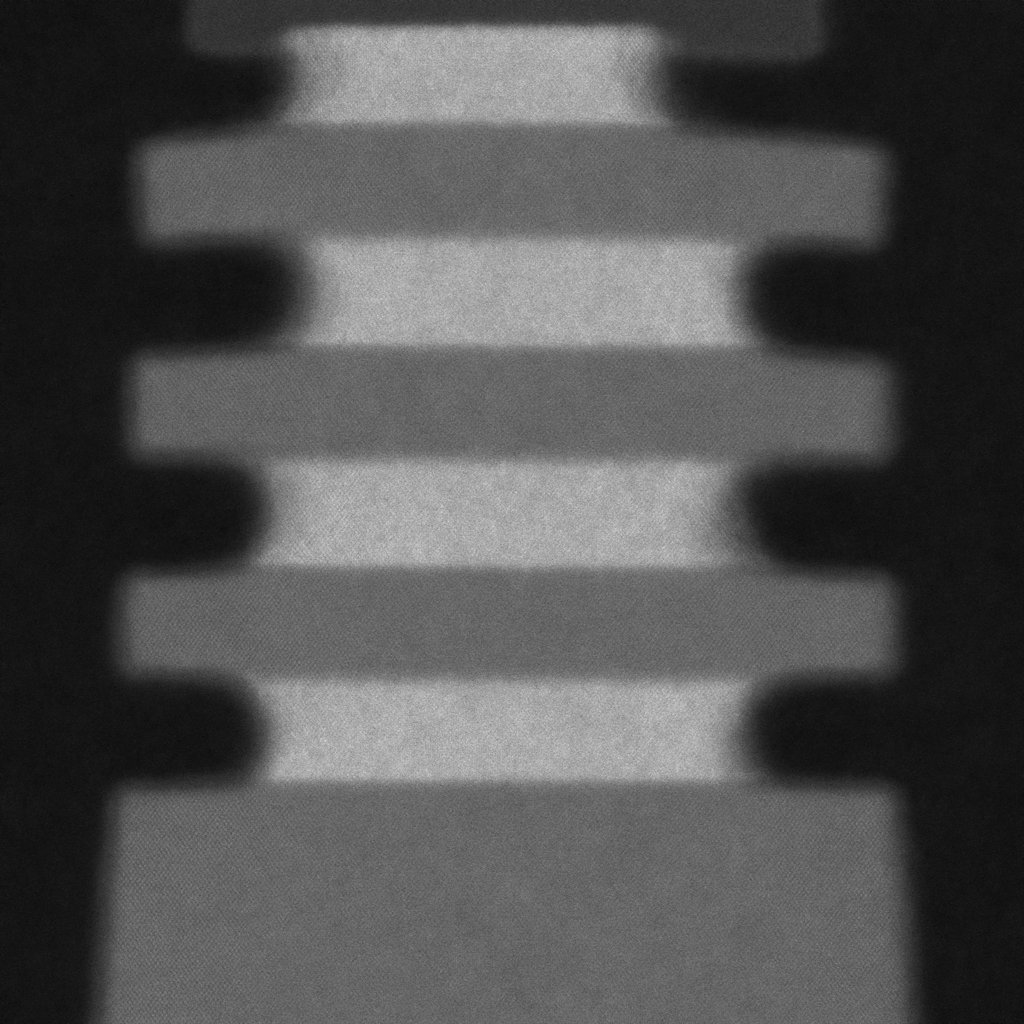}
\squareimg{imgGuided}{plots/class_dark_five_scale_5.0.jpg}
\squareimg{imgInpaint}{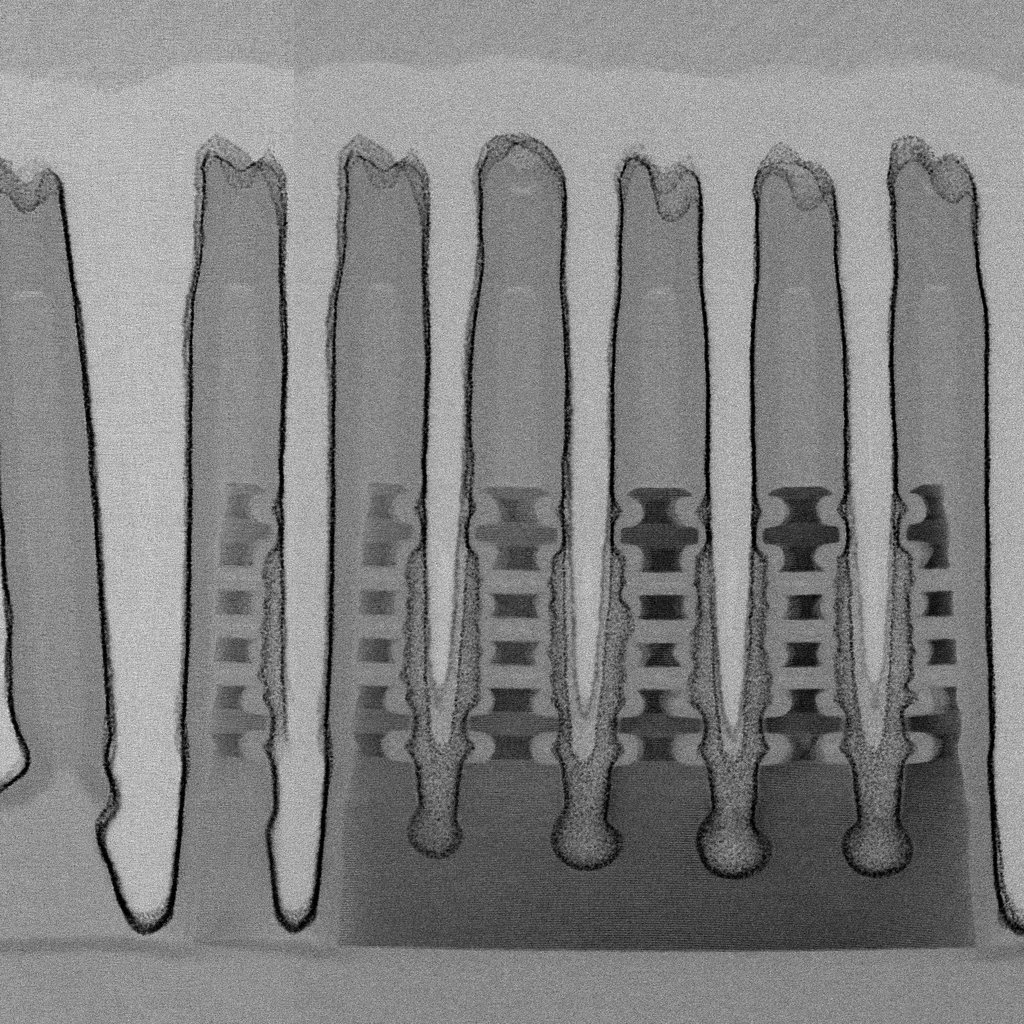}
\squareimg{imgDomain}{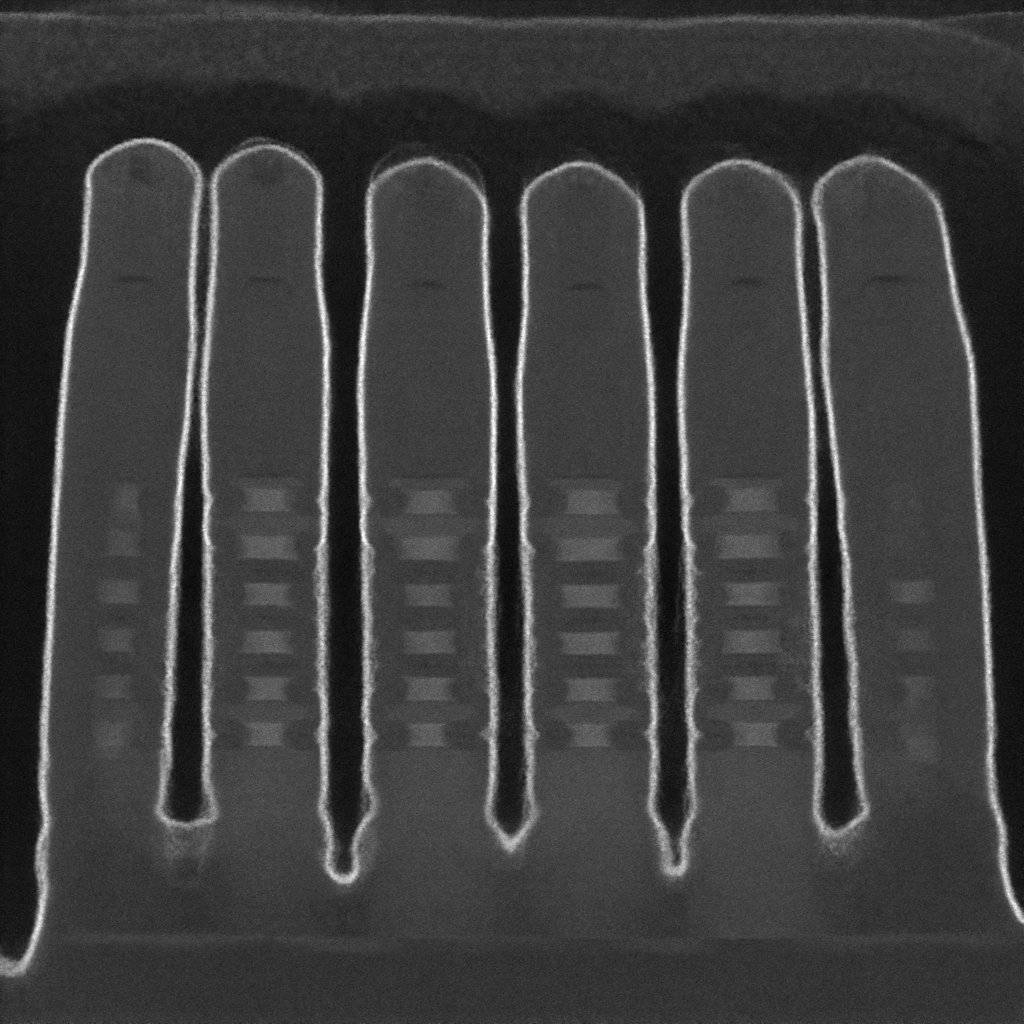}
\squareimg{imgSegment}{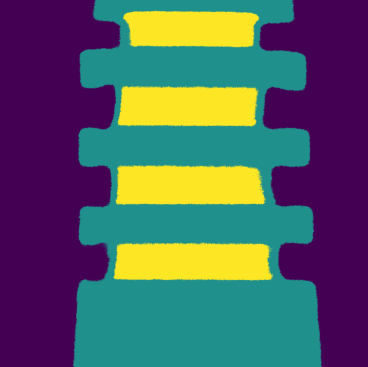}

\node[inner sep=0pt, anchor=east]
     at ($ (augPatch.west) + (-0.5mm,0) $)
     {\includegraphics[width=9mm]{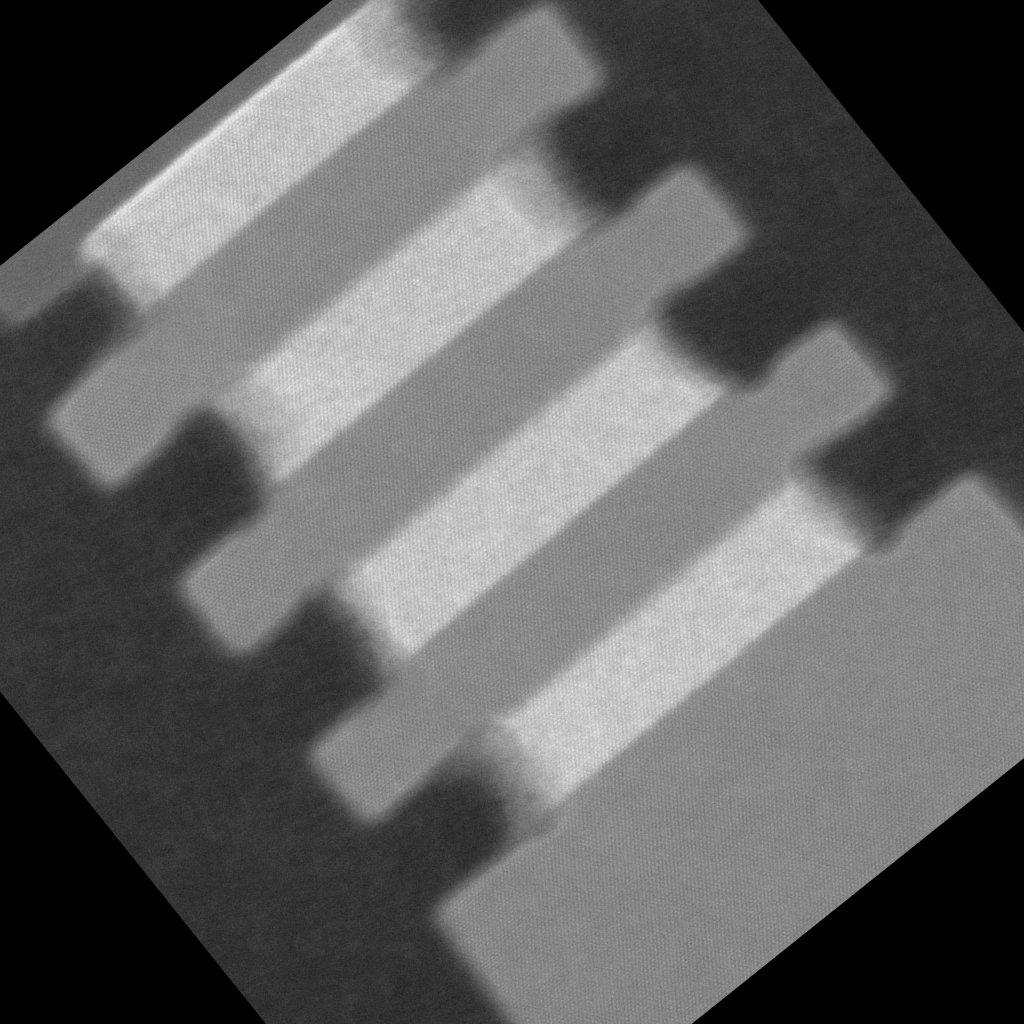}};
\node[inner sep=0pt, anchor=west]
     at ($ (augFull.east) + (0.5mm,0) $)
     {\includegraphics[width=9mm]{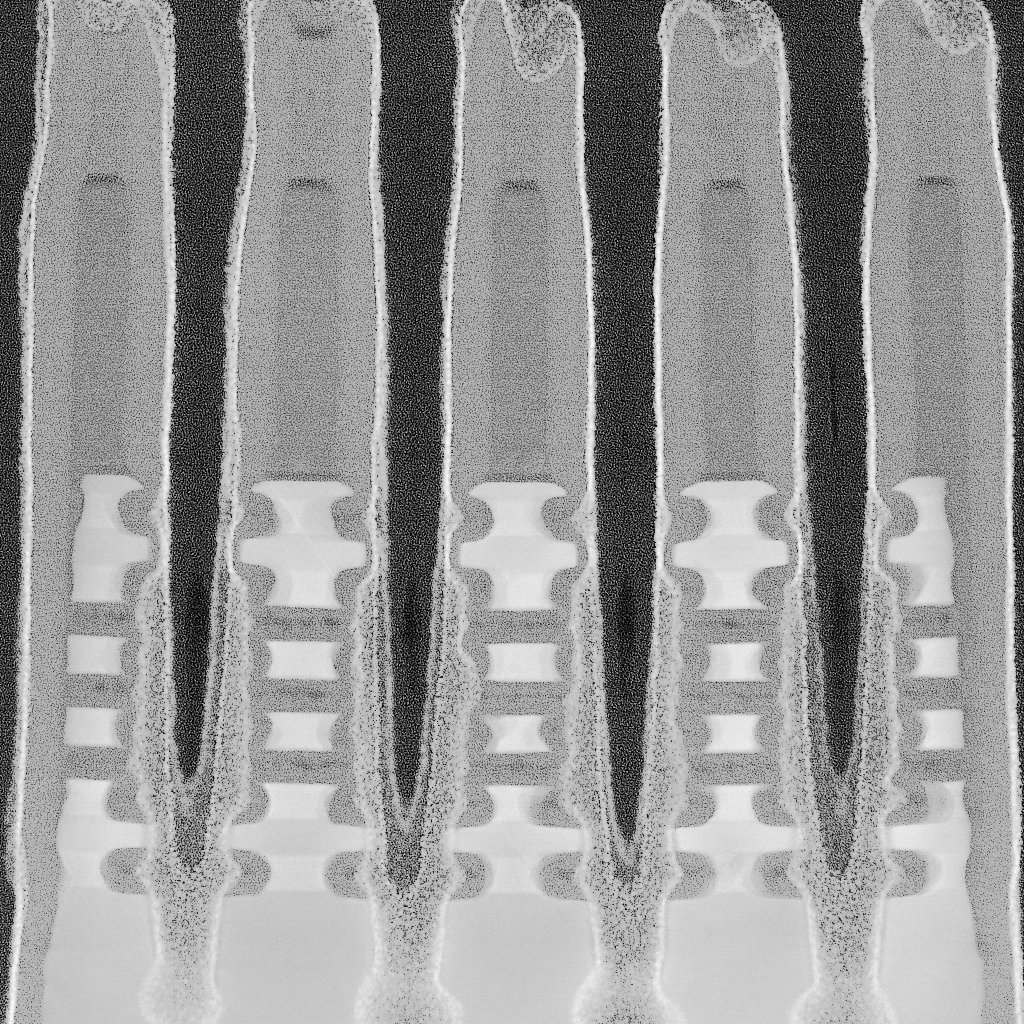}};

\end{tikzpicture}

%% file: appendix.tex
\subsection{Diffusion Framework}
\label{sec:diffusion_framework}
\FloatBarrier
\begin{figure}[H]
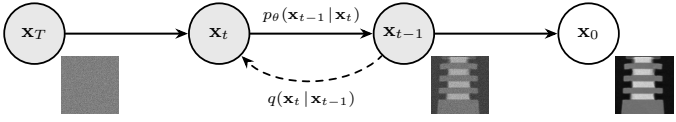

  \centering
  \includestandalone[width=\columnwidth]{figs/diffusion_framework}
  \caption{Framework of the forward and reverse diffusion process, adapted from \cite{ho_denoising_2020b}.}
  \label{fig:diffusion_framework}
\end{figure}
\FloatBarrier

\subsection{Model Architecture}
\label{sec:model_architecture}
Tab.~\ref{tab:architecture_config} summarizes the U-Net configuration shared by all progressive training stages.
\begin{table}[H]
\caption{U-Net configuration; levels are listed from the highest to the lowest spatial resolution}
\label{tab:architecture_config}
\centering
\renewcommand{\arraystretch}{1.2}
\resizebox{\columnwidth}{!}{%
\begin{tabular}{cccccc}
\toprule
\textbf{\textit{U-Net level}} & \textbf{\textit{Resolution}} & \textbf{\textit{Channels}} & \textbf{\textit{ResNet Blocks}} & \textbf{\textit{Attention Type}} & \textbf{\textit{Flash Attention}} \\
\midrule
0 & $1024\times1024$ & 64  & 2 & Linear & -- \\
1 & $512\times512$   & 64  & 2 & Linear & -- \\
2 & $256\times256$   & 128 & 2 & Linear & -- \\
3 & $128\times128$   & 256 & 2 & Linear & -- \\
4 & $64\times64$     & 512 & 2 & Linear & -- \\
5 & $32\times32$     & 512 & 2 & Linear & -- \\
6 & $16\times16$     & 512 & 2 & Full   & Training only \\
\midrule
Mid & $16\times16$   & 512 & 2 & Full   & Training only \\
\bottomrule
\end{tabular}}
\end{table}

\subsection{Augmentation Algorithm}
\label{sec:appendix_ta}

\begin{algorithm}[H]
\caption{TrivialAugment procedure for TEM images}
\label{alg:trivialaugment}
\begin{algorithmic}[1]
\Procedure{TA}{$x$: TEM image}
    \State Sample an augmentation $a$ from $\mathcal{A}$
    \State Sample a strength $m$ from $\{0, \ldots, 30\}$
    \State Apply rejection criteria for extreme values
    \State Reject if duplicate
    \State \textbf{return} $a(x, m)$
\EndProcedure
\end{algorithmic}
\end{algorithm}

\subsection{Computational Requirements}
\label{sec:computational_requirements}

\subsubsection{Training}
We experienced a minimum of 40 GB VRAM required for batch-size 1, full-scale ($1024 \times 1024$) training without torch.compile with the DEVICE-TEM dataset. With the NANO-TEM dataset, we utilized a NVIDIA 80 GB A100 GPU and batch-size 2. Batch-sizes increased by a factor of four at each smaller patch resolution stage. We employed a constant learning rate of $3\times 10^{-5}$, gradient accumulation across 16 batches, a dropout rate of 0.2, a maximum gradient norm of 0.5, a noise offset of 0.05 for earlier stages and 0.01 for the final stage, a minimum SNR gamma of 5, a cosine beta schedule and the v-prediction (velocity) parameterization. To further stabilize the training process, we used an Exponential Moving Average (EMA) model, updated every 10 iterations with a decay of 0.995.
The model trained with the smaller DEVICE-TEM dataset converged after 122{,}500 iterations at the final ($1024 \times 1024$) stage with 0.12 iterations per second, resulting in a training time of 7.64 hours at ($64 \times 64$) resolution, 7.78 hours at ($128 \times 128$), 18.94 hours at ($256 \times 256$), 39.20 hours at ($512 \times 512$) and 283.56 hours at the final ($1024 \times 1024$) stage, resulting in a total training time of 357.12 hours. For the last 2,500 iterations, we reduced the occurrence of samples in the training set that were over-represented during sampling. With the larger NANO-TEM dataset, the model converged after 220{,}000 iterations at the final ($1024 \times 1024$) stage with 0.09 iterations per second, resulting in a training time of 20.55 hours at ($64 \times 64$) resolution, 20.25 hours at ($128 \times 128$), 49.91 hours at ($256 \times 256$), 99.81 hours at ($512 \times 512$) and 679.01 hours at the final ($1024 \times 1024$) stage, resulting in a total training time of 869.53 hours. Due to computational constraints, we initially relied on periodic visual inspection of generated samples and subsequent quantitative analysis as detailed in Sec.~\ref{sec:similarity} of the main text. For the larger NANO-TEM model, training could have extended further. However, both visual evaluation and similarity metrics indicated that the generated samples achieved sufficient quality for our application requirements.

\subsubsection{Inference}
We observed an inference duration of 0.16 seconds per DDIM timestep \cite{song_denoising_2020} on an NVIDIA A100 40 GB GPU at a batch-size of 1. A batch-size of up to 6 was achievable, increasing to 13 by using torch.compile.

\subsection{Structural Similarity Metrics}
\label{sec:similarity_appendix}

The Peak Signal-To-Noise Ratio (PSNR) for the original images $f$ and the synthetic images $g$ is defined as:
\begin{equation}
\text{PSNR}(f, g) = 10\times\log_{10} \left( \frac{\text{MAX}^2}{\text{MSE}(f, g)} \right)
\label{eq:psnr}
\end{equation}
where $\text{MAX}$ is the maximum possible pixel value, here $\text{MAX} = 1$ for intensities normalized to $[0,1]$, and MSE is defined as:
\begin{equation}
\text{MSE}(f, g) = \frac{1}{M \times N} \sum_{i=1}^{M} \sum_{j=1}^{N} (f_{ij} - g_{ij})^2
\label{eq:mse}
\end{equation}
where $M \times N$ represents the image. A higher PSNR value indicates similarity to the original image. The Structural Similarity Index Measure (SSIM) \cite{wang_image_2004} is defined as:
\begin{equation}
\text{SSIM}(f, g) = l(f, g) \cdot c(f, g) \cdot s(f, g)
\label{eq:ssim}
\end{equation}

The SSIM compares images regarding their luminance ($l$), contrast ($c$) and structure ($s$), defined as:
\begin{equation}
\begin{aligned}
l(f, g) &= \frac{2\mu_f \mu_g + C_1}{\mu_f^2 + \mu_g^2 + C_1} \\[0.5em]
c(f, g) &= \frac{2\sigma_f \sigma_g + C_2}{\sigma_f^2 + \sigma_g^2 + C_2} \\[0.5em]
s(f, g) &= \frac{\sigma_{fg} + C_3}{\sigma_f \sigma_g + C_3}
\label{eq:lcs}
\end{aligned}
\end{equation}

The individual functions in~(\ref{eq:lcs}) are calculated using the mean luminance $\mu_f$ and $\mu_g$, standard deviation $\sigma_f$ and $\sigma_g$ and covariance $\sigma_{fg}$ of $f$ and $g$. $C_{1\text{--}3}$ are constants stabilizing the denominators. A SSIM of 1 indicates identity, 0 indicates no structural correlation.

The MS-SSIM \cite{wang_multiscale_2003} improves robustness by computing the SSIM across progressively down-sampled and low-pass-filtered scales and is defined as:
\begin{equation}
\text{MS-SSIM}(f, g)=[l_{M}(f, g)]^{\alpha_M}\cdot\prod\limits_{j=1}^{M}[c_{j}(f, g)]^{\beta_{j}}[s_{j}(f,g)]^{\gamma_{j}}
\end{equation}
where the exponents $\alpha_M$, $\beta_j$, and $\gamma_j$ weight the relative importance of each component and scale and are practically set equal within each scale to simplify parameter selection. Luminance $l(f, g)$ is calculated only at the coarsest scale $M$, as comparison is more meaningful at lower resolutions where local pixel variations are smoothed out.

The LPIPS metric \cite{zhang_unreasonable_2018a} leverages features from multiple layers of pre-trained networks (VGG, AlexNet, SqueezeNet) to measure perceptual similarity. It computes weighted L2 distances between channel-wise normalized features:
\begin{equation}
\text{LPIPS}(f, g) = \sum_{l} \frac{1}{H_l W_l} \sum_{h,w} \left\lVert \mathbf{w}_l \odot (\hat{\mathbf{y}}^l_{f,hw} - \hat{\mathbf{y}}^l_{g,hw}) \right\rVert_2^2
\label{eq:lpips}
\end{equation}
where $\mathbf{w}_l$ represents learned linear weights for layer $l$, calibrated on human perceptual judgments, $\hat{\mathbf{y}}^l$ indicates channel-wise unit normalized feature maps with spatial dimensions $(H_l, W_l)$, and $\odot$ denotes the Hadamard product. Smaller LPIPS values indicate higher perceptual similarity.

\subsection{Noise Estimation Metrics -- Formulas}
\label{sec:noise_appendix}

We estimated noise standard deviation using the Laplacian operator method \cite{immerkaer_fast_1996}, where an image $f$ with dimensions $H \times W$ and pixel intensities $f_{ij}$, normalized to $[0,1]$, is convolved with the discrete Laplacian kernel:
\begin{equation}
L = \begin{bmatrix} 0 & -1 & 0\\ -1 & 4 & -1\\ 0 & -1 & 0 \end{bmatrix}
\label{eq:laplacian_kernel}
\end{equation}
and the result is denoted as $y = f * L$. The noise standard deviation is then estimated as:
\begin{equation}
\sigma_{n}(f) = \sqrt{ c \cdot \frac{1}{HW} \sum_{i=1}^{H} \sum_{j=1}^{W} y_{ij}^{2} }
\label{eq:noise_std}
\end{equation}
where $c$ describes a correction factor accounting for the discrete nature of the Laplacian operator. Higher values of $\sigma_n$ indicate more noise. In our implementation $c = \pi/72$, and $y$ is computed with zero padding at the image border. Because $c$ enters as a global scale factor, it affects all reported noise standard deviations identically (and shifts all SNR values by the same constant offset in dB) and does not influence the comparisons between models. We follow the Laplacian-based approach of \cite{immerkaer_fast_1996}; the kernel of Eq.~(\ref{eq:laplacian_kernel}) and the squared-magnitude form above differ from the estimator in that reference, so absolute values are specific to this implementation.

The Signal-to-Noise Ratio (SNR) quantifies the ratio of signal strength and noise. It is computed as:
\begin{equation}
\text{SNR}_{\text{dB}}(f) = 20 \log_{10}\left( \frac{\mu_f}{\sigma_{n}(f)} \right)
\label{eq:snr}
\end{equation}
where $\mu_f$ is the mean intensity of the image. Higher SNR values indicate better signal quality in relation to noise levels. Because $\mu_f$ is the image mean, the SNR defined here also reflects differences in mean brightness between models. We therefore report it alongside, rather than in place of, the noise standard deviation.

We define the High-Frequency Noise Ratio (HFNR) as the ratio of the mean Fourier magnitude in the high-frequency region to the mean Fourier magnitude over all frequencies:
\begin{equation}
\text{HFNR}(f) = \frac{\text{mean}(|F_{uv}| \text{ high frequencies})}{\text{mean}(|F_{uv}| \text{ all frequencies})}
\label{eq:hfnr}
\end{equation}
Here, $F$ represents a 2D discrete Fourier transform of $f$ and $|F_{uv}|$ denotes the magnitude of the Fourier coefficient at frequency coordinates $(u,v)$. High frequencies are defined as those outside a centered disk of radius $\lfloor\min(H,W)/4\rfloor$. We compute Eq.~(\ref{eq:hfnr}) identically for original and synthetic images and use it only as a relative measure of spectral concentration. Its absolute scale is set by the chosen cut-off radius and by the frequency partition of the implementation, so individual values should not be read as an absolute high-frequency fraction; only differences between models, and distances from the original, are interpreted.

\subsection{Patch-based Training and Layer Freezing Ablation Study}
\label{sec:ablation_patches}

\begin{table}[H]
\caption{Ablation study of the proposed patch-based DDPM training with (ours) and without layer-freezing against non-patch-based baseline DDPM training after 122,500 iterations on the DEVICE-TEM dataset. 600 samples were created using 120 DDIM timesteps and an $\eta$-value of 0.5. Structural similarity and perceptual metrics were computed against every original image. Only per-metric best results, indicating an optimal match, were recorded. We report mean results $\pm$ standard deviation; bold: best model per metric, underlined: second best.}
\label{tab:similarity_metrics_device_ddpm_variants}
\centering
\resizebox{\columnwidth}{!}{%
\begin{tabular}{cccc}
\toprule
\textbf{Metric} &
\makecell{\textbf{DDPM}\\\textbf{(Ours)}} &
\makecell{\textbf{DDPM}\\\textbf{(Baseline)}} &
\makecell{\textbf{DDPM}\\\textbf{(No freeze)}} \\
\midrule
MS-SSIM $(\uparrow)$ & \bm{$0.621 \pm 0.191$} & \underline{$0.581 \pm 0.181$} & $0.577 \pm 0.170$ \\
SSIM $(\uparrow)$    & \bm{$0.561 \pm 0.175$} & $0.509 \pm 0.159$ & \underline{$0.523 \pm 0.161$} \\
PSNR (dB) $(\uparrow)$ & \bm{$18.96 \pm 3.88$} & $18.35 \pm 3.16$ & \underline{$18.54 \pm 2.91$} \\
MSE $(\downarrow)$   & \underline{$0.017 \pm 0.010$} & $0.018 \pm 0.009$ & \bm{$0.016 \pm 0.008$} \\
LPIPS $(\downarrow)$ & \bm{$0.331 \pm 0.158$} & $0.367 \pm 0.149$ & \underline{$0.365 \pm 0.137$} \\
FID $(\downarrow)$   & $62.93$ & \bm{$58.31$} & \underline{$61.93$} \\
KID $(\to 0)$        & \bm{$-0.003 \pm 0.013$} & \underline{$-0.008 \pm 0.011$} & $-0.009 \pm 0.009$ \\
\bottomrule
\end{tabular}}
\end{table}

\begin{table}[H]
\caption{Ablation study noise metrics, same methodology as in Tab.~\ref{tab:similarity_metrics_device_ddpm_variants}; bold: closest to the originals, underlined: second closest.}
\label{tab:noise_comparison_ddpm_variants_with_original}
\centering
\resizebox{\columnwidth}{!}{%
\begin{tabular}{ccccc}
\toprule
\textbf{Metric} & \textbf{Original} &
\makecell{\textbf{DDPM}\\\textbf{(Ours)}} &
\makecell{\textbf{DDPM}\\\textbf{(Baseline)}} &
\makecell{\textbf{DDPM}\\\textbf{(No freeze)}} \\
\midrule
Noise std $(=)$ & $0.022 \pm 0.011$ & \underline{$0.016 \pm 0.009$} & \bm{$0.017 \pm 0.009$} & \underline{$0.016 \pm 0.009$} \\
SNR (dB) $(=)$  & $23.78 \pm 2.66$ & \underline{$26.81 \pm 3.50$} & $27.19 \pm 3.15$ & \bm{$26.71 \pm 3.02$} \\
HFNR $(=)$      & $1.141 \pm 0.031$ & \underline{$1.147 \pm 0.022$} & \bm{$1.146 \pm 0.024$} & $1.153 \pm 0.023$ \\
\bottomrule
\end{tabular}}
\end{table}

\begin{figure*}[!t]
    \centering
    \subfloat[MS-SSIM $(\uparrow)$]{\includegraphics[width=0.48\textwidth]{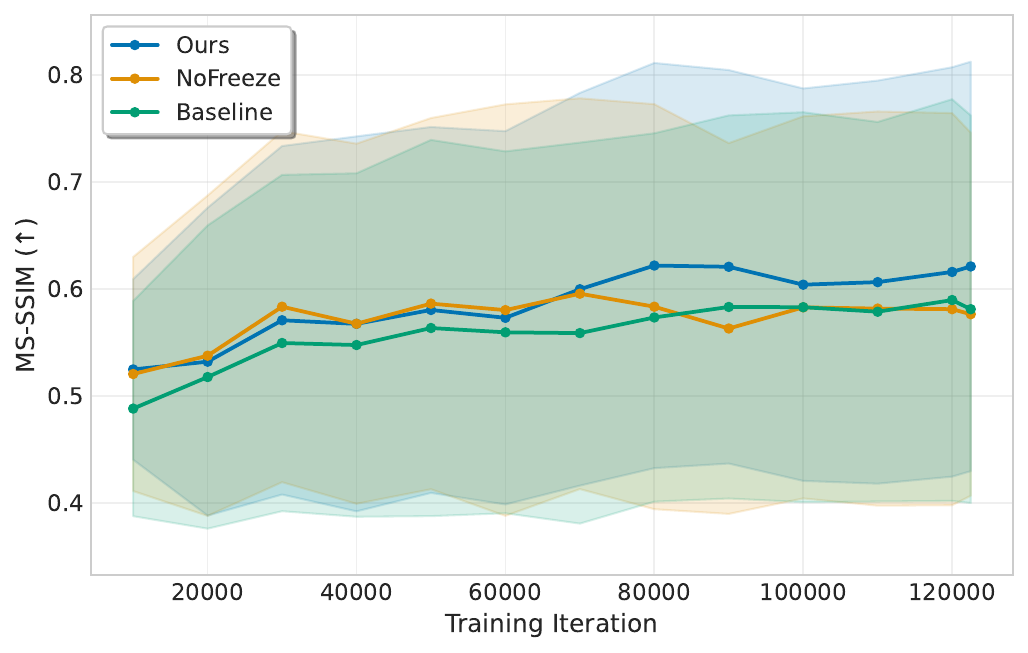}}
    \hfill
    \subfloat[SSIM $(\uparrow)$]{\includegraphics[width=0.48\textwidth]{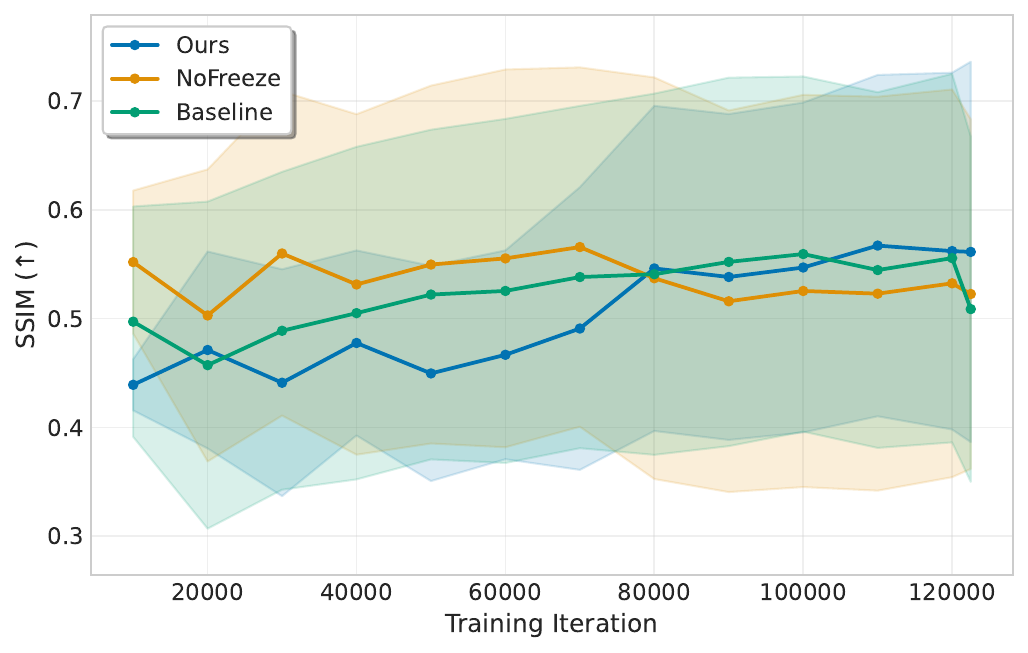}}

    \vspace{0.3em}

    \subfloat[PSNR (dB) $(\uparrow)$]{\includegraphics[width=0.48\textwidth]{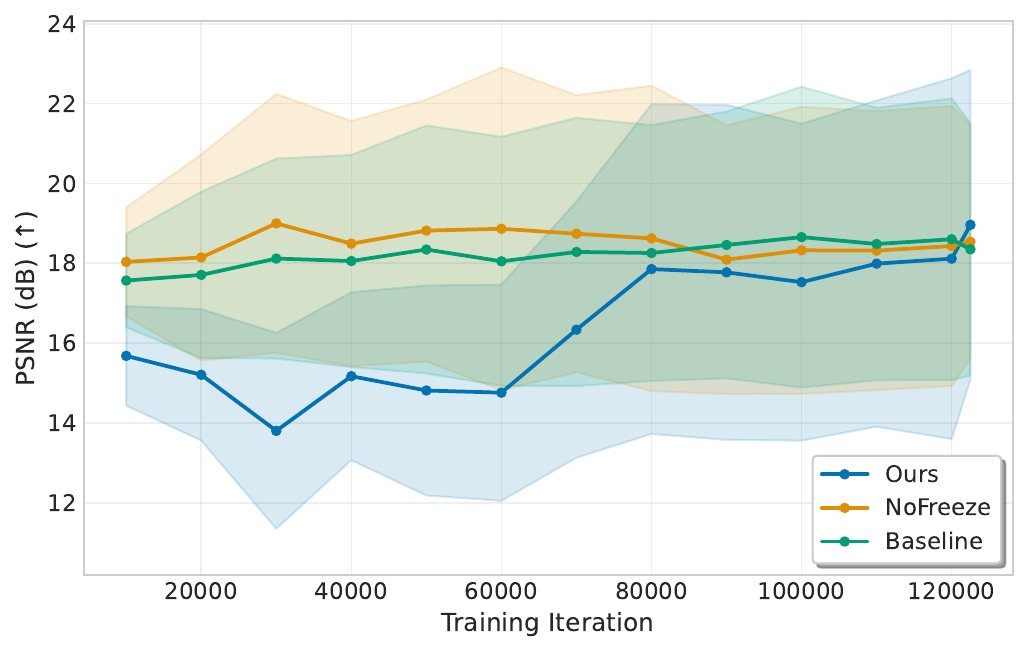}}
    \hfill
    \subfloat[MSE $(\downarrow)$]{\includegraphics[width=0.48\textwidth]{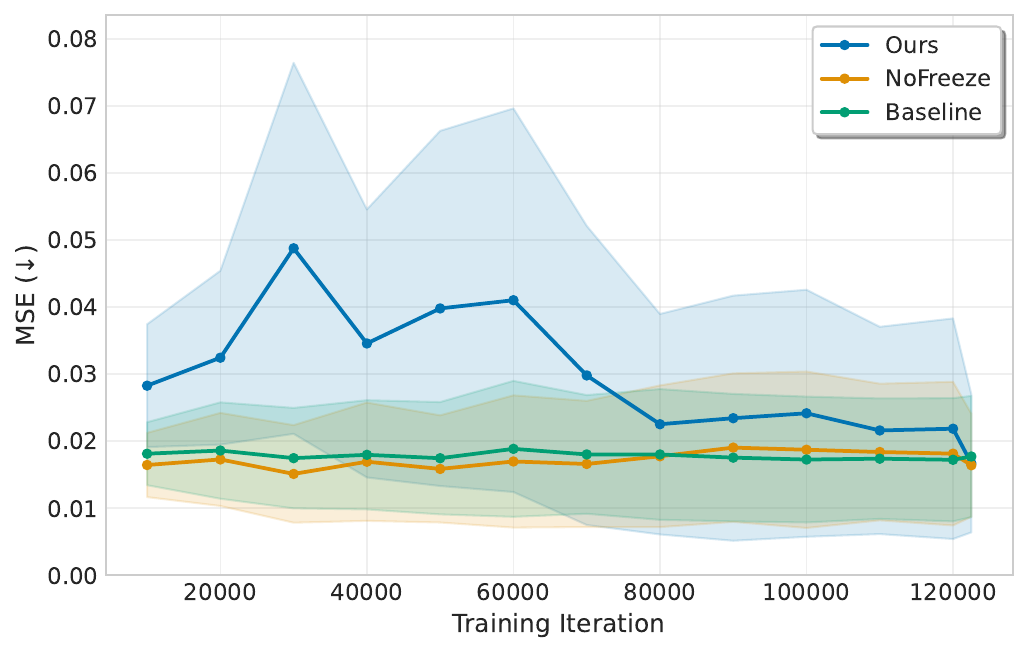}}

    \vspace{0.3em}

    \subfloat[LPIPS $(\downarrow)$]{\includegraphics[width=0.48\textwidth]{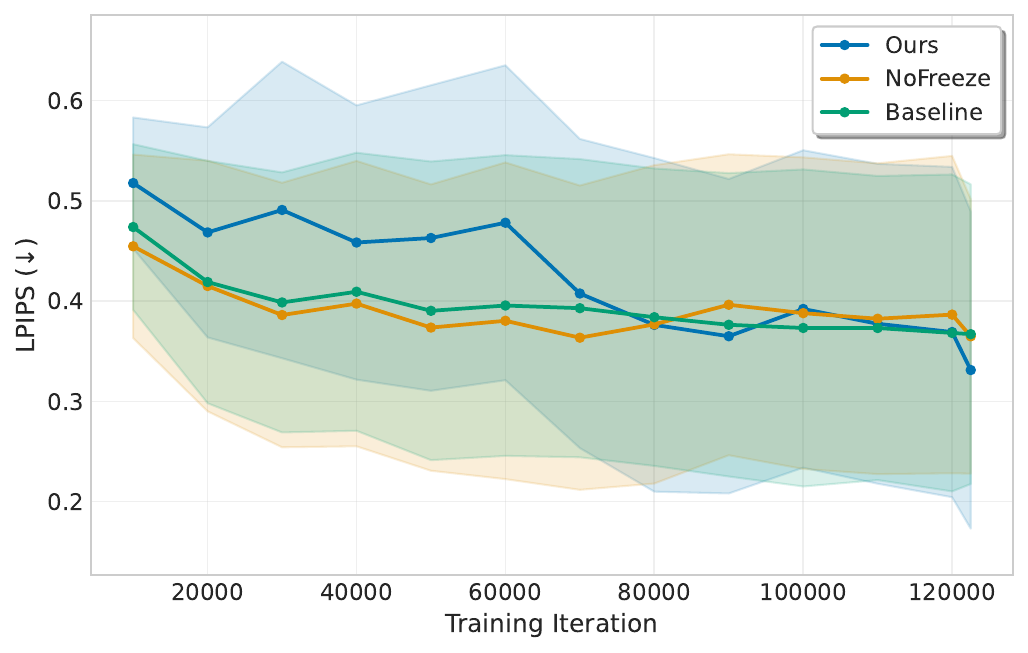}}
    \hfill
    \subfloat[FID $(\downarrow)$]{\includegraphics[width=0.48\textwidth]{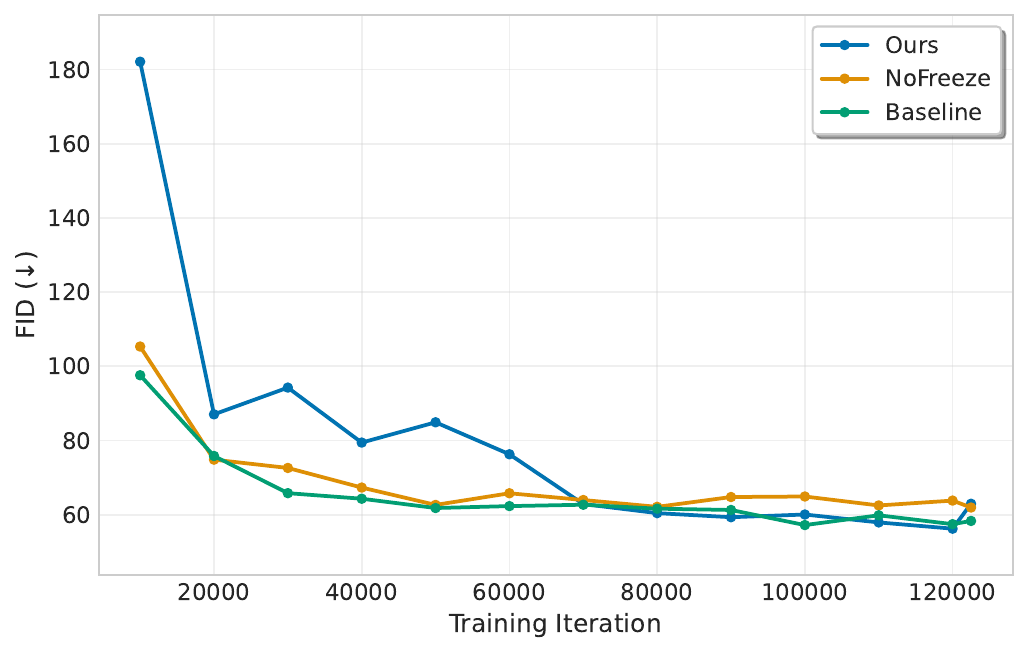}}

    \caption{Per training run, every 10,000 iterations, the same methodology as reported in Tab.~\ref{tab:similarity_metrics_device_ddpm_variants} was applied (part 1).}
    \label{fig:ablation-metrics}
\end{figure*}

\begin{figure*}[!t]
    \centering
    \subfloat[KID $(\to 0)$]{\includegraphics[width=0.48\textwidth]{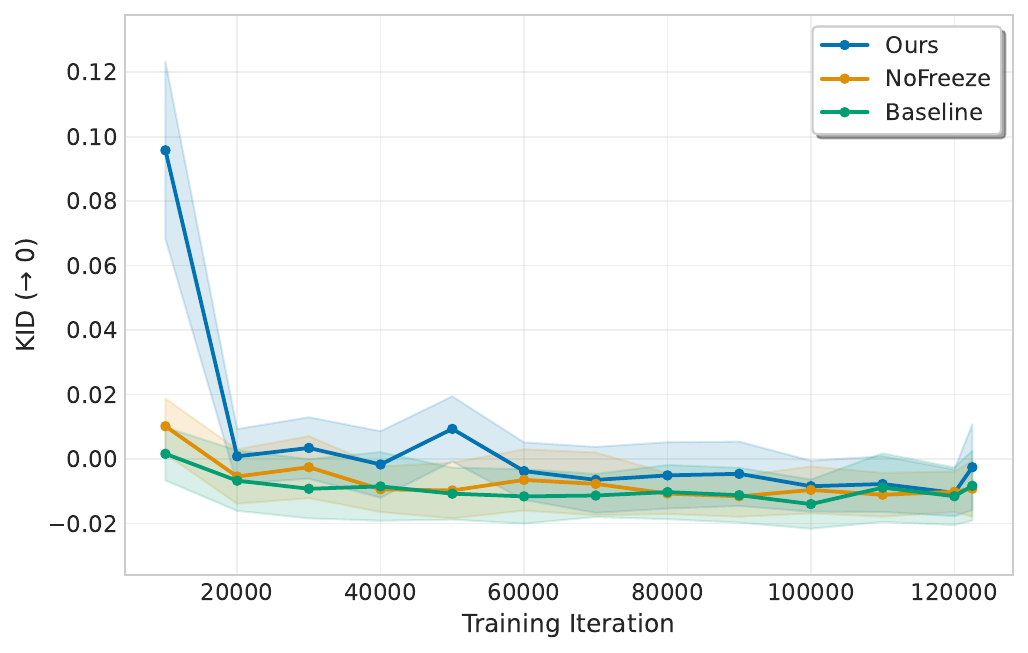}}
    \hfill
    \subfloat[Noise Std $(=)$]{\includegraphics[width=0.48\textwidth]{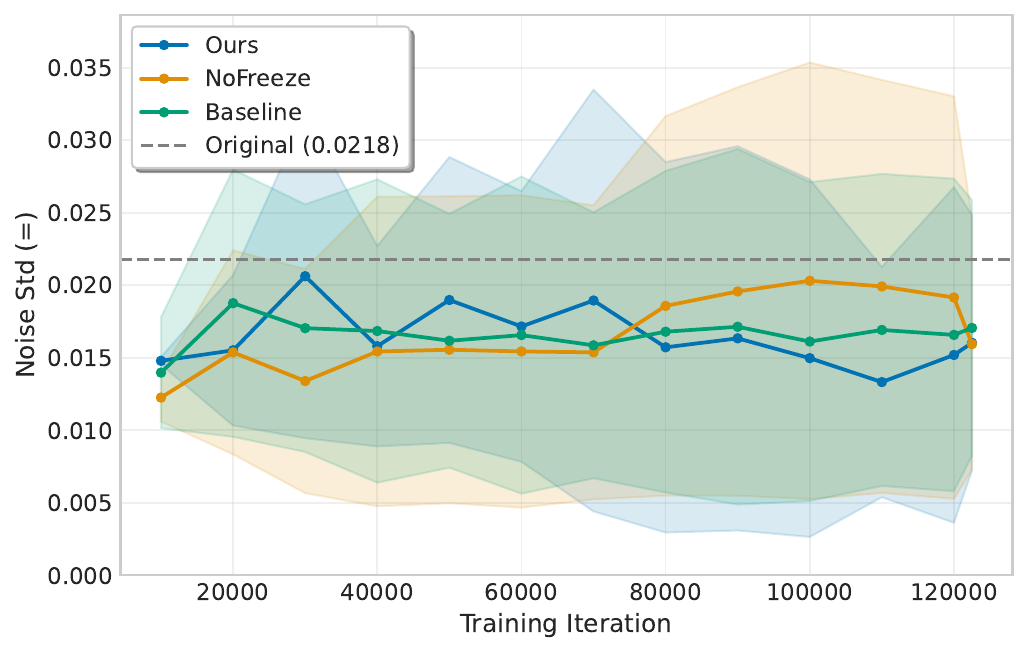}}

    \vspace{0.3em}

    \subfloat[SNR (dB) $(=)$]{\includegraphics[width=0.48\textwidth]{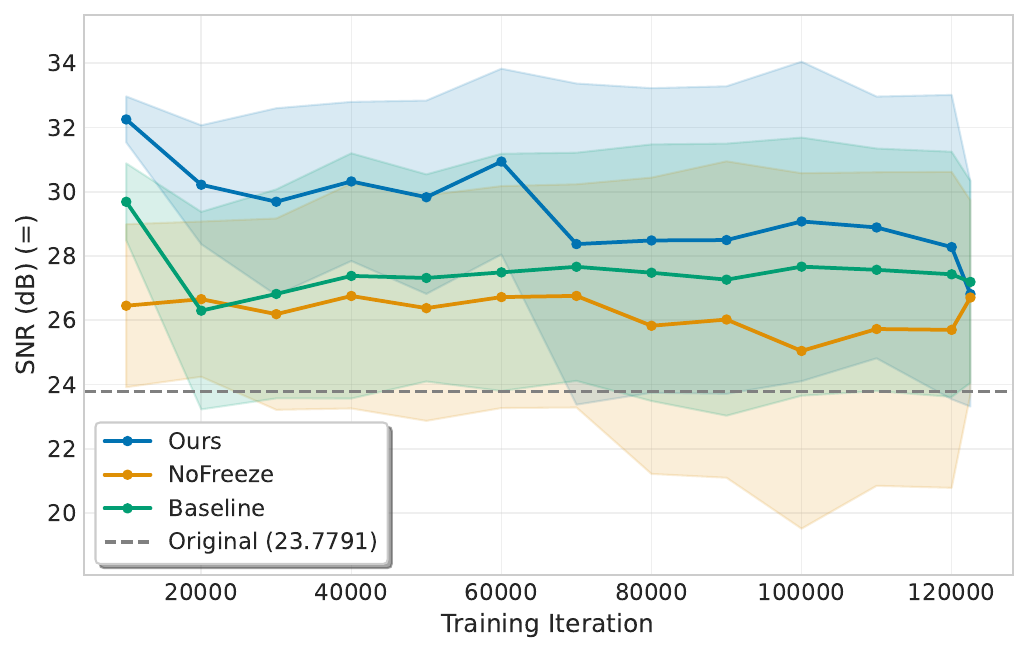}}
    \hfill
    \subfloat[HFNR $(=)$]{\includegraphics[width=0.48\textwidth]{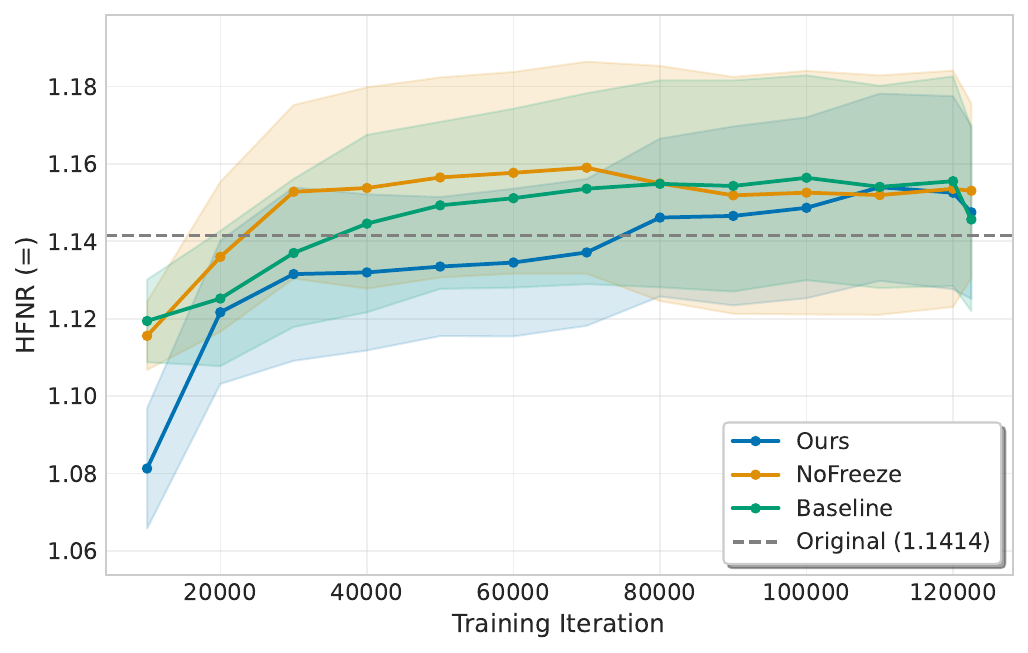}}

    \caption{Per training run, every 10,000 iterations, the same methodology as reported in Tab.~\ref{tab:similarity_metrics_device_ddpm_variants} was applied (part 2).}
\end{figure*}
\FloatBarrier

\onecolumn
\subsection{Classifier Guidance UMAP}
\label{sec:classifier_guidance_umap}
\begin{figure}[H]
    \centering
    \includegraphics[width=\textwidth]{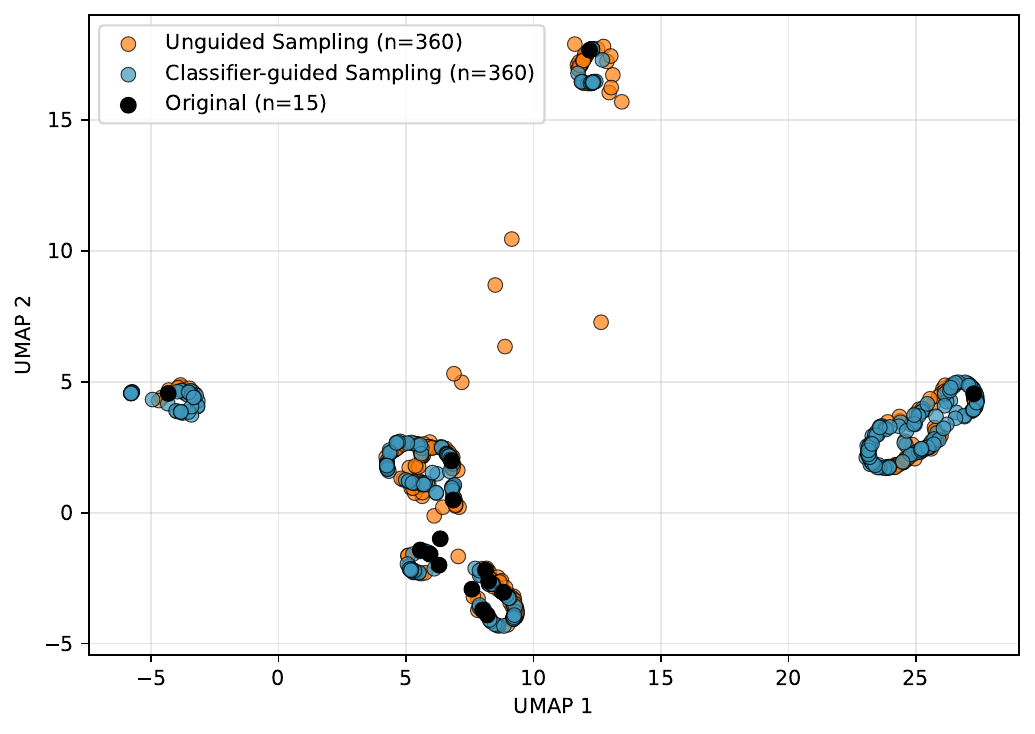}
    \caption{UMAP \cite{mcinnes_umap_2020} of DEVICE-TEM samples generated without guidance and with guidance scale ($s \in [0.5, 10]$, $\Delta s = 0.5$) with 15 samples for each scale.}
    \label{fig:umap}
\end{figure}

Fig.~\ref{fig:umap} shows a UMAP projection of DEVICE-TEM samples generated with and without classifier guidance.
\FloatBarrier

\begin{figure*}[!p]
\centering

\subfloat[]{%
\includegraphics[width=0.25\textwidth]{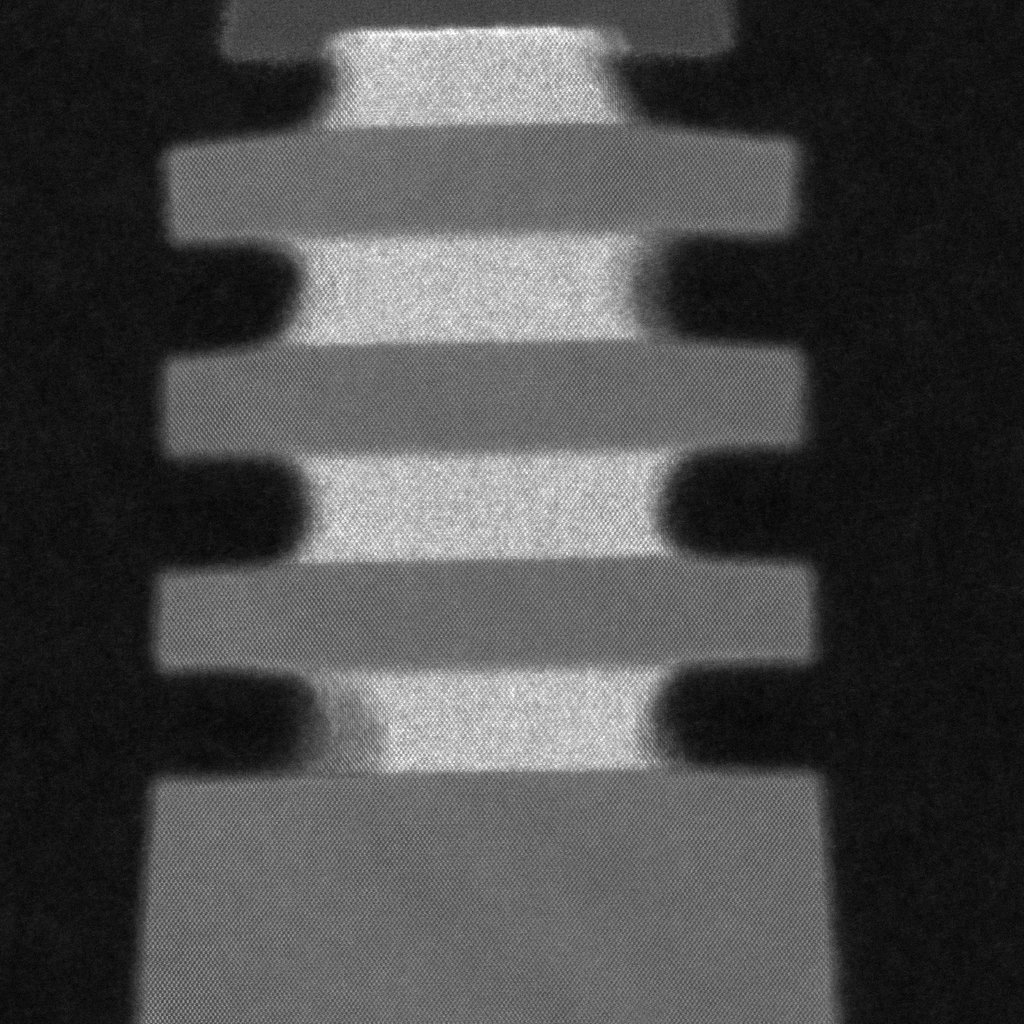}}
\quad
\subfloat[]{%
\includegraphics[width=0.25\textwidth]{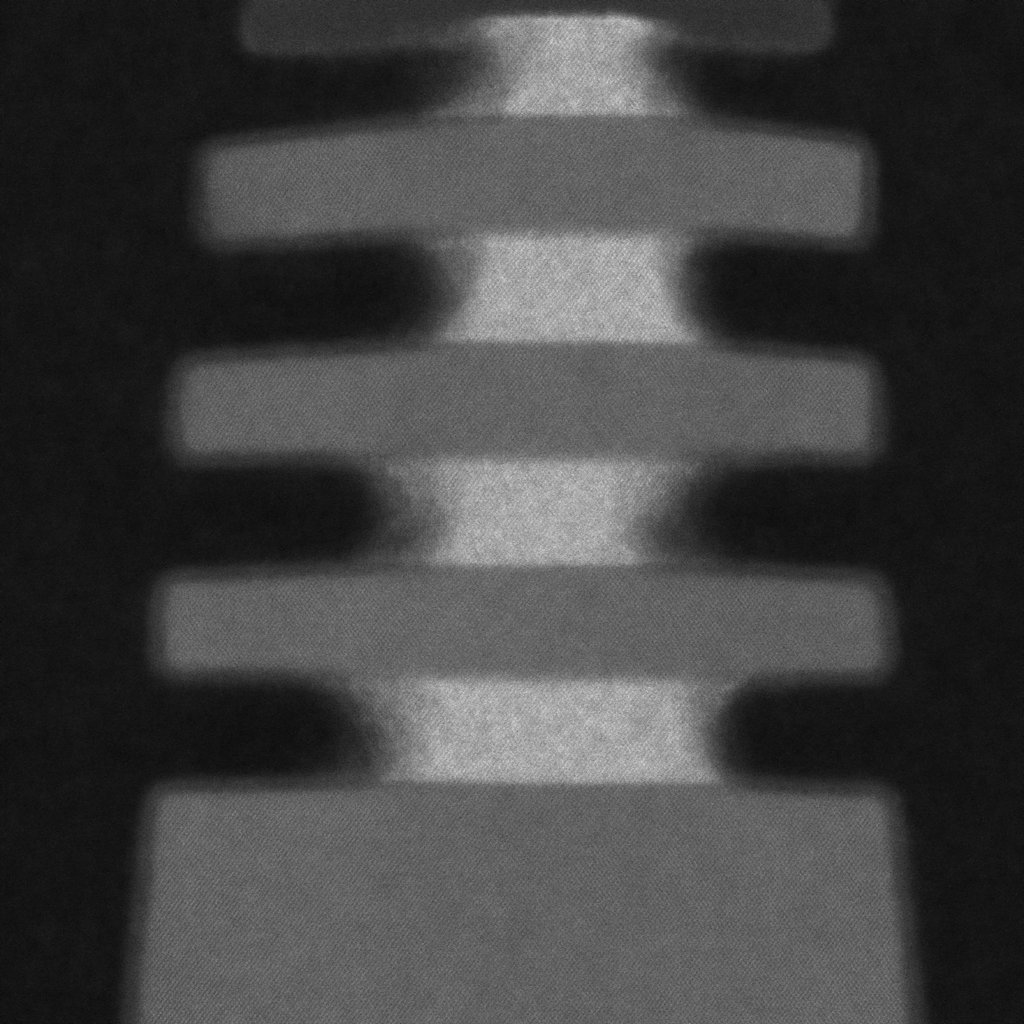}}
\quad
\subfloat[]{%
\includegraphics[width=0.25\textwidth]{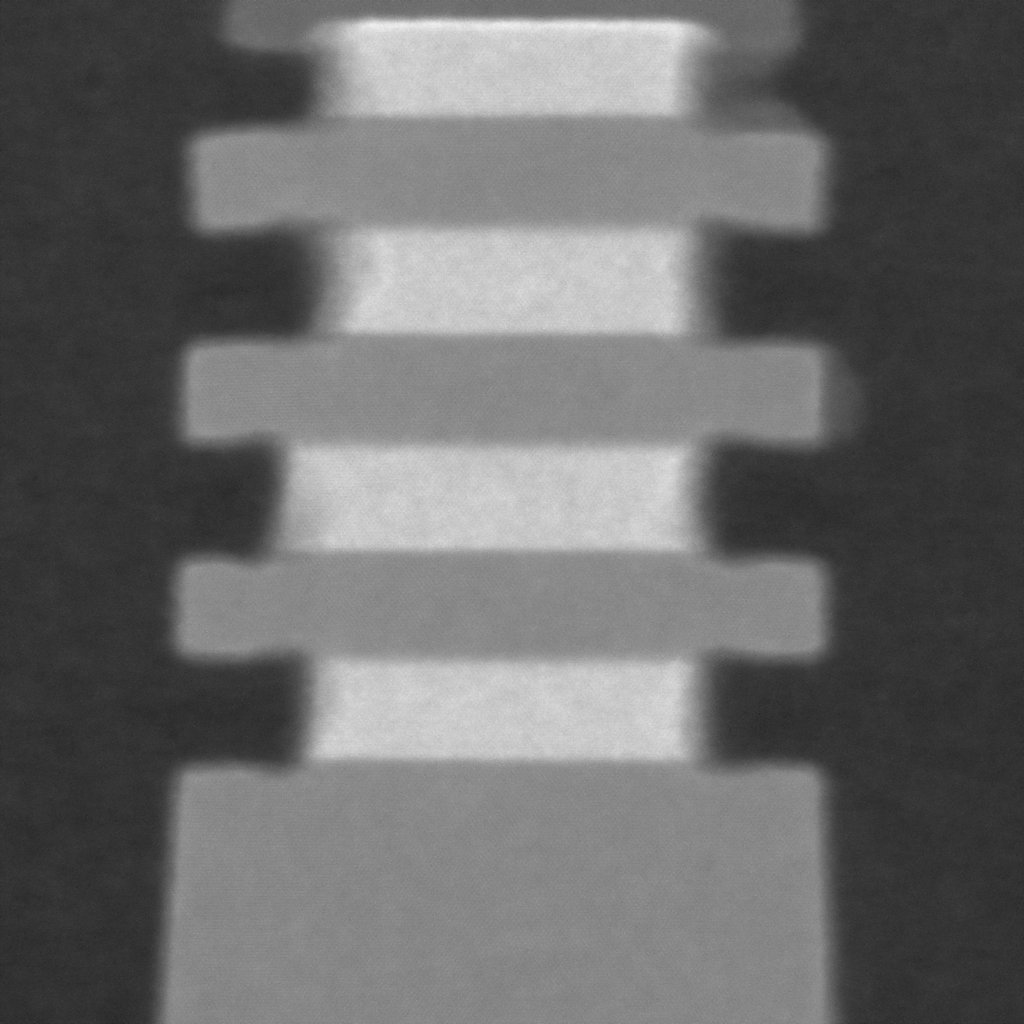}}

\vspace{0.3em}

\subfloat[]{%
\includegraphics[width=0.25\textwidth]{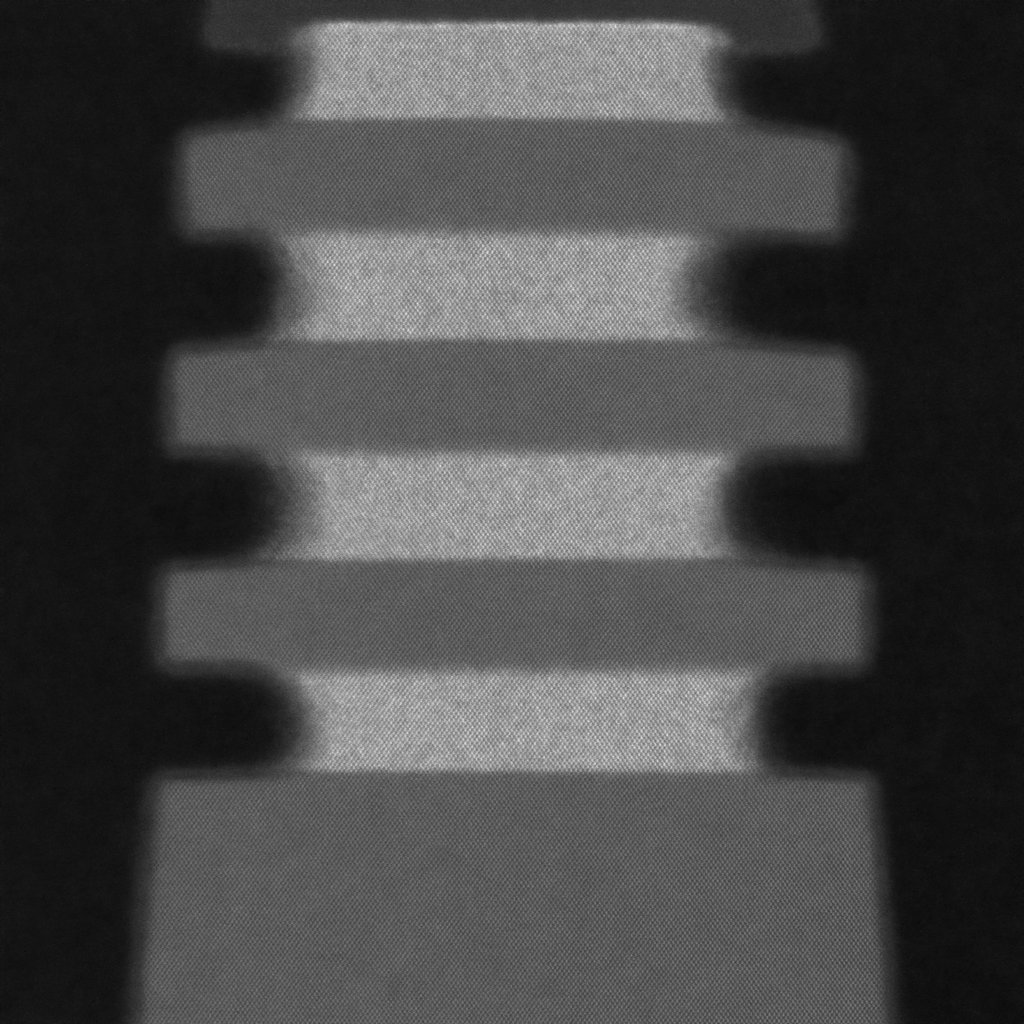}}
\quad
\subfloat[]{%
\includegraphics[width=0.25\textwidth]{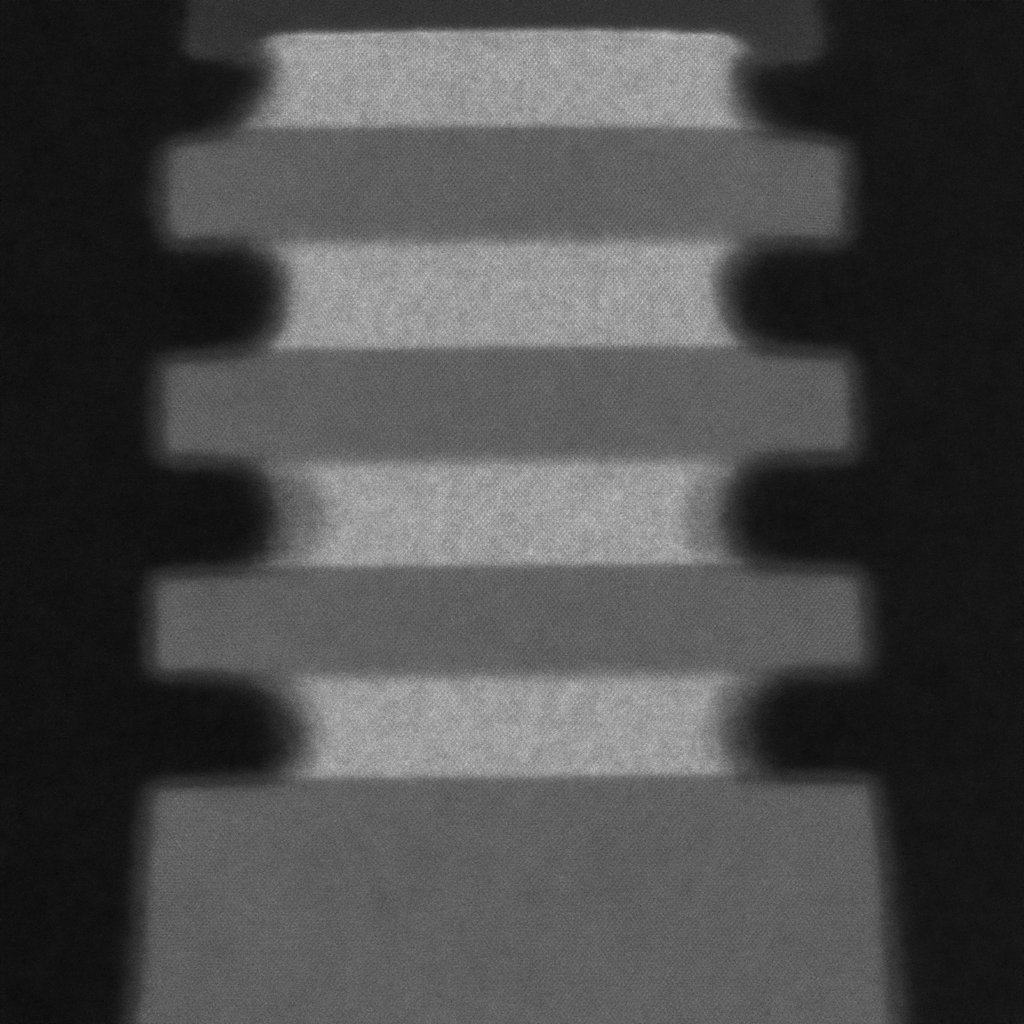}}
\quad
\subfloat[]{%
\includegraphics[width=0.25\textwidth]{{plots/syn_nano6}.jpg}}

\vspace{0.6em}

\subfloat[]{%
\includegraphics[width=0.25\textwidth]{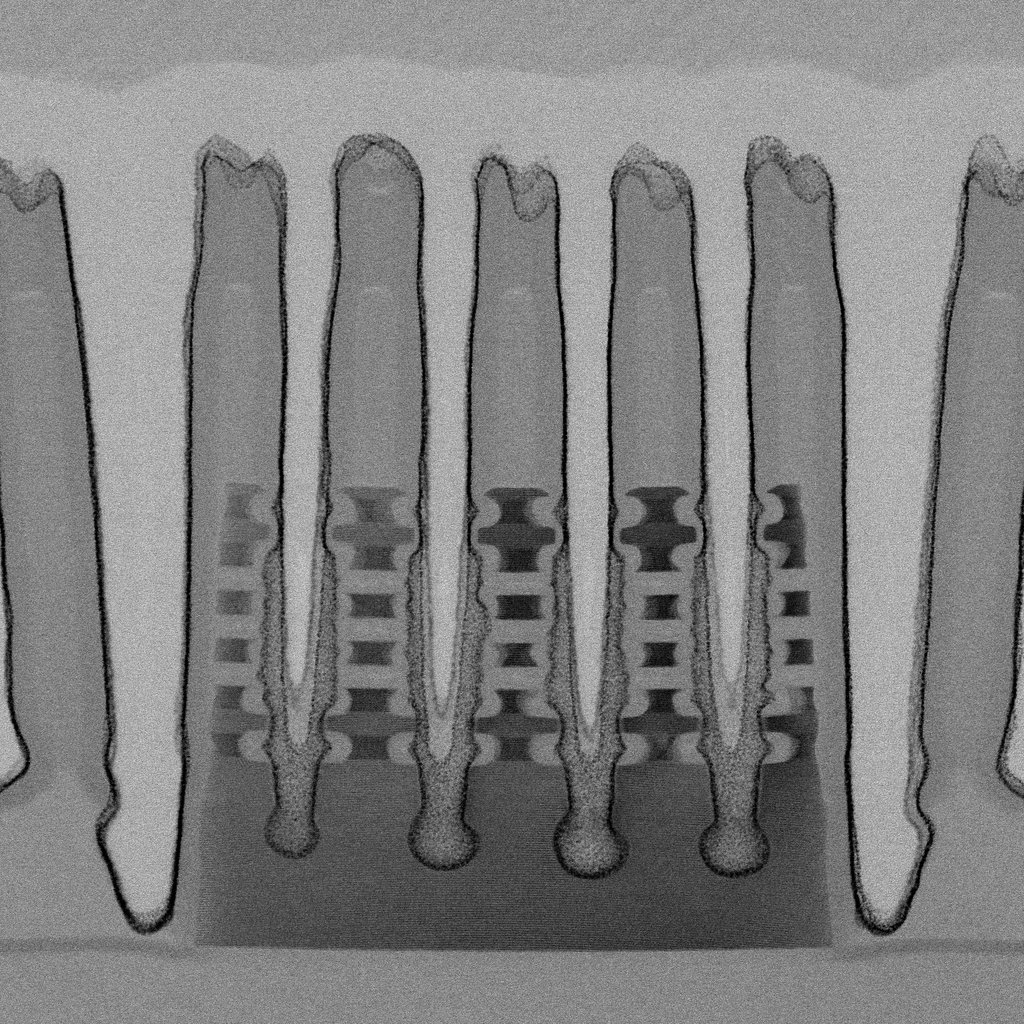}}
\quad
\subfloat[]{%
\includegraphics[width=0.25\textwidth]{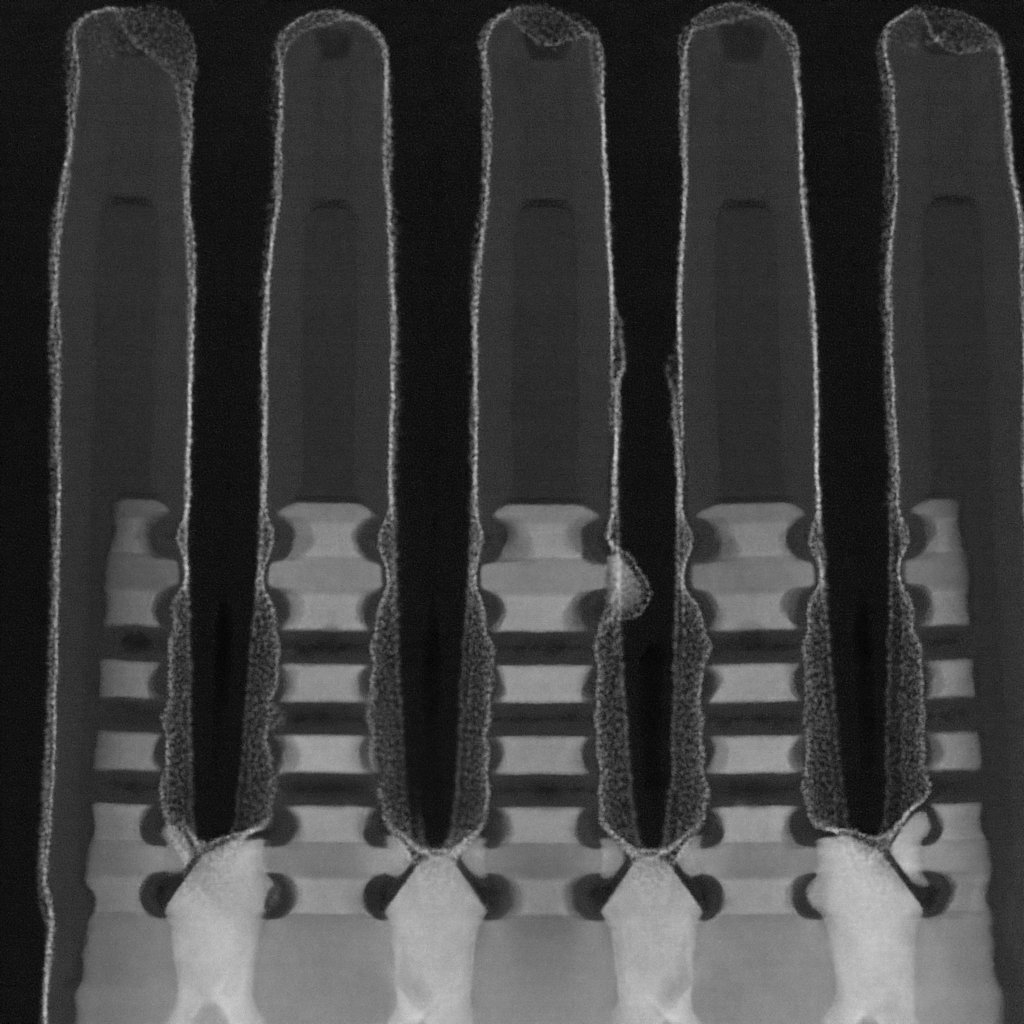}}
\quad
\subfloat[]{%
\includegraphics[width=0.25\textwidth]{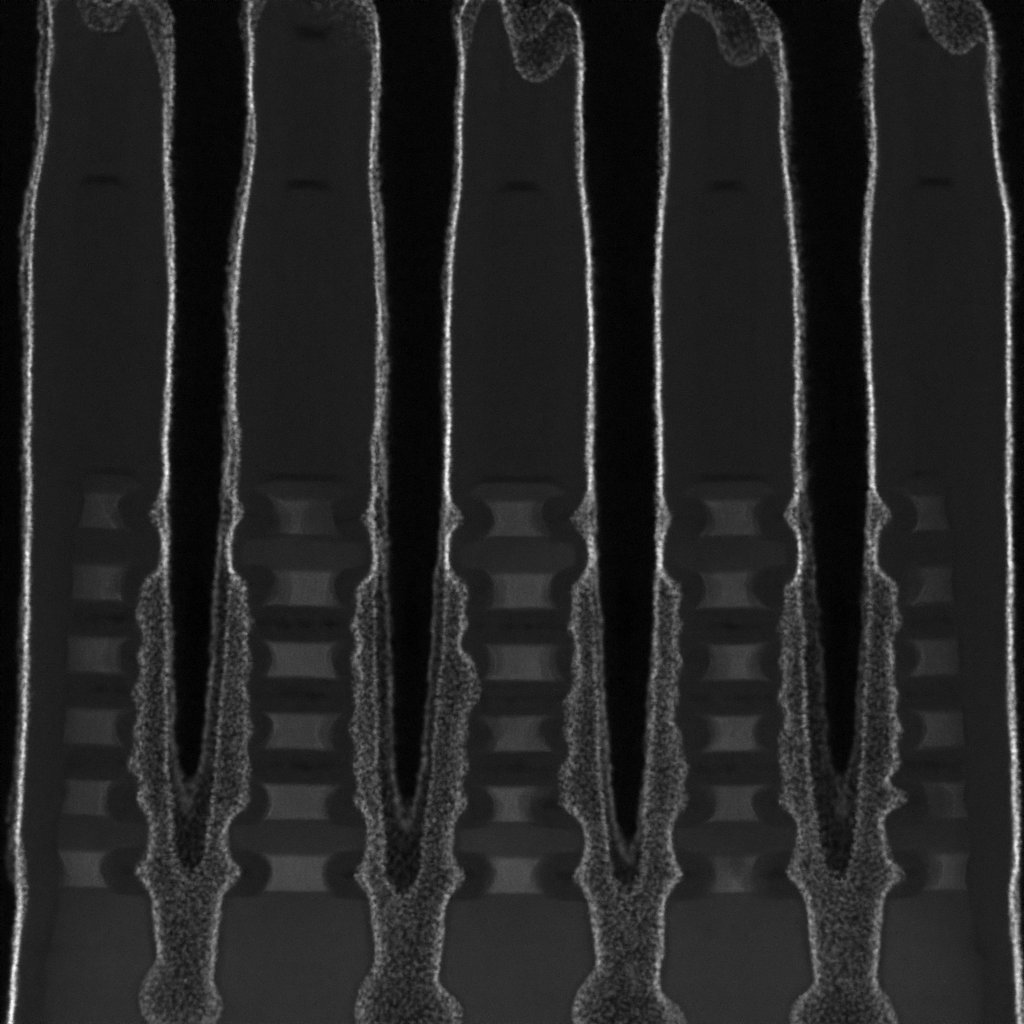}}

\vspace{0.3em}

\subfloat[]{%
\includegraphics[width=0.25\textwidth]{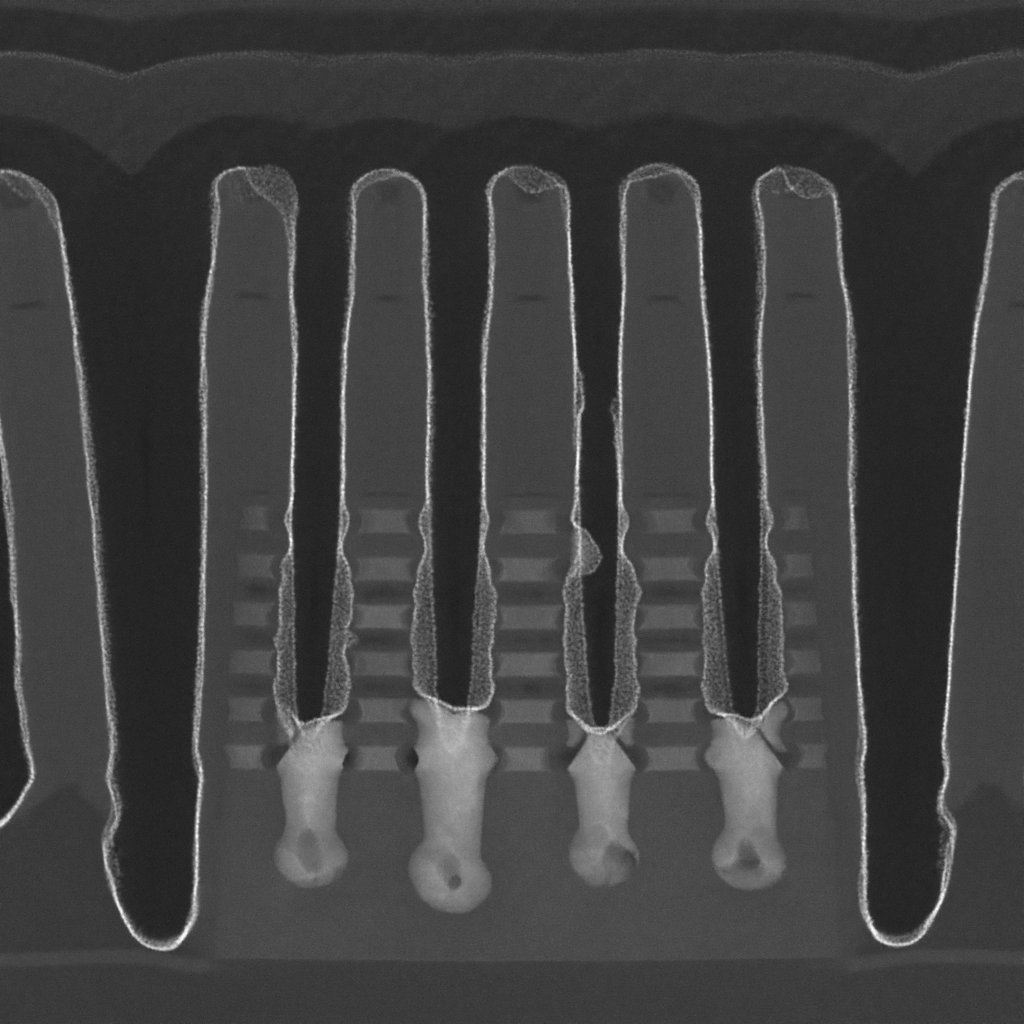}}
\quad
\subfloat[]{%
\includegraphics[width=0.25\textwidth]{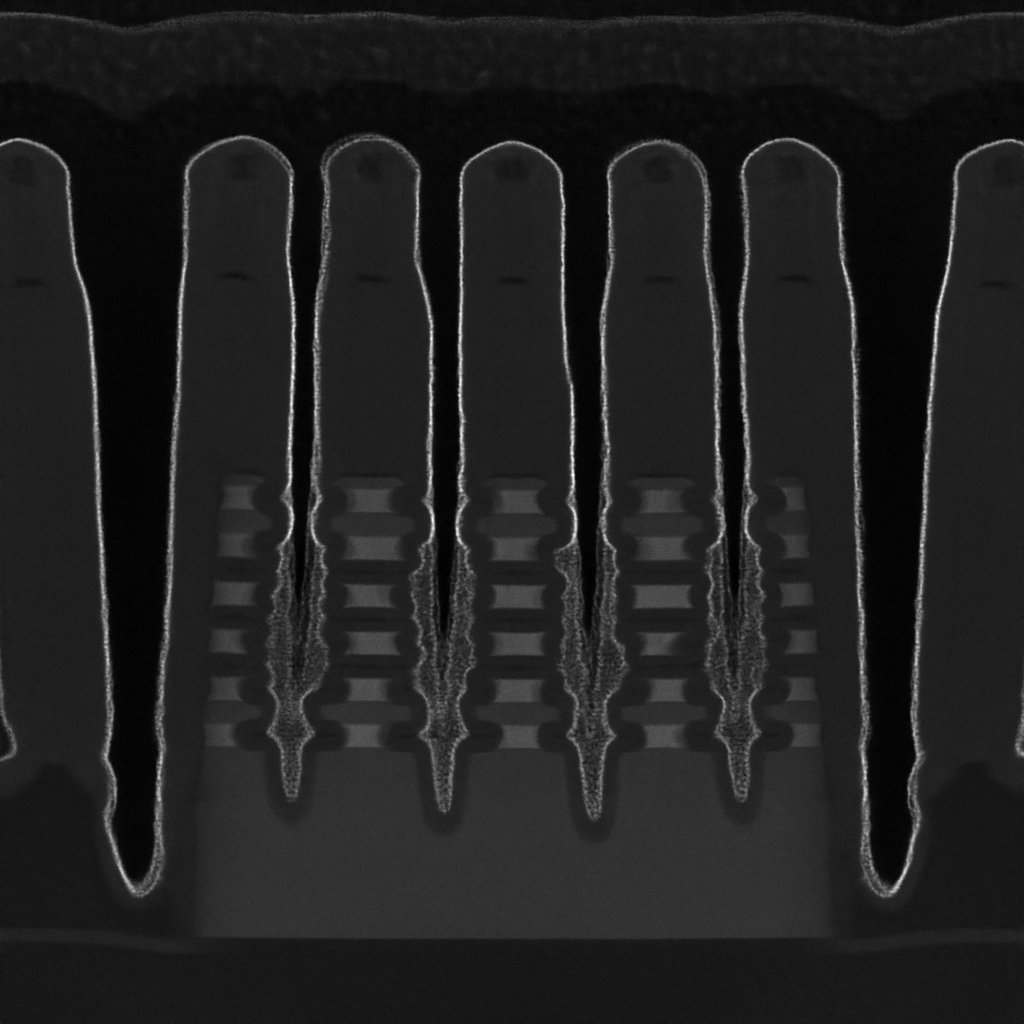}}
\quad
\subfloat[]{%
\includegraphics[width=0.25\textwidth]{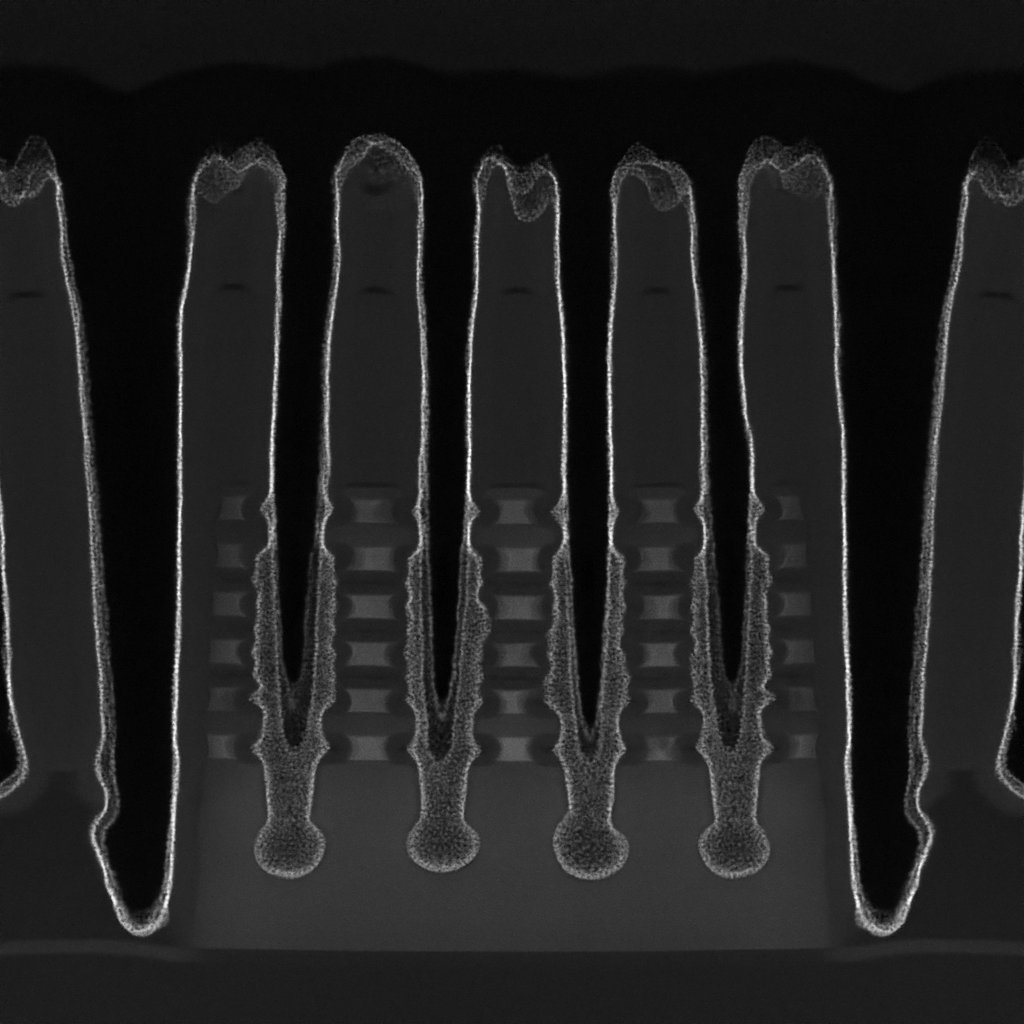}}

\caption{Original NANO-TEM (a) and DEVICE-TEM (g) images with synthetic samples (b--f) and (h--l), respectively.}
\label{fig:example_grid}
\end{figure*}

%% file: figs/diffusion_framework.tex
\begin{tikzpicture}[
  every node/.style={font=\small},
  line cap=round, line join=round,
  >=Stealth,
  >={Stealth[length=2.0mm,width=1.6mm]} 
]
  \tikzset{
    state/.style={circle, draw, minimum size=9.5mm, inner sep=0pt, outer sep=0pt, line width=0.8pt},
    filledstate/.style={state, fill=gray!18},
    lab/.style={font=\scriptsize},
    img/.style={inner sep=0pt, outer sep=0pt, anchor=north west}
  }
  \coordinate (c0) at (0,0);
  \coordinate (c1) at (2.9,0);
  \coordinate (c2) at (5.8,0);
  \coordinate (c3) at (8.7,0);
  \node[filledstate] (xT)   at (c0) {$\mathbf{x}_T$};
  \node[filledstate] (xt)   at (c1) {$\mathbf{x}_t$};
  \node[filledstate] (xtm1) at (c2) {$\mathbf{x}_{t-1}$};
  \node[state]       (x0)   at (c3) {$\mathbf{x}_0$};
  
  \draw[->, line width=0.9pt, shorten >=0.4pt, shorten <=0.4pt] (xT.east) -- (xt.west);
  \draw[->, line width=0.9pt, shorten >=0.4pt, shorten <=0.4pt]
    (xt.east) -- node[above=0.35mm, lab] {$p_{\theta}(\mathbf{x}_{t-1}\!\mid\!\mathbf{x}_t)$} (xtm1.west);
  \draw[->, line width=0.9pt, shorten >=0.4pt, shorten <=0.4pt] (xtm1.east) -- (x0.west);
  
  \path[->, dashed, line width=0.8pt, dash phase=1pt, shorten >=0.6pt, shorten <=0.6pt]
    (xtm1) edge[out=225, in=-45, looseness=0.86]
    node[midway, below=0.35mm, lab] {$q(\mathbf{x}_t\!\mid\!\mathbf{x}_{t-1})$}
    (xt);
  
  \node[img] at ([xshift=0.9mm,yshift=-0.2mm]xT.south east)
    {\includegraphics[width=0.90cm]{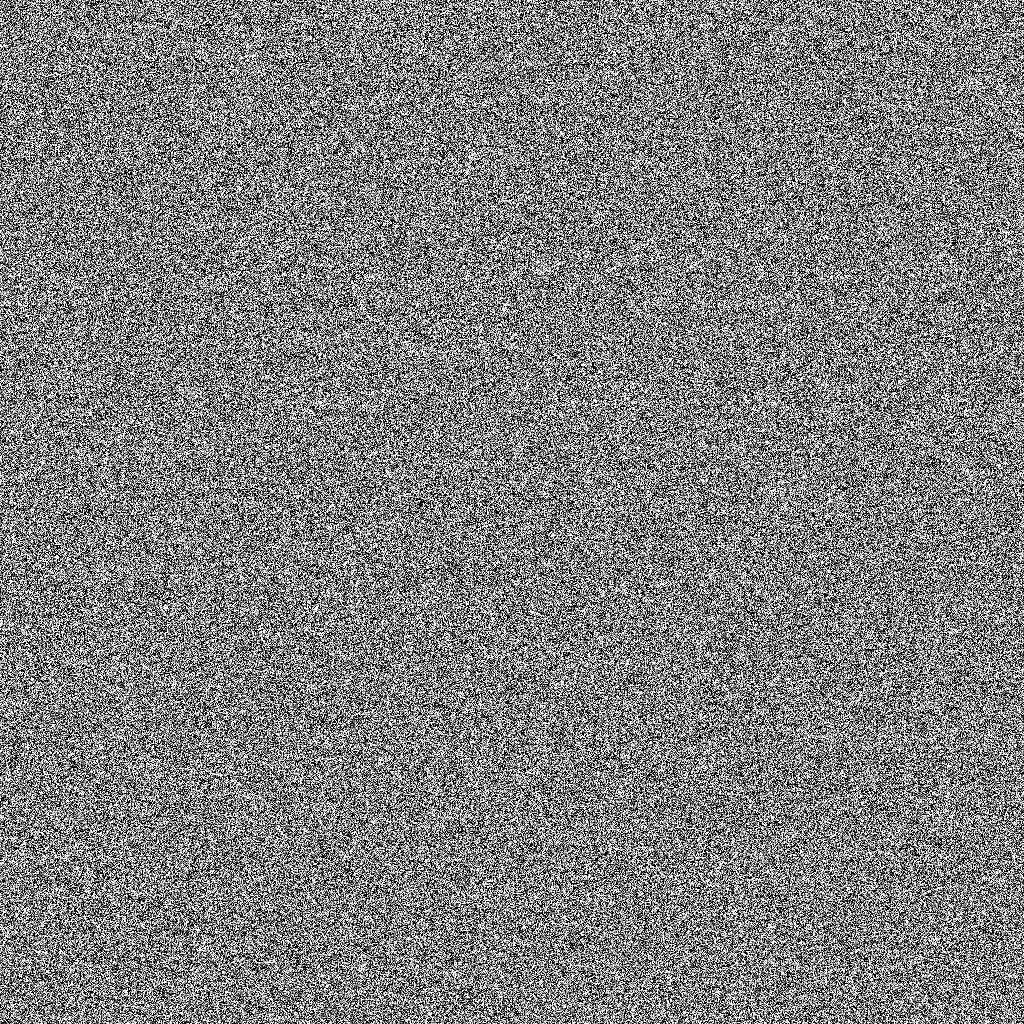}};
  \node[img] at ([xshift=0.9mm,yshift=-0.2mm]xtm1.south east)
    {\includegraphics[width=0.90cm]{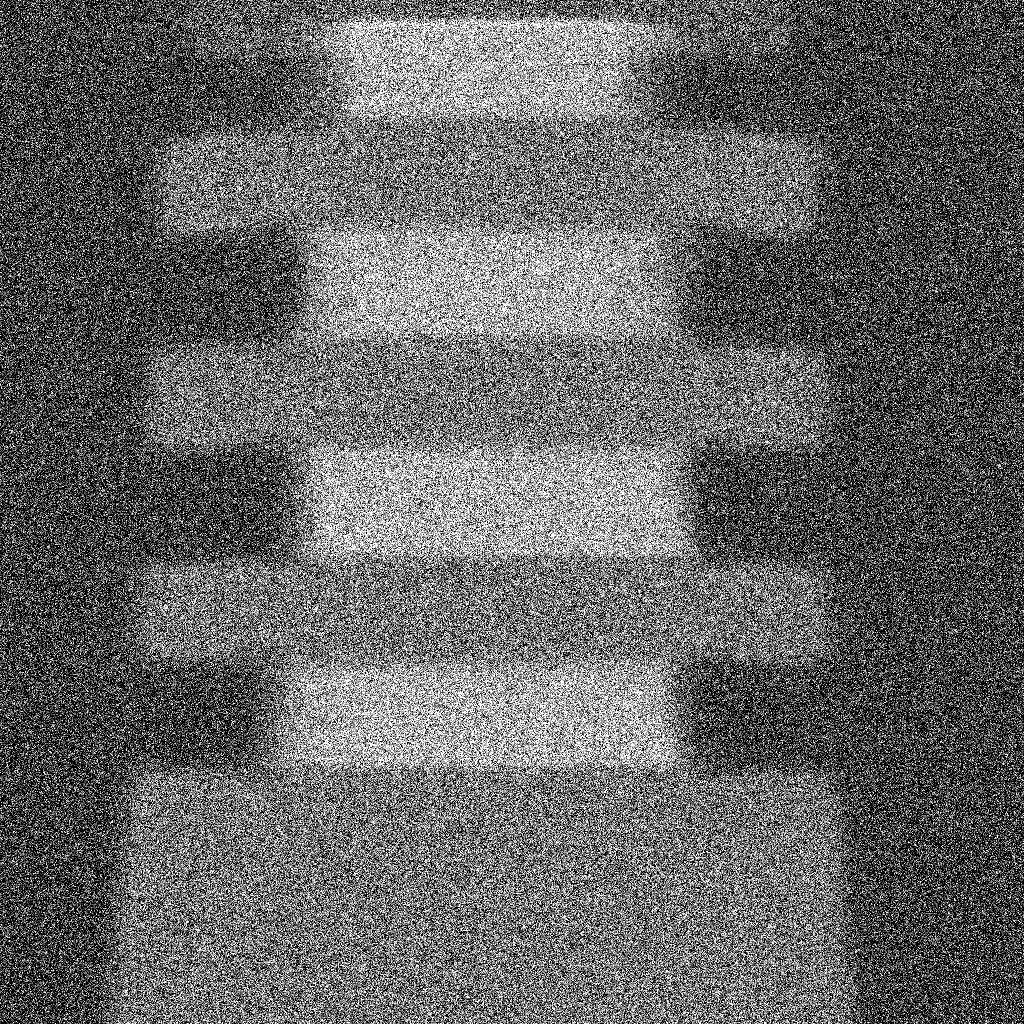}};
  \node[img] at ([xshift=0.9mm,yshift=-0.2mm]x0.south east)
    {\includegraphics[width=0.90cm]{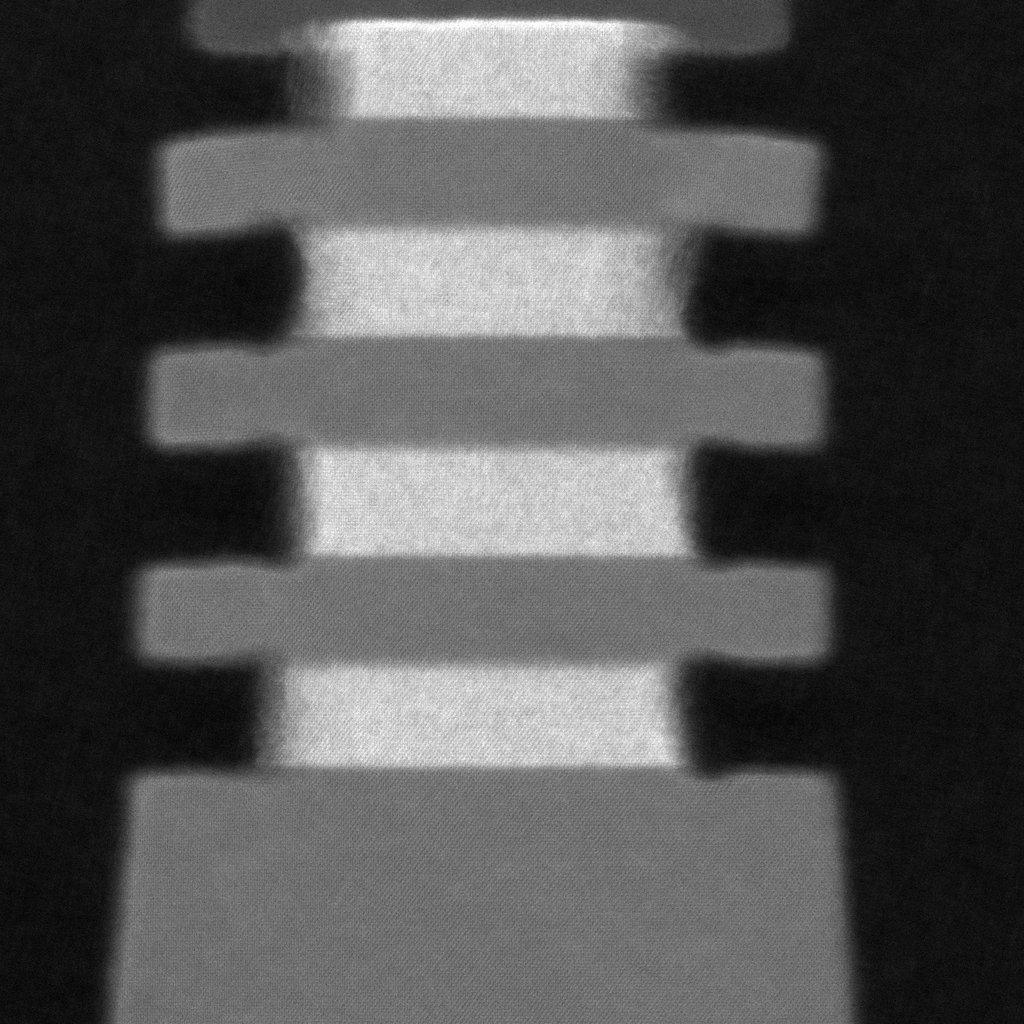}};
\end{tikzpicture}